%% file: mam.tex
\documentclass{article} %
\usepackage[
  backref=page,
  bookmarks=false,]{hyperref} 
\usepackage[accepted]{icml2024}

\usepackage{multicol}
\usepackage{url}            %
\usepackage{booktabs}       %
\usepackage{amsmath}
\usepackage{amsthm}
\usepackage{amssymb}
\usepackage{bbm}
\usepackage{nicefrac}       %
\usepackage{microtype}      %
\usepackage{wrapfig}
\usepackage{graphicx}
\usepackage{subcaption}
\usepackage{enumitem}
\usepackage{pgffor}
\usepackage{pgfmath}
\usepackage{pdfpages}
\usepackage{graphicx}
\usepackage{float}
\usepackage{comment}
\usepackage{tabularx}
\usepackage{tablefootnote}
\usepackage{svg}

\usepackage[most]{tcolorbox}
\colorlet{shadecolor}{gray!20}
\definecolor{commentcolor}{RGB}{110,154,155}   %

\usepackage{algpseudocode}
\usepackage{algorithm}
\usepackage{ragged2e}
\usepackage[capitalize,noabbrev]{cleveref}
\usepackage{macros}
\usepackage[framemethod=TikZ]{mdframed}
\usepackage{colortbl}
\usepackage{listings}
\usepackage{tcolorbox}
\usepackage{soul}

\newtcolorbox{myquote}{colback=teal!10!white, colframe=gray!75!black, boxrule=0.5pt}

\newcommand{\ours}{{\fontsize{10pt}{13.6pt}\textsc{MaM}}}
\newcommand{\sv}{{\bigtriangleup}}
\newcommand{\am}{{\blacklozenge}}
\newcommand{\halfam}{{\lozenge}}

\renewcommand*{\backref}[1]{}
\renewcommand*{\backrefalt}[4]{%
  \ifcase #1 \hfill(Not cited)%
  \or        \hfill(page~#2)%
  \else      \hfill(pages~#2)%
  \fi}

\begin{document}

\twocolumn[
\icmltitle{Generative Marginalization Models}

\icmlsetsymbol{equal}{*}

\begin{icmlauthorlist}
\icmlauthor{Sulin Liu}{yyy}
\icmlauthor{Peter J. Ramadge}{yyy}
\icmlauthor{Ryan P. Adams}{yyy}
\end{icmlauthorlist}

\icmlaffiliation{yyy}{Princeton University}

\icmlcorrespondingauthor{Sulin Liu}{sulinl@princeton.edu}

\icmlkeywords{Machine Learning, ICML}

\vskip 0.3in
]

\printAffiliationsAndNotice{}  %

\begin{abstract}
  We introduce \textit{marginalization models} (\ours{}s), a new family of generative models for high-dimensional discrete data. They offer scalable and flexible generative modeling 
  by explicitly modeling all induced marginal distributions. Marginalization models enable fast approximation of arbitrary marginal probabilities with a single forward pass of the neural network, which overcomes a major limitation of arbitrary marginal inference models, such as any-order autoregressive models. 
  \ours{}s also address the scalability bottleneck encountered in training any-order generative models for high-dimensional problems under the context of \textit{energy-based training}, where the goal is to match the learned distribution to a given desired probability (specified by an unnormalized log-probability function such as energy or reward function).
  We propose scalable methods for learning the marginals, grounded in the concept of ``\textit{marginalization self-consistency}''.
  We demonstrate the effectiveness of the proposed model on a variety of discrete data distributions, including images, text, physical systems, and molecules, for \textit{maximum likelihood} and \textit{energy-based training} settings. \ours{}s achieve orders of magnitude speedup in evaluating the marginal probabilities on both settings. For energy-based training tasks, \ours{}s enable any-order generative modeling of high-dimensional problems beyond the scale of previous methods. Code is available at \href{https://github.com/PrincetonLIPS/MaM}{github.com/PrincetonLIPS/MaM}.
\end{abstract}

\input{1_introduction}

\input{2_background}

\input{3_methodology}
\input{4_training}

\input{5_related}

\input{6_experiment}
\input{7_conclusion}

\section*{Acknowledgments}
We thank members of the Princeton Laboratory for Intelligent Probabilistic Systems and anonymous reviewers for valuable discussions and feedback. We thank Andrew Novick and Eric Toberer for valuable discussions on energy-based training in scientific applications. We thank Akshay Subramanian and Soojung Yang for valuable discussions on designing the drug-or-photodiode out-of-distribution evaluation task. This work was partially supported by NSF grants IIS-2007278 and OAC-2118201.

\section*{Impact Statement}
As a deep learning model, \ours{} has the risk of low robustness on data from unseen domain or manifold. In practice, one should not blindly apply it to data that is far away from the training data distribution and expect the marginal likelihood estimate can be trusted. For the same reason, \ours{} will also be susceptible to adversarial attacks just as other commonly deep learning models.

MaM enables training of a new type of generative model. Access to fast marginal likelihood is helpful for many downstream tasks such as outlier detection, protein/molecule design or screening. By allowing the training of order-agnostic discrete generative models scalable for distribution matching, it enhances the flexibility and controllability of generation towards a target distribution. This also poses the potential risk of deliberate misuse, leading to the generation of content/designs/materials that could cause harm to individuals.

\newpage
\bibliography{genbib}
\bibliographystyle{icml2024}

\input{appendix}

\end{document}

%% file: 1_introduction.tex
\section{Introduction}
Deep generative models have %
enabled remarkable progress 
across diverse fields, 
including image generation, audio synthesis, natural language modeling, and scientific discovery. However, %
there remains a pressing need to better support efficient probabilistic inference for key questions involving  marginal probabilities
$p(\boldx_\mcS)$ and conditional probabilities $p(\boldx_\mcU|\boldx_\mcV)$, 
for appropriate subsets $\mcS,\mcU,\mcV$ of the 
variables. The ability to
directly address such quantities 
is critical in 
applications such as outlier or machine-generated content detection~\cite{ren2019likelihood, mitchell2023detectgpt}, masked language modeling~\cite{devlin2018bert, yang2019xlnet}, image inpainting~\citep{yeh2017semantic}, and constrained protein/molecule design~\citep{wang2022scaffolding, schneuing2022structure}. Furthermore, the capacity to conduct such inferences for arbitrary subsets of variables empowers users 
to leverage the model according to their specific needs and preferences. For instance, in protein design, 
scientists may want to manually guide the generation of a protein from a user-defined substructure under a particular path over the relevant variables. 
This requires the generative model to perform arbitrary marginal inferences.

\begin{figure}[ht!]
  \centering
  \includegraphics[width=0.7\linewidth]{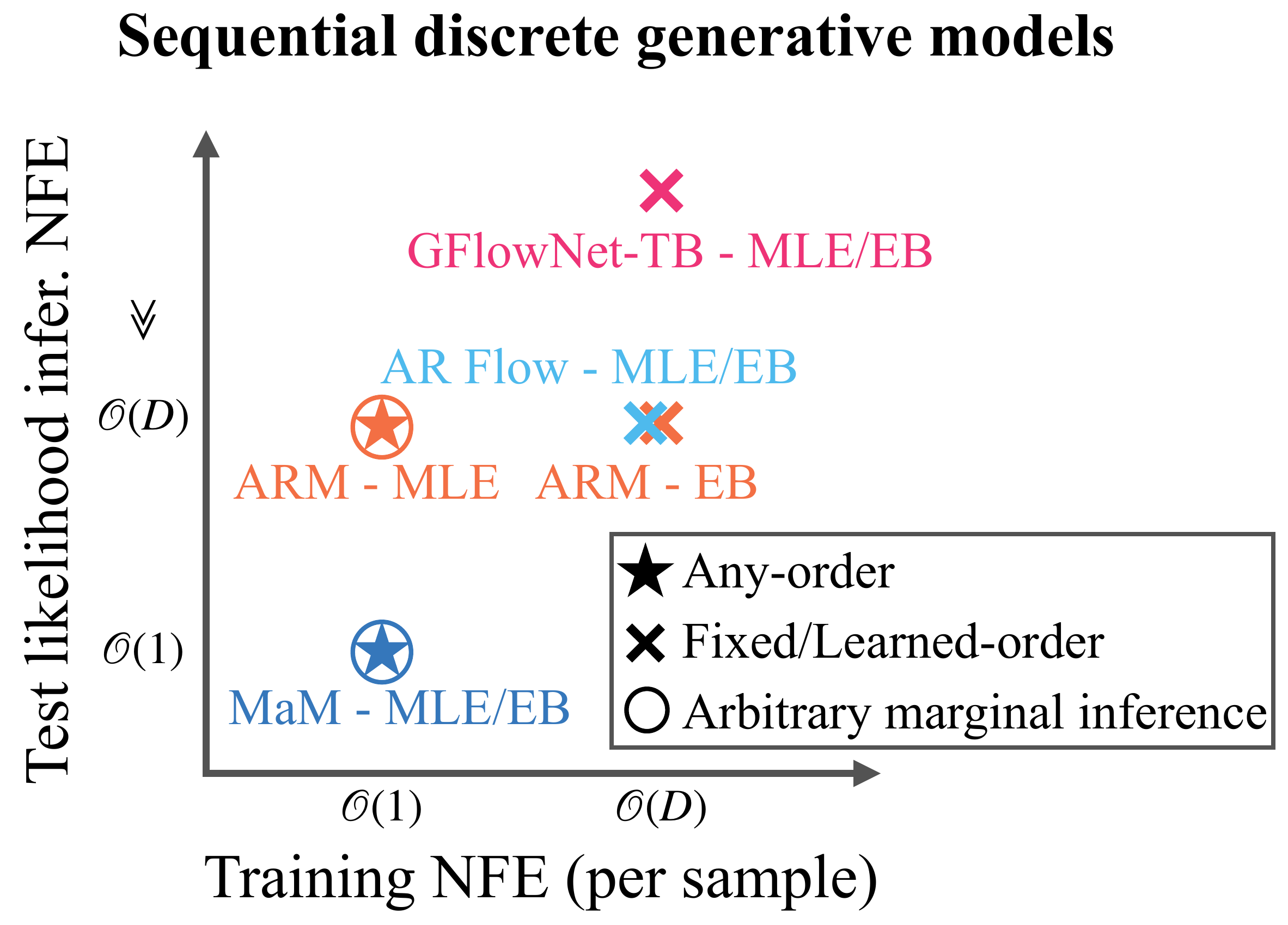}
  \caption{\small Training and test time scalability of sequential discrete generative models. The unit is number of function (i.e. NN) evaluations (NFE).}
  \label{fig:compare_methods} 
\end{figure}

\begin{figure*}[t]
  \centering
  \includegraphics[width=1.0\textwidth]{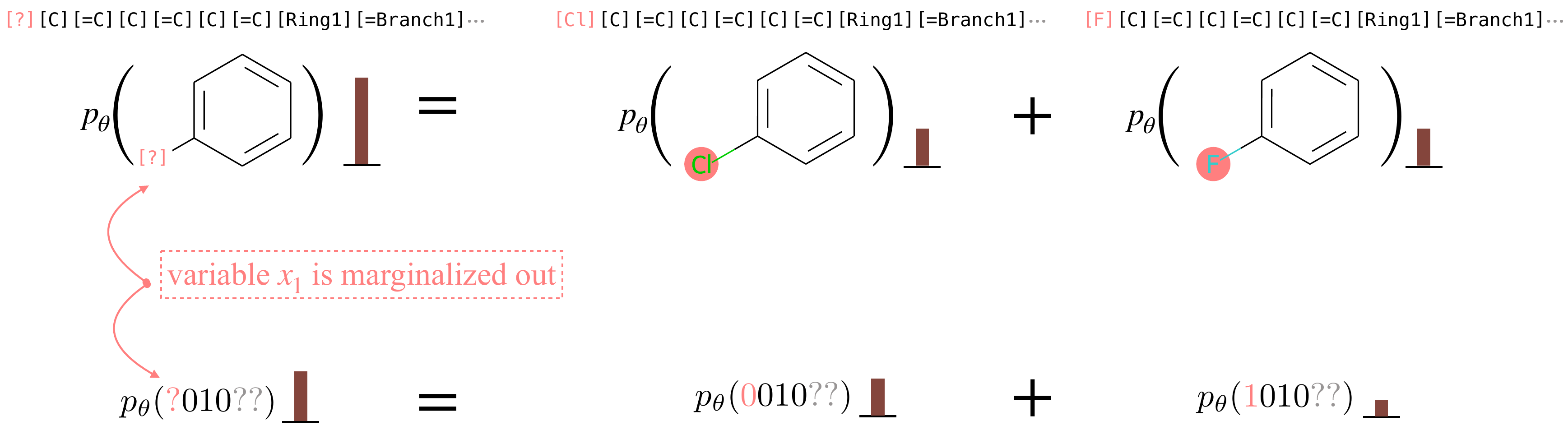}
  \caption[figure of marg]{\small Marginalization models (\ours{}s) enable estimation of any marginal probability with a neural network $\theta$ that learns to ``marginalize out'' variables (represented by ``$?$''). The figure illustrates marginalization of a single variable on bit strings (representing molecules) with two alternatives for clarity (versus $K$ in general). The bars represent probability masses.}
  \label{fig:marginalization}
\end{figure*}

Towards this end, neural autoregressive models (ARMs)~\citep{bengio2000taking, larochelle2011neural} have shown great performance in conditional/marginal inference based on the idea of modeling a high-dimensional joint distribution as a factorization of univariate conditionals using the chain rule of probability. Many efforts have been made to scale up ARMs and enable any-order generative modeling under the setting of maximum likelihood estimation (MLE)~\citep{larochelle2011neural, uria2014deep, hoogeboom2021autoregressive}, and great progress has been made in applications such as masked language modeling~\citep{yang2019xlnet} and image inpainting~\citep{hoogeboom2021autoregressive}. However, marginal likelihood evaluation on a sequence of $D$ variables is limited by $\mcO(D)$ neural network passes with the most widely-used modern neural network architectures (e.g., Transformers~\citep{vaswani2017attention} and U-Nets~\citep{ronneberger2015u}). This scaling makes it difficult to evaluate likelihoods on long sequences arising in data such as natural language and proteins.  
In addition to MLE, the setting of \textit{energy-based training} (EB) has recently received growing interest with its applications in science domains~\citep{noe2019boltzmann, damewood2022sampling, kohler2023rigid}.
Instead of empirical data samples, we only have access to an unnormalized (log) probability function (specified by a reward or energy function) that can be evaluated pointwise for the generative model to match. 
In such settings, ARMs are limited to fixed-order generative modeling and lack scalability in training. The subsampling techniques developed to scale the training of conditionals for MLE are no longer applicable when matching log probabilities in energy-based training (see \Cref{sec:4_arm_limits} for details).

To enhance scalability and flexibility in the generative modeling of discrete data,
we propose a new family of generative models, \textbf{marginalization models} (\ours{}s), that directly model the marginal distribution $p(\boldx_\mcS)$ for any subset of variables $\boldx_\mcS$ in $\boldx$. Direct access to marginals has two important advantages: 1) \textit{significantly speeding up inference for any marginal}, and 2) \textit{enabling scalable training of any-order generative models under both MLE and EB settings}.

The unique structure of the model allows it to simultaneously represent the coupled collection of all marginal distributions of a given discrete joint probability mass function.
For the model to be valid, it must be consistent with the sum rule of probability, a condition we refer to as ``\textit{marginalization self-consistency}'' (see \Cref{fig:marginalization}); learning to enforce this with scalable training objectives is one of the key contributions of this work.

We show that \ours{}s can be trained under both maximum likelihood and energy-based training settings with scalable learning objectives. We demonstrate the effectiveness of \ours{}s in both settings on a variety of discrete data distributions, including binary images, text, physical systems, and molecules. We empirically show that \ours{}s achieve orders of magnitude speedup in marginal likelihood evaluation. For energy-based training, \ours{}s are able to scale training of any-order generative models to high-dimensional problems that previous methods fail to achieve.

%% file: 2_background.tex
\section{Background}\label{sec:background}

We first review two prevalent settings for training a generative model: \textit{maximum likelihood estimation} and \textit{energy-based training}. Then we introduce autoregressive models.

\textbf{Maximum likelihood (MLE)}\quad
Given a dataset $\mcD = \{\boldx^{(i)}\}_{i=1}^N$ drawn i.i.d. from a data distribution $p = p_\text{data}$, we aim to learn the distribution $p_\theta(\boldx)$ via maximum likelihood estimation:
\begin{equation}
    \max_\theta \; \bbE_{\boldx \sim p_{\text{data}}} \left[\log p_\theta(\boldx)\right] \approx \frac{1}{N}\sum\nolimits_{i=1}^N \log p_\theta(\boldx^{(i)}) \label{eq:ml_obj}
\end{equation}
which is equivalent to minimizing the Kullback-Leibler divergence under the empirical distribution, i.e., minimizing $ D_\text{KL}(p_\text{data}(\boldx) || p_{\theta}(\boldx))$. This is the setting that is most commonly used in generation of images (e.g., diffusion models \citep{sohl2015deep, ho2020denoising, song2019generative}) and language (e.g. GPT \citep{radford2019language}) where we can easily draw observed data from the distribution.

\textbf{Energy-based training (EB)}\quad
In other cases, data from the distribution are not always available. Instead, we have access to an unnormalized probability distribution $f(\cdot)$ typically specified as  ${f(\boldx) = \exp(r(\boldx)/\tau)}$ where $r(\boldx)$ is an energy (or reward) function and $\tau>0$ is a temperature parameter.
In this setting, the objective is to match $p_\theta(\boldx)$ to $f(\boldx)/Z$, where $Z$ is the normalization constant of $f$. This can be done by minimizing the KL divergence
~\citep{noe2019boltzmann, wu2019solving,damewood2022sampling},
\begin{equation}
\min_\theta \; D_\text{KL}
\left(p_{\theta} (\mathbf{x}) \| \frac{f(\mathbf{x})}{Z} \right) 
    \label{eq:eb_kl_obj}
 \end{equation}
The reward function $r(\boldx)$ can be defined either by human preferences or by the physical system from first principles. 
For example, 
(a) In aligning large language models, $r(\boldx)$ can represent human preferences \citep{ouyang2022training, chatgpt};
(b) In molecular/material design, it can specify the proximity of a sample's measured or calculated properties to some functional desiderata \citep{bengio2021flow}; and 
(c) In modeling the thermodynamic equilibrium ensemble of physical systems, it is the (negative) energy function of a given sample \citep{noe2019boltzmann,wu2019solving,damewood2022sampling,kohler2023rigid}.

The training objective in \Cref{eq:eb_kl_obj} can be optimized using a Monte Carlo estimate of the gradient using the REINFORCE algorithm \citep{williams1992simple}. A generative model $\theta$ allows us to efficiently generate samples approximately from the distribution, which would otherwise be much more expensive via running MCMC with the energy function $f(\cdot)$.

\textbf{Autoregressive models}\quad
Autoregressive models (ARMs) \citep{bengio2000taking,larochelle2011neural} model a complex high-dimensional distribution $p(\boldx)$ by factorizing it into univariate conditionals using the chain rule:
\begin{align}
    \log p_\phi(\boldx)=\sum\nolimits_{d=1}^D \log p_\phi\left(x_d \mid \boldx_{<d}\right),
\end{align}
where $\boldx_{<d} = \{x_1, \ldots, x_{d-1}\}$.
ARMs generate examples by sequentially drawing $x_1$ under $p_\phi(x_1),$ then $x_2$ under $p_\phi(x_2 | x_1),$ and so on. 
The ARM approach has produced successful discrete-data neural models for natural language, proteins~\citep{shin2021protein, lin2023evolutionary, madani2023large}, and molecules~\citep{segler2018generating, flam2022language}.

\textbf{Any-order ARMs (AO-ARMs)}\quad
~\citet{uria2014deep} propose to learn the conditionals of ARMs for arbitrary orderings that include all permutations of $\{1,\ldots,D\}$. Under the MLE setting, the model $\phi$ is trained by maximizing a lower-bound objective \citep{uria2014deep,hoogeboom2021autoregressive} using an expectation under the uniform distribution of orderings.
This objective allows scalable training of AO-ARMs with architectures such as the U-Net~\citep{ronneberger2015u} and Transformers~\citep{vaswani2017attention}, by leveraging efficient parallel evaluation of multiple one-step conditionals for all next-tokens in one forward pass.
However, modeling conditionals alone with ARMs results in limitations in both inference and training (more details in \Cref{sec:4_arm_limits}):
\begin{enumerate}
    \item \textit{Test-time marginal likelihood inference}: evaluation of $p_\phi(\boldx)$ or $p_\phi(\boldx_s)$ requires up to $D$ neural network passes, making it costly for high-dimensional data.
    \item \textit{Energy-based training for high-dimensional problems:} the objective in \Cref{eq:eb_kl_obj} requires evaluating $\log p_\phi(\boldx)$ in full with $D$ network forward passes in order to calculate the difference of $\log p_\phi(\boldx)$ and $f(\boldx)/Z$. Monte Carlo estimate of $\log p_\phi(\boldx)$ no longer works since the objective is matching $\log p$'s instead of maximizing $\log p$ (the MLE case). As a result, this significantly limits ARM's training scalability under the EB setting when $D$ becomes large.
\end{enumerate}

%% file: 3_methodology.tex
\section{Marginalization Models}
\label{sec:method}

We propose \textit{marginalization models} (\ours{}s), a new type of generative model that enables scalable any-order generative modeling on high-dimensional problems as well as efficient marginal evaluation, for both maximum likelihood and energy-based training. The flexibility and scalability of marginalization models are enabled by the explicit modeling of the marginal distribution and scalable training objectives derived from \textit{marginalization self-consistency}.

In this paper, we focus on generative modeling of discrete structures using vectors of discrete variables. The vector representation encompasses various real-world problems with discrete structures, including language sequence modeling, protein design, and molecules with string-based representations (e.g., SMILES~\citep{weininger1989smiles} or SELFIES~\citep{krenn2020self}). Moreover, vector representations are inherently applicable to any discrete problem, since it is feasible to encode any discrete object into a vector of discrete variables.

\textbf{Definition}\quad Let $p(\boldx)$ be a discrete probability distribution, where $\boldx = (x_1, \ldots, x_D)$ is a $D$-dimensional vector and each $x_d$ takes $K$ possible values, i.e. $x_d \in \mcX \triangleq \{1,\ldots,K\}$.

\textbf{Marginalization}\quad 
For a subset of indices $\mathcal{S} \subseteq \{1, \ldots, D\}$, let $\boldx_\mathcal{S}$ and $\boldx_{\mathcal{S}^c}$ denote the subvectors corresponding to $\mathcal{S}$ and the complement set, $\mathcal{S}^c = \{1, \ldots, D\} \setminus \mathcal{S}$.
The marginal of $\boldx_\mcS$ is obtained by summing over all values of $\boldx_{\mcS^c}$: 
\begin{equation}
  p(\boldx_\mcS) = \sum\nolimits_{\boldx_{\mcS^c}} p(\boldx_\mcS, \boldx_{\mcS^c}) \label{eq:marginal}
\end{equation}
We refer to~\eqref{eq:marginal} as the ``\textit{marginalization self-consistency}'' that a valid distribution should follow. 
The goal of a marginalization model $\theta$ is to estimate the marginals $p(\boldx_\mcS)$ for any subset of variables $\boldx_\mcS$ as closely as possible. To achieve this, we train a deep neural network that fits $p_\theta(\boldx)$ to a target distribution $p(\boldx)$ while fitting the marginals $p_\theta(\boldx_\mcS)$ through the marginalization self-consistency principle.
In other words, \ours{} learns to approximately inference the marginals of an arbitrary subset of variables with a single neural net forward pass.\footnote{Estimating $p(\boldx)$ is a special case of marginal inference where there are no variables to be marginalized.}

\textbf{Parameterization}\quad
A marginalization model parameterized by a neural network $\theta$ takes in $\boldx_\mcS$ and outputs the marginal log probability $ f_\theta(\boldx_\mcS) = \log p_\theta(\boldx_\mcS)$. Note that for different subsets $\mcS$ and $\mcS^\prime$, $\boldx_\mcS$ and $\boldx_\mcS^\prime$ lie in different vector spaces. To unify the vector space that is fed into the NN, we introduce an augmented vector space that additionally includes the ``marginalized out'' variables $\boldx_{\mcS^c}$. By defining a special symbol ``$\sv$'' to denote the missing values of the ``marginalized out'' variables, the augmented vector representation is $D$-dimensional and is defined to be:
$
\boldx_\mcS^\text{aug} (i) = \begin{cases} x_i, & \text{if } i \in \mcS\\ \sv, & \text{otherwise} \end{cases}.
$
Now, the augmented vector representation $\boldx_\mcS^\text{aug}$ of all possible $\boldx_\mcS$'s has the same dimension $D$, and for any $i$-th dimension $\boldx_\mcS^\text{aug}(i) \in \mcX^\text{aug} \triangleq \{1,\cdots,K,\sv\}$. To given an example, when $D=4$ and $\mcX = \{0,1\}$, for ${\boldx_\mcS = \{x_1, x_3\}}$ taking values $x_1=0$ and $x_3=1$, ${\boldx_\mcS^\text{aug} = (0, \sv, 1, \sv)}$, with the corresponding marginal $p(\boldx_\mcS^\text{aug}) = \sum_{x_2}\sum_{x_4} p(0,x_2,1,x_4)$. From here onwards we will use $\boldx_\mcS^\text{aug}$  and $\boldx_\mcS$ interchangeably.

\textbf{Sampling}\quad
With the marginalization model, one can sample from the learned distribution by picking an arbitrary order and sampling one variable or multiple variables at a time. 
In this paper, we focus on the sampling procedure that generates one variable at a time. Sampling multiple variables jointly can also be done in a similar way (see \Cref{sec:sample_with_marginal} for ablation studies).
To get the conditionals at each step for generation, we can use the product rule of probability: $ p_\theta(x_{\sigma(d)} | \boldx_{\sigma(<d)}) = p_\theta(\boldx_{\sigma (\leq d)})/ p_\theta(\boldx_{\sigma (< d)}).$
However, the above conditional distribution is not exactly valid when the following single-step marginalization consistency in \eqref{eq:one_step_marginal} is only approximately enforced,
\begin{align}
  & p_\theta(\boldx_{\sigma (< d)}) \approx \sum\nolimits_{x_{\sigma (d)}} p_\theta(\boldx_{\sigma (\leq d)}), \label{eq:one_step_marginal} \\
  & \forall \sigma \in S_D, \boldx \in \{1,\cdots,K\}^D, d \in [1:D] \,, \nonumber
\end{align}
since the estimated probabilities might not sum up exactly to one. Hence we use following normalized conditional:
\begin{equation}
  p_\theta(x_{\sigma(d)} | \boldx_{\sigma(<d)}) = \frac{p_\theta([\boldx_{\sigma (< d)}, x_{\sigma(d)}])} {\sum\nolimits_{x_{\sigma (d)}} p_\theta([\boldx_{\sigma (< d)}, x_{\sigma(d)}])} \,.
\end{equation}
\textbf{Scalable training of marginalization self-consistency}\quad 
In training, we can impose the marginalization self-consistency by minimizing the \textit{squared error} of the constraints in~\eqref{eq:one_step_marginal} in log-space.
Evaluation of each marginalization constraint in~\eqref{eq:one_step_marginal} requires $K$ NN forward passes, where $K$ is the number of discrete values $x_d$ can take. This makes mini-batch training challenging to scale when $K$ is large. To address this issue, we augment the marginalization models with learnable conditionals parameterized by another neural network $\phi$. The marginalization constraints in~\eqref{eq:one_step_marginal} can be further decomposed into $K$ parallel marginalization constraints\footnote{To make sure $p_\theta$ is normalized, we can either additionally enforce $p_\theta\left(\left(\sv, \cdots, \sv \right)\right) = 1$ or let $Z_\theta = p_\theta\left(\left(\sv, \cdots, \sv \right)\right)$ be the normalization constant.}. 
\begin{align}
  & p_\theta(\boldx_{\sigma (< d)}) p_\phi(\boldx_{\sigma (d)} | \boldx_{\sigma (< d)}) \approx p_\theta(\boldx_{\sigma (\leq d)}),  \label{eq:one-step-marginal-conditional}\\
  & \forall \sigma \in S_D, \boldx \in \{1,\cdots,K\}^D, d \in [1:D]. \nonumber
\end{align}
The consistency error for each constraint can be defined correspondingly as follows:
\begin{align*}
& \text{ConsistencyError}(\boldx, \sigma, d) \\
= & \left[ \log \left( p_\theta(\boldx_{\sigma (< d)}) p_\phi(\boldx_{\sigma (d)} | \boldx_{\sigma (< d)})\right) - \log p_\theta(\boldx_{\sigma (\leq d)}) \right]^2.
\end{align*}
Other distances such as KL divergence can also be considered. We choose squared distance for its flexibility in selecting the $q(\boldx)$, allowing us to fit marginals for various use cases with different $q(\boldx)$ at test time.
It's also worth noting that training with KL divergence and squared distance are quite similar (see \citet{malkin2022gflownets}). The REINFORCE gradient of 
$D_\text{KL}\left(p_\theta(\mathbf{x}_{<\sigma(d)}) p_\phi(x_{\sigma(d)} | \mathbf{x}_{<\sigma(d)}) \parallel p_\theta(\mathbf{x}_{\leq \sigma(d)}).\mathrm{detach}() \right) 
$
is equivalent to the squared distance loss when $q$ is set to $p_\theta$.

By breaking the original marginalization self-consistency in \Cref{eq:marginal} into highly parallel marginalization self-consistency in \Cref{eq:one-step-marginal-conditional}, we introduce a total of $K^D{\times}D!{\times}D{\times}K$ constraints.
Although this increases the number of constraints, it becomes \textit{highly scalable} to train on the marginalization self-consistency via randomly sampling constraints following a specified distribution $q(\boldx)$ and $q(\sigma)$. In our experiments, $q(\sigma)$ is set to the uniform distribution over all orderings $\mcU(S_D)$ and $q(\boldx)$ is set to the data distribution of interest for marginal inference, such as the empirical data distribution $p_\text{data}(\boldx)$ or the generative model's distribution $p_{\theta,\phi}(\boldx)$. We found that a training objective that decomposes into highly parallel terms for sampling is key to effectively fitting marginals with scalability.

%% file: 4_training.tex
\section{Training the Marginalization Models}
\label{sec:training}

\subsection{Maximum Likelihood Estimation Training}
\label{subsec:mle}
In this setting, we train \ours{}s with the maximum likelihood objective while additionally enforcing the marginalization constraints in \Cref{eq:one_step_marginal}:
\begin{align}
  \max_{\theta, \phi} \quad & \bbE_{\boldx \sim p_{\text{data}}} \log p_\theta(\boldx)   \label{eq:mar_ml_constr}
 \\
  \text{s.t.} \quad & p_\theta(\boldx_{\sigma (< d)}) p_\phi(\boldx_{\sigma (d)} | \boldx_{\sigma (< d)}) \approx p_\theta(\boldx_{\sigma (\leq d)}), \nonumber \\
  & \forall \sigma \in S_D, \boldx \in \{1,\cdots,K\}^D, d \in [1:D]. \nonumber
\end{align}

\textbf{Two-stage training}\quad
A typical way to solve the above optimization problem is to convert the marginalization constraint into another objective and optimize both objectives jointly. However,
maximizing $\log p_\theta(\boldx_{\leq D})$ directly in \Cref{eq:mar_ml_constr} is unbounded, we empirically found this causes the training to be slow and unstable by over-emphasizing likelihood at the expense of self-consistency.
Instead, we identify an theoretically equivalent two-stage optimization formulation that leads to more effective training strategy based on the following observation:

\begin{proposition}  \label{thm:mar_ml}
  Solving the optimization problem in \eqref{eq:mar_ml_constr} is equivalent to the following two-stage optimization procedure, under mild assumptions about the neural networks used being universal approximators:
  \begin{align*}
     & \text{\normalfont \bf Stage 1:}\max_{\phi} \; \bbE_{\boldx \sim p_{\text{data}}} \bbE_{\sigma}  \;
     \sum\nolimits_{d=1}^D \log p_\phi\left(x_{\sigma(d)} | \boldx_{\sigma(<d)}\right)
    \nonumber \\
     & \text{\normalfont \bf Stage 2:}
    \min_{\theta} \; \bbE_{\boldx \sim q(\boldx)} \bbE_{\sigma} \bbE_{d}  \;
    \text{ConsistencyError}(\boldx, \sigma, d)
  \end{align*}
where $\sigma \sim \mcU(S_D)$, $d \sim \mcU(1,\cdots,D)$ and $q(\boldx)$ is the distribution of interest for marginal likelihood inference.
 
\end{proposition}

The first stage can be interpreted as \textit{fitting the conditionals} in the same way as AO-ARMs~\citep{uria2014deep,hoogeboom2021autoregressive} and the second stage acts as \textit{distilling the marginals} from conditionals through training on \textit{marginalization self-consistency}.
The intuition comes from the simple chain rule of probability: we first observe a one-to-one correspondence between the optimal conditionals $\log p_\phi$ and marginals $\log p_\theta$, i.e. $\log p_\theta(\boldx) = \sum\nolimits_{d=1}^{D}  \log p_\phi\left(x_{\sigma(d)} |\boldx_{\sigma(<d)}\right)$ for any $\sigma$ and $\boldx$.  By assuming neural networks are universal approximators, we can split the joint optimization problem into two steps by first finding the optimal conditionals $p_\phi$, and then solving for the corresponding optimal marginals $p_\theta$. 
We provide proof details in \Cref{app:claim_proof}. 

There are two main advantages with the reformulated two-stage training. First, the maximum likelihood objective based on conditionals is now bounded and can be optimized in parallel. Secondly, even when compared with joint training with the reformulated conditional-based likelihoods, the decoupled two-stage training leads to improved efficiency, since it avoids wasted compute on fitting marginals on conditionals that are still being actively updated throughout training. Additionally, the two-stage approach eliminates the need to sweep over the hyperparameter that balances the two objectives. In \Cref{subsec:exp_two_stage_joint}, we validate this with experiments by comparing two-stage v.s. joint training, both using the reformulated objectives.
This aligns with findings in diffusion model distillation \citep{song2023consistency, berthelot2023tract}, where training a standard diffusion model followed by distillation proves easier than training a distilled model from scratch.

\subsection{Energy-based Training}

In this setting, the two-stage training introduced in \Cref{subsec:mle} becomes impractical for high-dimensional problems. Stage 1 training (fitting conditionals with $ \mathcal L_\mathrm{KL} = \mathbb E_{p_\theta} \left[ \sum_{d=1}^D \log p_\phi \left( x_{\sigma(d)} \mid \mathbf x_{\sigma(<d)}\right) - \log p(x) \right] $ ) scales poorly with $D$ as it requires $D$ NN forward passes per datapoint.
Therefore, for scalability, we train \ours{}s by jointly minimizing the KL divergence objective over the marginals and the self-consistency loss term that include both marginals and conditionals:
\begin{equation}
  \min_{\theta, \phi}  D_\text{KL}\!\left( p_{\theta} \! \parallel\! p \right) + \lambda \,\bbE_{\boldx \sim q(\boldx)} \bbE_{\sigma} \bbE_{d} \text{ConsistencyError}(\boldx, \sigma, d),
  \nonumber %
\end{equation}
where $\sigma \sim \mcU(S_D)$, $d \sim \mcU(1,\cdots,D)$ and $q(\boldx)$ is the distribution of interest for marginal likelihood inference.

Unlike the unbounded likelihood maximization in \Cref{subsec:mle}, matching $\log p_\theta(\mathbf{x})$ with $\log p(\mathbf{x})$ in the KL term does not lead to training instability issues. However, joint training introduces complex dynamics, necessitating careful hyperparameter selection. We find that a wide range of small $\lambda$ yield best performance. More experiments and discussion are provided in Section \ref{subsec:exp_two_stage_joint}.

\textbf{Scalable training}\quad
The gradient of the KL divergence term is estimated with the REINFORCE estimator \citep{williams1992simple}:
\begin{align}
  & \nabla_\theta D_\text{KL}(p_{\theta}(\boldx)||p(\boldx)) \nonumber \\
  =& \bbE_{\boldx \sim p_{\theta}(\boldx)} \left[ \nabla_\theta \log p_{\theta}\left(\boldx\right) \left( \log p_{\theta}\left(\boldx\right) - \log f\left( \boldx \right) \right) \right]  \label{eq:reinforce} \\
  \approx& \frac{1}{N}  \sum\nolimits_{i=1}^N \nabla_\theta \log p_{\theta}(\boldx^{(i)}) \left( \log p_{\theta}(\boldx^{(i)}) - \log f(\boldx^{(i)}) \right) \nonumber
\end{align}
The consistency-error term can be estimated by randomly sampling data $\boldx$, ordering $\sigma$ and step $d$ from the specified distribution.

\textbf{Efficient sampling with persistent MCMC}\quad
To efficiently generate approximate samples of $p_\theta$ for the REINFORCE estimator, a persistent set of Markov chains are maintained by taking block-wise Gibbs sampling steps following a random ordering using the conditional distribution $p_\phi(\boldx_{\sigma (d)} | \boldx_{\sigma (< d)})$ (full algorithm in \Cref{app:algorithm}), in a similar fashion to persistent contrastive divergence~\citep{tieleman2008training}.
The samples from the conditional network $p_\phi$ serve as good approximation to samples from the marginal network $p_\theta$, since they are close to each other when conditionals and marginals are approximately consistent with each other. In experiments, we validate this by observing that the log-probabilities from $p_\theta$ and $p_\phi$ are highly consistent on both random and on-policy samples. In cases when there is a strong discrepancy between $p_\theta$ and $p_\phi$ during training, we can additionally use importance sampling to get an unbiased estimate. 

\subsection{Addressing Limitations of ARMs}\label{sec:4_arm_limits}

1) \textbf{Test-time marginal likelihood inference}\quad Evaluation of a marginal $p_\phi(\boldx_o)$ with ARMs (or an arbitrary marginal with AO-ARMs) requires applying the conditional $p_\phi$
up to $D$ times, which is inefficient in time and memory for high-dimensional data. In contrast, \ours{}s can approximate any arbitrary marginal with just one NN forward pass. This is achieved through explicitly modeling the marginals and training with scalable self-consistency objectives.

\begin{figure}[ht]
  \centering
    \vspace{-1mm}
    \includegraphics[width=0.7\linewidth]{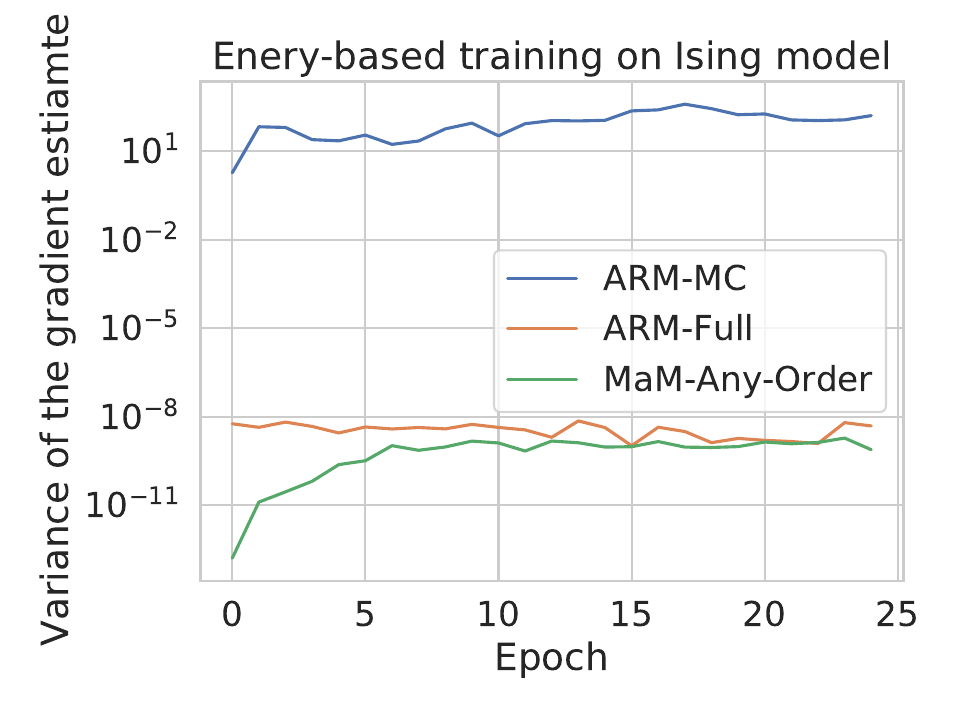}
  \vspace{-1mm}
  \caption{\small Approximating $\log p_\phi(\boldx)$ with one-step conditional (ARM-MC) results in extremely high gradient variance in energy-based training.}
  \label{fig:arm_mc_variance}
\end{figure}

2) \textbf{EB training for high-dimensional problems}\quad
There are two factors that limit the scalability of ARMs for EB training. First, the KL divergence objective in EB training requires evaluating $\log p_\phi(\boldx)$ in full with $D$ network forward passes in order to calculate the difference of $\log p_\phi(\boldx)$ and $f(\boldx)/Z$. One might consider estimating $p_\phi(\boldx)$ with a single-step Monte Carlo estimate $p_\phi(x_{\sigma(d)} | \boldx_{\sigma(<d})$, but this leads to high variance of the REINFORCE gradient in \Cref{eq:reinforce} due to the product of the score function and distance terms, which are both of high variance (validated in experiments, see \Cref{fig:arm_mc_variance}). Consequently, training ARMs for energy-based training necessitates a sequence of $D$ conditional evaluations to compute the gradient of the objective function. This constraint leads to an effective batch size of $B \times D$ for batch of $B$ samples, significantly limiting the training scalability of ARMs to high-dimensional problems. 

\ours{}s circumvent the first limiting factor by breaking down the original distribution matching problem into two sub-problems: 1) minimizing the KL divergence between the model's marginal probability estimate $p_\theta(\boldx)$ and the energy function $f(\boldx)$, and 2) ensuring marginals $\log p_\theta$ and conditionals $\log p_\phi$ are self-consistent. The first sub-problem requires evaluating the marginal probability $p_\theta(\boldx)$ with just one network forward pass for each $\boldx$ sample. The training objective for the second sub-problem is also scalable via simply sampling the highly parallel self-consistency error objectives developed in \Cref{eq:one-step-marginal-conditional}.

The other limiting factor is associated with obtaining Monte Carlo samples for the REINFORCE gradient estimator. 
Previous methods that use ARMs for energy-based training \citep{wu2019solving, damewood2022sampling} assume a fixed ordering and require $D$ sequential sampling steps to generate samples, which is slow and costly when the dimension is large. \ours{} proposes a more cost-effective sampling procedures through the utilization of persistent block-wise Gibbs sampling.

%% file: 5_related.tex
\section{Related Work}

\textbf{Autoregressive models}\quad
Developments in deep learning have greatly advanced the performance of ARMs across different modalities, including images, audio, and text. Any-order (Order-agnostic) ARMs were first introduced in~\citep{uria2014deep} by training with the any-order lower-bound objective for the maximum likelihood setting. Recent work, ARDM~\citep{hoogeboom2021autoregressive}, demonstrates state-of-the-art performance for any-order discrete modeling of image/text/audio.
\citet{germain2015made} train an auto-encoder with masking that outputs the sequence of all one-step conditionals for a given ordering, but does not perform as well as methods \citep{van2016pixel, yang2019xlnet, hoogeboom2021autoregressive} that predict one-step conditionals under the given masking.
\citet{douglas2017universal} train an AO-ARM as a proposal distribution and uses importance sampling to estimate arbitrary conditional probabilities in a DAG-structured Bayesian network, but with limited experiment validation on a synthetic dataset. \citet{shih2022training} utilizes a modified training objective of ARMs for better marginal inference performance but loses any-order generation capability. In the energy-based training setting, ARMs are applied to science problems \citep{damewood2022sampling, wu2019solving}, but suffer in scaling to when $D$ is large.
\ours{}s and ARMs are compared in detail in \Cref{sec:4_arm_limits}.

\textbf{Arbitrary conditional/marginal models}\quad
For continuous data, VAEAC~\citep{ivanov2018variational} and ACFlow~\citep{li2020acflow} extend conditional variational encoder and normalizing flow to arbitrary conditional modeling. ACE \citep{strauss2021arbitrary} improves the expressiveness of arbitrary conditional models by directly modeling the energy function, which reduces the constraints on parameterization but increases computation costs due to the need to approximate the normalizing constant. 
Instead of using neural networks as function approximators, probabilistic circuits (PCs)~\citep{chow1968approximating, darwiche2003differential, poon2011sum, choi2020probabilistic,peharz2020einsum} offer tractable probabilistic models for both conditionals and marginals by building a computation graph with sum and product operations following specific structural constraints. \citet{peharz2020einsum} improved the scalability of PCs by combining arithmetic operations into a single monolithic einsum-operation and automatic differentiation. Further improvements of PCs are achieved through distilling latent variables from pre-trained deep generative models \citep{liu2022scaling, liu2023understanding} . All methods mentioned above focus on MLE settings.
\ours{}s focus on scalable approximate marginal inference using neural networks as function approximators on both MLE and EB settings.

\textbf{GFlowNets}\quad
GFlowNets~\citep{bengio2021flow, bengio2023gflownet} formulate the problem of generation as matching the probability flow at terminal states to the target normalized density. Compared to ARMs, GFlowNets allow flexible modeling of the generation process by assuming learnable generation paths through a directed acyclic graph (DAG). The advantages of learnable generation paths come with the trade-off of sacrificing the flexibility of any-order generation and exact likelihood evaluation. Under a fixed generation path, GFlowNets reduce to fixed-order ARMs \citep{zhang2022unifying}. In \Cref{app:mam_gflownet}, we further discuss the connections and differences between GFlowNets and AO-ARMs/\ours{}s. For discrete problems,~\citet{zhang2022generative} train GFlowNets on the squared distance loss with the trajectory balance objective \citep{malkin2022trajectory}. This is not scalable for large $D$ (for the same reason as ARMs in \Cref{sec:4_arm_limits}) and doesn't provide direct access to marginals.
In the MLE setting, an energy function is additionally learned from data so that the model can be trained with energy-based training.

%% file: 6_experiment.tex
\section{Experiments}

\begin{figure*}[htbp!]
  \centering
  \includegraphics[width=1\textwidth]{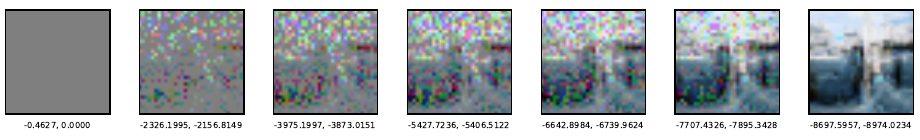}
  \caption{\small An example of the marginal estimates of an ImageNet32 image along the generation trajectory using a random ordering. The numbers in the captions show that the learned (log) marginals (left) v.s. learned (log) conditionals (right) are approximately self-consistent.}
  \label{fig:gen}
\end{figure*}

We evaluate marginalization models (\ours{}) on both MLE and EB settings for discrete problems including images, text, molecules and phyiscal systems. 
We compare \ours{}s with baselines that support arbitrary marginal inference\footnote{We use $\am$ to denote that the model supports arbitrary marginal inference. $\halfam$ is used for ARMs with fixed ordering since they only partially support arbitrary marginal inference.}: Any-order ARM$^\am$~\citep{hoogeboom2021autoregressive}, ARM$^\halfam$~\citep{larochelle2011neural}, Parallel Any-order ARMs (P-AO-ARM)~\citep{hoogeboom2021autoregressive} and Probabilistic Circuit (PC)$^\am$~\citep{peharz2020einsum}.
We also include state-of-the-art generative models on various tasks for comparison: GFlowNet~\citep{zhang2022generative}, Discrete Flow~\citep{tran2019discrete}, PixelCNN++~\citep{salimans2017pixelcnn++}, Variational Diffusion Models~\citep{kingma2021variational}, Sparse Transformers \citep{child2019generating,jun2020distribution} and D3PM~\cite{austin2021structured}. 
We follow training settings or results from the literature for all baselines. 
In \Cref{sec:test_msc}, we present additional studies on measuring the marginal self-consistency with a carefully curated synthetic experiment.
Neural network architecture and training hyperparameter details are given in \Cref{sec:app_exp}. 

\subsection{Maximum Likelihood Estimation Training}
We focus on three metrics: test data negative log likelihood (NLL), marginal inference time and marginal inference quality. The later two are only available with baselines that support arbitrary marginal inference. The marginals are evaluated on a randomly sampled mini-batch data of the test set (batch size $=128$, metrics are averaged over batches). To evaluate marginal estimation quality, the marginal estimates of each model are compared with the marginal estimates of the best-performing model (in terms of NLL). Pearson correlation is reported to measure the quality of marginal likelihoods\footnote{When measuring AO-ARM against itself, two random orderings are measured against each other.}. ($1.0$ means a perfect linear correlation with the best model's marginal estimates.)
For evaluating NLL, the conditional network and marginal network perform similarly in ablation studies (see \Cref{sec:sample_with_marginal}). We use the conditional network for evaluating NLL. The marginal network is used for evaluating marginals.

\textbf{Image}\quad
We evaluate \ours{}s on Binary MNIST \citep{salakhutdinov2008quantitative}, CIFAR-10 \citep{krizhevsky2009learning} and ImageNet32 \citep{deng2009imagenet,chrabaszcz2017downsampled}. The image dimension is $1\times 28 \times 28$ for MNIST and $3 \times 32 \times 32$ for CIFAR-10 and ImageNet32. \ours{}s achieve competitive NLL on all tasks, equaling the best-performing arbitrary marginal inference models. In terms of marginal inference, \ours{} produces \textit{high quality} marginal estimates while achieving close to \textit{$4$ orders of magnitude speed-up} in computation time. The Pearson correlation coefficients are close to $1.0$, which means the marginal estimates are consistent with the best marginal estimates. It can also be interpreted as a measure of marginalization self-consistency, since the the marginals of \ours{} are evaluated against the same conditionals of AO-ARM and \ours{}.

\begin{table}[t]
  \centering
  \caption{Pixel modeling on Binary-MNIST}
  \label{tab:mnist}
    \begin{center}
    \begin{small}
  \begin{tabular}{lccc}
    \toprule
    & {NLL (bpd) $\downarrow$} &{Pearson $\uparrow$}  & {Time (s) $\downarrow$}\\
    \midrule
    GflowNet \citep{zhang2022generative} & 0.189 & -- & -- \\
    AO-ARM$^\am$ \citep{hoogeboom2021autoregressive} & {0.146} & {0.99} & 132.4 $\pm$ 0.03\\
    PC (EiNets)$^\am$ \citep{peharz2020einsum} & 0.187 & 0.75 & {0.015 $\pm$ 0.00} \\
    \rowcolor{shadecolor} \ours{}$^\am$ & {0.146} & {0.99} & {0.018 $\pm$ 0.00} \\
    \bottomrule
  \end{tabular}
    \end{small}
    \end{center}
\end{table}

\begin{table}[t]
  \centering
  \caption{Pixel modeling on CIFAR-10}
  \label{tab:cifar10}
    \begin{center}
    \begin{small}
  \begin{tabular}{lccc}
    \toprule
    & {NLL (bpd) $\downarrow$} &{Pearson $\uparrow$}  & {Time (s) $\downarrow$}\\
    \midrule
    D3PM \citep{austin2021structured} & 3.44 & -- & --\\
    PixelCNN++ \citep{salimans2017pixelcnn++} & 2.99 & -- & -- \\
    VDM \citep{kingma2021variational} & 2.49 & -- & -- \\
    Sparse Transformer \citep{child2019generating, jun2020distribution} & 2.56 & -- & -- \\
    PC (LVD-PG)$^\am$ \citep{liu2023understanding} & 3.87 & -- & -- \\
    AO-ARM$^\am$ (800 epochs) & 2.88 & {0.99} & 2401 $\pm$ 1\\
    \rowcolor{shadecolor} \ours{}$^\am$ (800 epochs) & 2.88 & {0.98} & {0.495 $\pm$ 0.00} \\
    \bottomrule
  \end{tabular}
    \end{small}
    \end{center}
\end{table}

\begin{table}[t]
  \centering
  \caption{Pixel modeling on ImageNet32}
  \label{tab:imagenet}
    \begin{center}
    \begin{small}
  \begin{tabular}{lccc}
    \toprule
     & {NLL (bpd) $\downarrow$} &{Pearson $\uparrow$}  & {Time (s) $\downarrow$}\\
    \midrule
    Image Transformer \citep{parmar2018image} & 3.77 & -- & -- \\
    VDM \citep{kingma2021variational} & 3.72 & -- & -- \\
    PC (LVD-PG)$^\am$ \citep{liu2023understanding} & 4.06 & -- & -- \\
    AO-ARM$^\am$ (16 epochs) & 3.60 & {0.99} & 4995 $\pm$ 1\\
    \rowcolor{shadecolor} \ours{}$^\am$ (16 epochs) & 3.60 & {0.98} & {1.243 $\pm$ 0.00} \\
    \bottomrule
  \end{tabular}
    \end{small}
    \end{center}
\end{table}

\begin{table}[t!]
  \centering
  \caption{Character modeling on text8}
  \label{tab:text8}
  \small
  \begin{tabular}{lccc}
    \toprule
    & {NLL (bpc) $\downarrow$}  &{Pearson $\uparrow$}  & {Time (s) $\downarrow$}\\
    \midrule
    D3PM \citep{austin2021structured} & 1.47 & -- & -- \\
    Discrete Flow \citep{tran2019discrete} & 1.23 & -- & -- \\
    Transformer \citep{vaswani2017attention} & 1.35 & -- & -- \\
    AO-ARM$^\am$ (3000 epochs) & {1.48} & {0.987} & 41.40 $\pm$ 0.01  \\
    \rowcolor{shadecolor} \ours{}$^\am$ (3000 epochs) & {1.48} & 0.945 & {0.005 $\pm$ 0.00} \\
    \bottomrule
  \end{tabular}
\end{table}

\textbf{Molecule}\quad
We evaluate \ours{} on
the molecular generation benchmark MOSES~\citep{polykovskiy2020molecular} refined from the ZINC database~\citep{sterling2015zinc}. The generated molecules from \ours{} and AO-ARM are comparable to standard state-of-the-art molecular generative models, such as CharRNN~\citep{segler2018generating}, JTN-VAE~\citep{jin2018junction}, and LatentGAN~\citep{prykhodko2019novo} (see \Cref{tab:moses-sota,tab:moses}), with added controllability and flexibility in any-order generation. \ours{} supports much much faster marginal inference, which is useful for domain scientists to easily reason about the likelihood of (sub)structures. Generated molecules and property histogram plots of are available in \Cref{sec:moses_additional_results}.

\textbf{Text}\quad
We train a character-level generative model on Text8~\citep{mahoney2011large}, which consists of 100M characters from Wikipedia in chunks of 250 character. \ours{} achieves significant speed-up in marginal inference while maintaining comparable performance as an arbitrary marginal inference model. The test NLL is close to a Transformer model that is trained to only model one ordering (from left to right).

\subsection{Energy-based Training}
In the existing literature, only ARM with fixed variable order has been used for this training setting (for example in \citet{wu2019solving, damewood2022sampling}). We additionally implement two more baselines: ARM-MC that uses single-step conditional as a Monte Carlo estimate to $\log p_\phi$ and GFlowNet~\citep{malkin2022trajectory}.  
The effective batch size for ARM and GFlowNet is $B \times D$ for batch of $B$ data samples (due to reasons mentioned in \Cref{sec:4_arm_limits}), and $B \times 1$ for ARM-MC and \ours{} .
\ours{} and ARM use the REINFORCE gradient estimator with baseline. GFlowNet is trained on per-sample gradient of squared distance~\citep{zhang2022generative}. Note that \ours{} is an any-order generative model, which is a more difficult learning task than ARM that uses fixed ordering and GFlowNet that uses learned ordering.

\begin{table}[th!]
  \centering
  \caption{Energy-based modeling of Ising model ($D=100$)}
    \small
  \label{tab:ising-100D}
  \begin{tabular}{lccc}
    \toprule
      & {NLL (bpd) $\downarrow$} & {KL div. $\downarrow$} & {Time (s) $\downarrow$}\\
    \midrule
    ARM-One-Order$^\halfam$ \citep{damewood2022sampling} & 0.79 & {-78.63} & 5.3$\pm$0.1e-01 \\
    ARM-MC-One-Order$^\halfam$ & 24.84 & -18.01 & 5.3$\pm$0.1e-01 \\
    GFlowNet \citep{zhang2022generative} & {0.78} & -78.17 & -- \\
    \rowcolor{shadecolor} \ours{}-Any-Order$^\am$ & 0.80 & -77.77 & {3.7$\pm$0.1e-04} \\
    \bottomrule
  \end{tabular}
\end{table}

\begin{table}[th!]
  \centering
  \caption{Energy-based modeling of Ising model ($D=900$)}
    \small
  \label{tab:ising-900D}
  \begin{tabular}{lccc}
    \toprule
      & {NLL (bpd) $\downarrow$} & {KL div. $\downarrow$} & {Time (s) $\downarrow$}\\
    \midrule
    ARM-One-Order$^\halfam$ \citep{damewood2022sampling} & 
    \multicolumn{3}{c}{-- Out of GPU memory --} \\
    Random Samples & {1.00} & -623.9 & -- \\
    \rowcolor{shadecolor} \ours{}-Any-Order$^\am$ & 0.83 & -685.8 & {3.7$\pm$0.1e-04} \\
    \bottomrule
  \end{tabular}
\end{table}

\begin{figure}[ht]
  \centering
  \begin{subfigure}{0.49\linewidth}
    \centering
    \includegraphics[width=\linewidth]{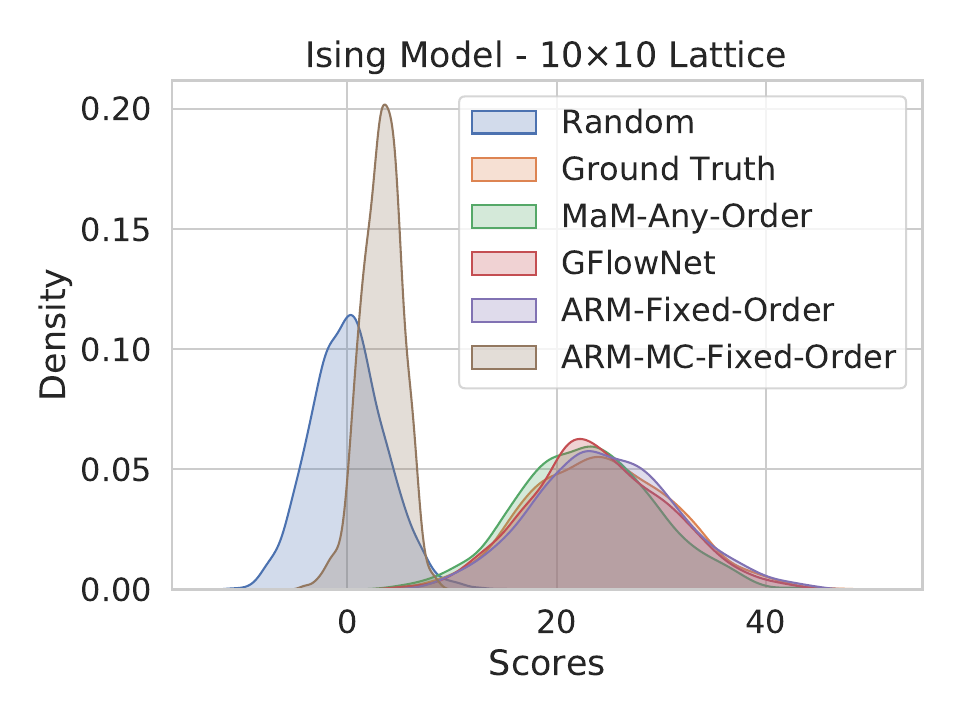}
  \end{subfigure}
  \hfill
  \begin{subfigure}{0.49\linewidth}
    \centering
    \includegraphics[width=\linewidth]{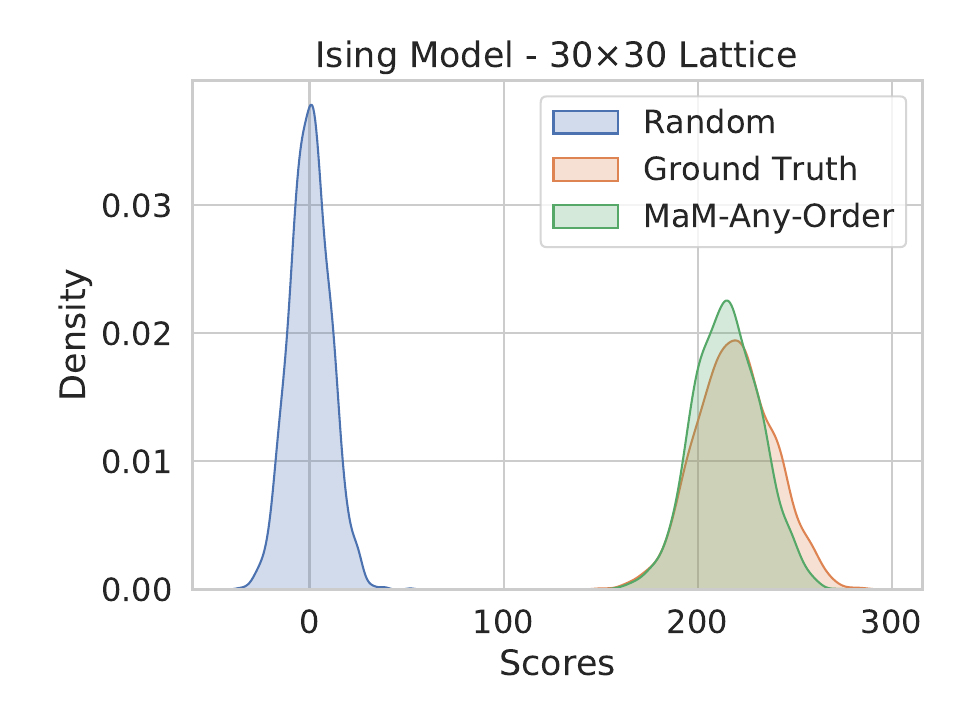}
  \end{subfigure}
  \caption{\small Ising model. Left: $D = 10 \times 10$. Right: $D = 30 \times 30$}
  \label{fig:ising}
\end{figure}

\textbf{Physical systems}\quad 
Ising models~\citep{ising1925beitrag} model interacting spins on a square lattice and are widely studied in mathematics and physics (see~\citet{mackay2003information}). The spins of the $D$ sites are represented by a $D$-dimensional binary vector $\boldx$, whose distribution ${p^*(\mathbf{x}) \propto \exp \left(-E_\boldJ(\mathbf{x})\right)}$ is determined by the energy function ${{E}_{\boldJ}(\mathbf{x}) \triangleq-\mathbf{x}^{\top} \boldJ \mathbf{x}-\boldsymbol{\theta}^{\top} \mathbf{x}}$, with $\boldJ$ being the adjacency matrix. 
We compare \ours{} with ARM, ARM-MC, and GFlowNet on a ${10 \!\times\! 10}$ ($D \!=\! 100$) and a larger ${30 \!\times\! 30}$ ($D\! =\! 900$) Ising model where ARMs and GFlowNets fail to scale. 
We generate 2000 ground truth samples following~\citet{grathwohl2021oops} and we measure test negative log-likelihood on those samples.
We also measure $D_\text{KL}(p_{\theta}(\boldx)||p^*)$ by sampling from the learned model and evaluating ${\sum_{i=1}^M(\log p_\theta(\boldx_i) \!-\! \log f^*(\boldx_i))}$. \Cref{fig:ising} contains KDE plots of $-E_{\boldJ}(\mathbf{x})$ for the generated samples. 
We validate the analysis in \Cref{sec:4_arm_limits}, the ARM-MC gradient has high variance which leads to non-convergence or mode collapse. \ours{} achieves significant speedup in marginal inference and is the only model that supports any-order generative modeling. The performance in terms of KL divergence and likelihood are only slightly worse than models with fixed/learned order, which is expected since any-order modeling is harder than fixed-order modeling, and \ours{} is solving a more complicated task of jointly learning conditionals and marginals. On a $30 \times 30$ ($D=900$) Ising model, \ours{} achieves a bpd of $0.835$ while ARM and GFlowNet fail to fit in the GPU memory (see \Cref{fig:ising} and \Cref{tab:ising-900D}).

\begin{figure}[t]
  \centering
  \begin{subfigure}{0.49\linewidth}
    \centering
    \includegraphics[width=\linewidth]{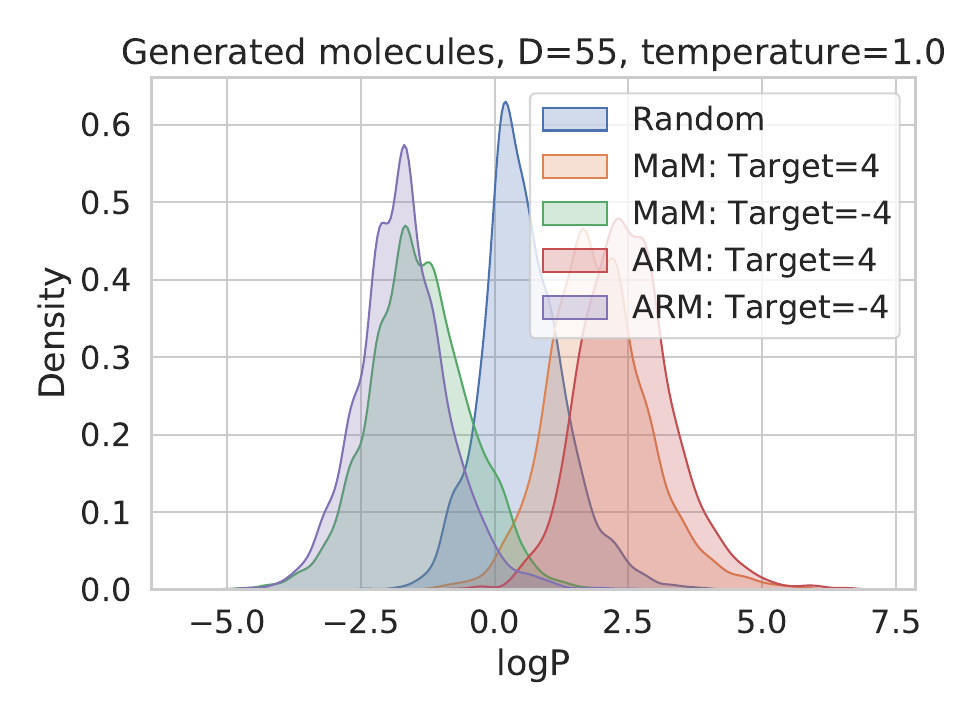}
  \end{subfigure}
  \hfill
  \begin{subfigure}{0.49\linewidth}
    \centering
    \includegraphics[width=\linewidth]{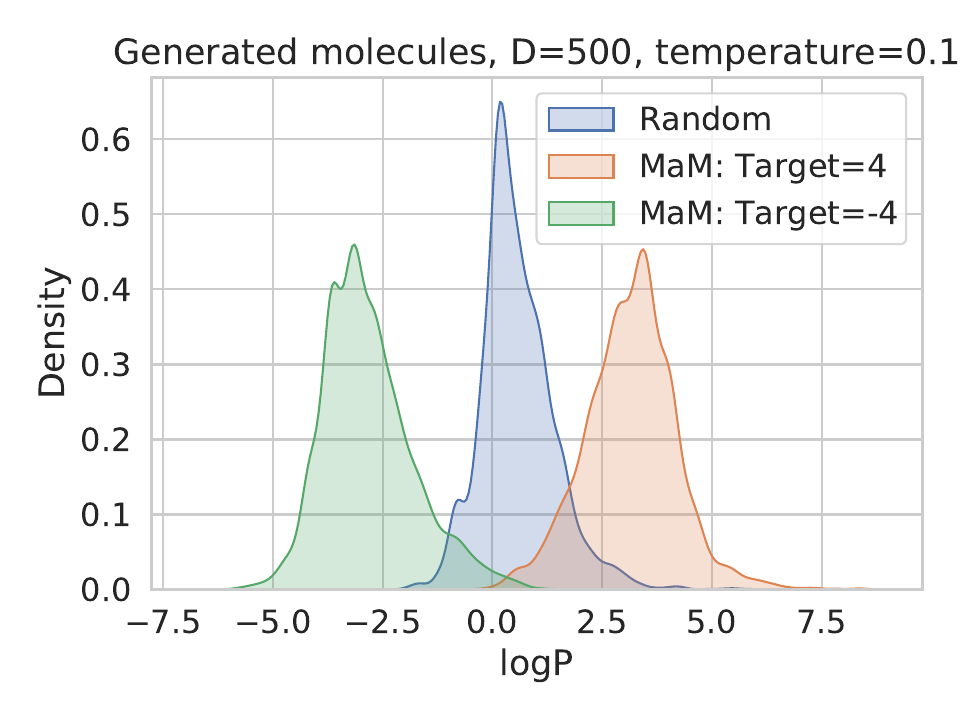}
  \end{subfigure}
  \caption{\small EB molecule targeted generation. Left: $55$d. Right: $500$d}
  \label{fig:logp}
\end{figure}

\textbf{Molecular generation with target property}\quad
In this task, we are interested in training generative models towards a specific target property of interest $g(x)$, such as lipophilicity (logP), synthetic accessibility (SA), etc. We define the distribution of molecules to follow ${p^*(x) \propto \exp (-(g(x) - g^*)^2/\tau)}$, where $g^*$ is the target value of the property and $\tau$ is a temperature parameter. We train ARM and \ours{} for lipophilicity of target values $4.0$ and $-4.0$, both with $\tau=1.0$ and $\tau=0.1$. Both models are trained for $4000$ iterations with batch size $512$. Results are shown in \Cref{fig:logp} (additional results in \Cref{sec:app_exp}). Findings are consistent with the Ising model experiments. Again, \ours{} performs just marginally below ARM. However, only \ours{} supports any-order modeling and scales to high-dimensional problems. \Cref{fig:logp} (right) shows molecular generation with \ours{} for $D = 500$.

\begin{figure}[ht!]
  \centering
  \includegraphics[width=1.0\linewidth]{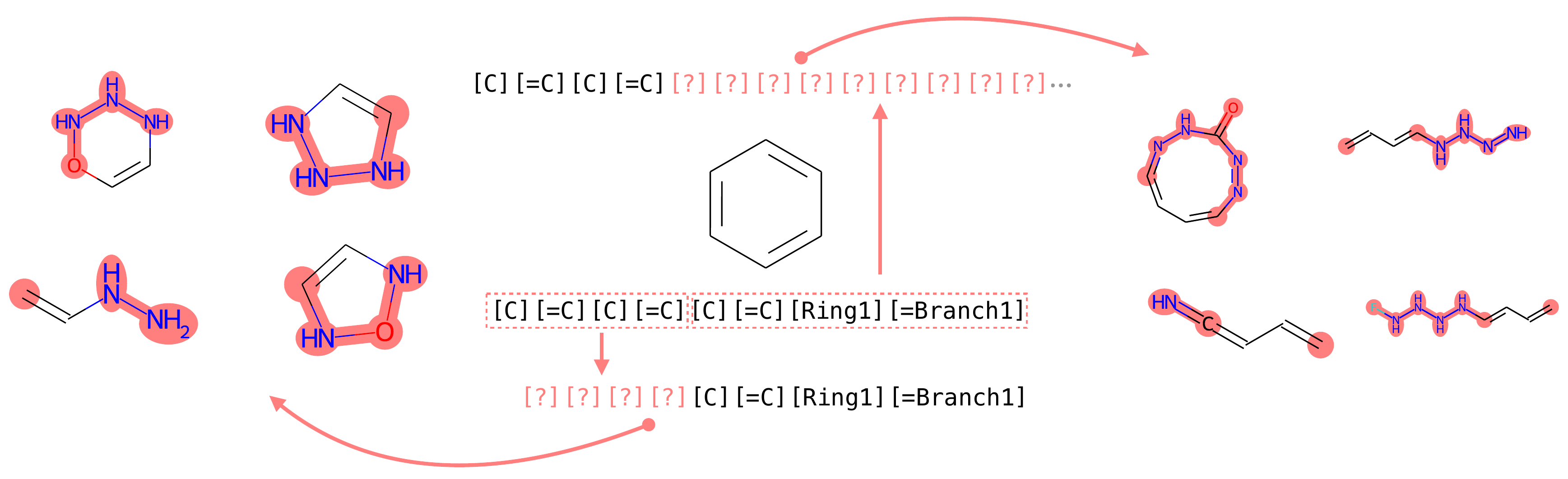}
  \caption{\small Illustration of conditional design of molecules towards low lipophilicity from a user-defined substructure in a given order. Left: Impainting the left 4 characters. Right: Impainting the right 4-20 characters. Shaded regions denote the impainted structures.}
  \label{fig:dm_mol_gen}
\end{figure}

\subsection{Comparison with Parallel AO-ARMs}

\begin{figure}[t]
  \centering
  \begin{subfigure}{0.49\linewidth}
    \centering
    \includegraphics[width=\linewidth]{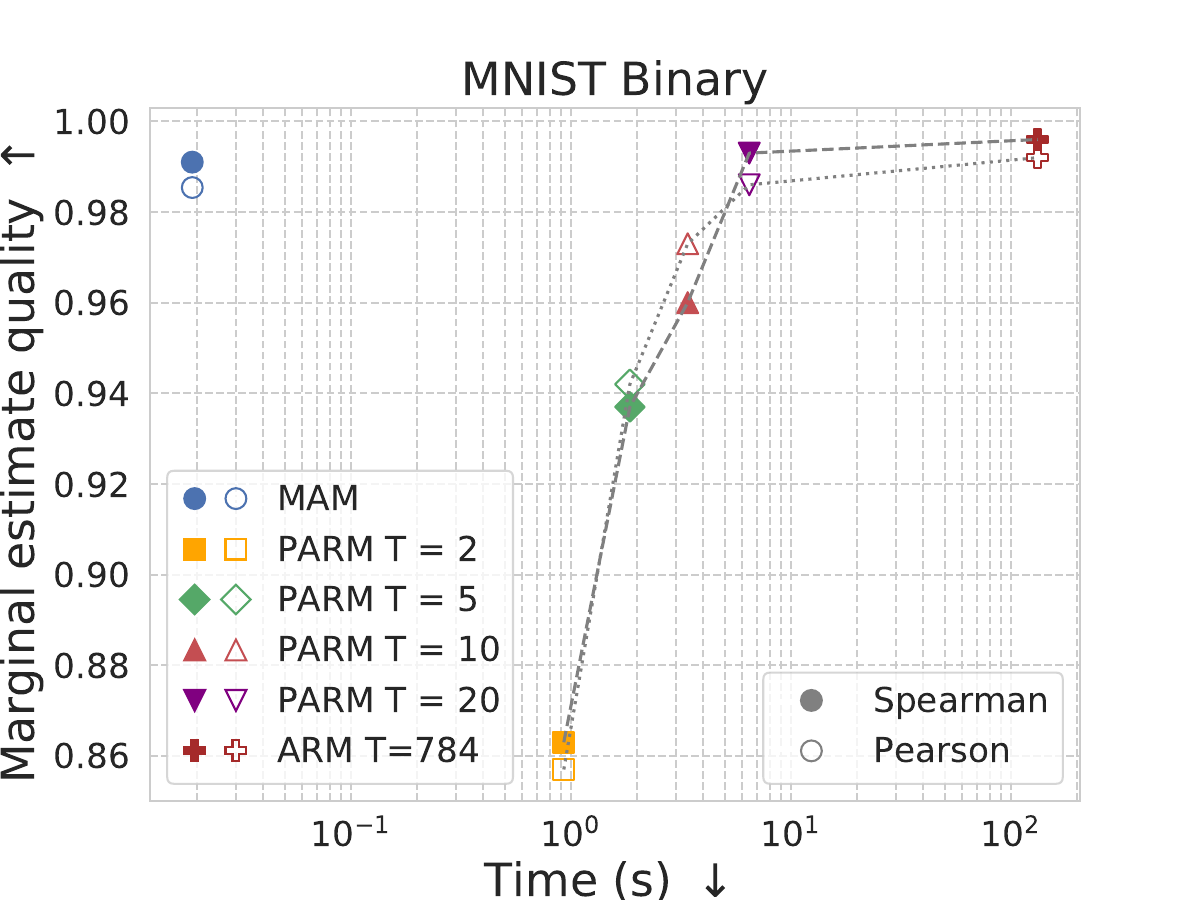}
  \end{subfigure}
  \hfill
  \begin{subfigure}{0.49\linewidth}
    \centering
    \includegraphics[width=\linewidth]{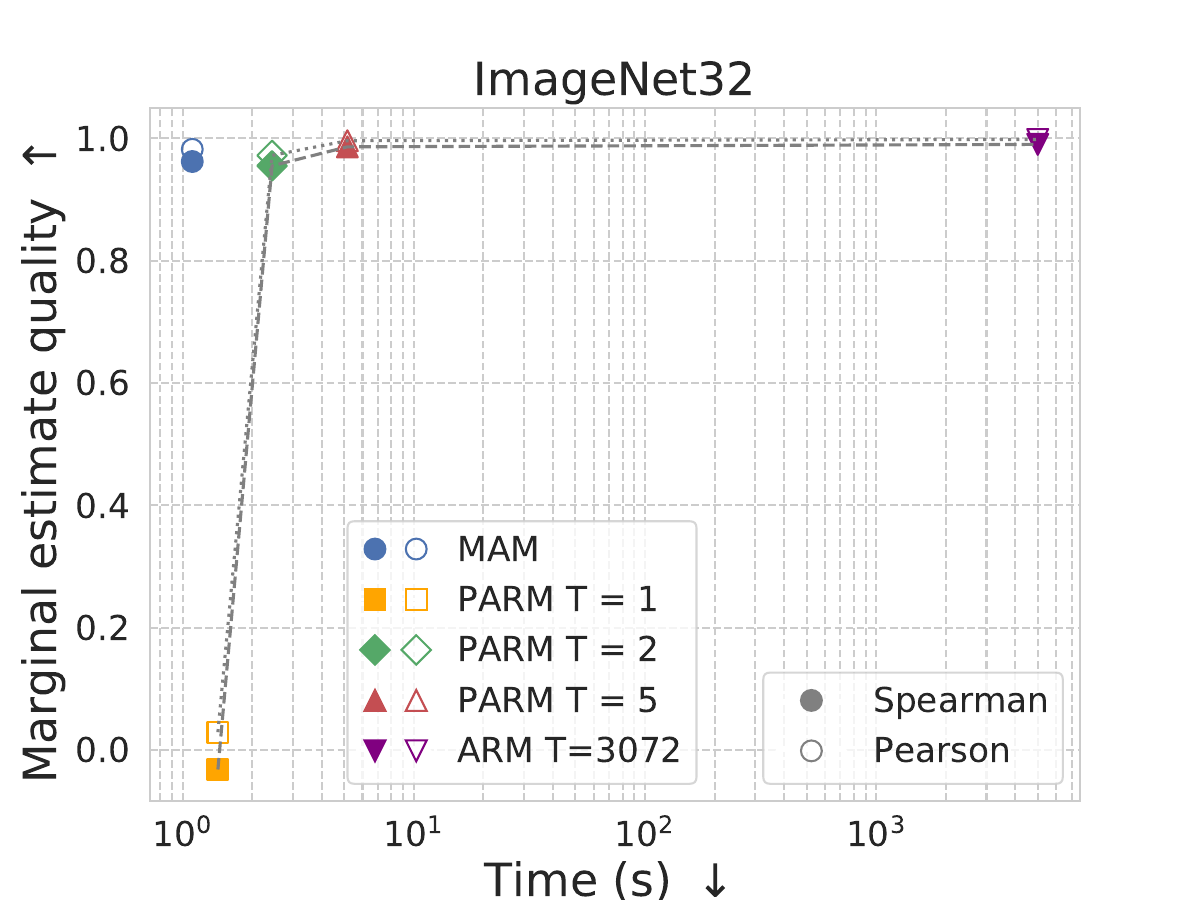}
  \end{subfigure} \\
    \begin{subfigure}{0.49\linewidth}
    \centering
    \includegraphics[width=\linewidth]{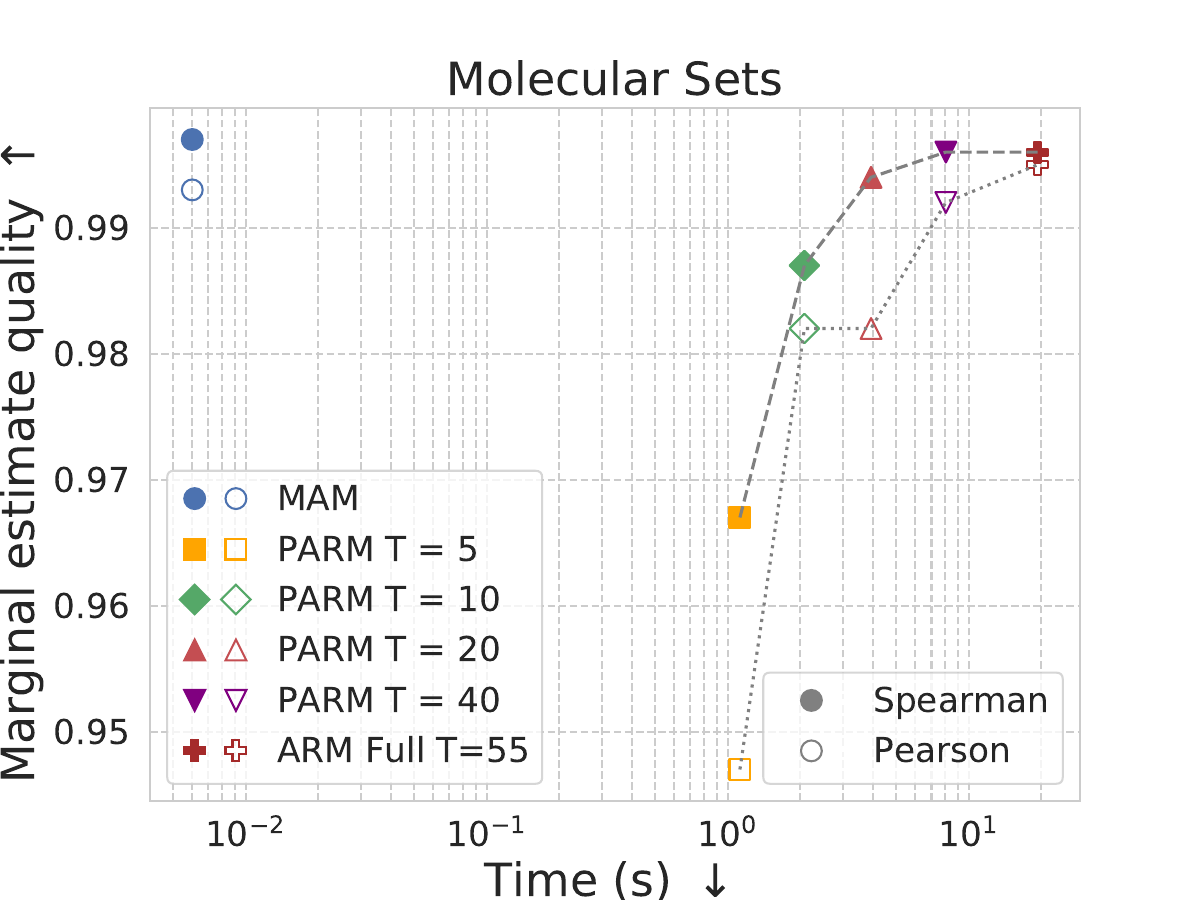}
    \end{subfigure}
    \hfill
    \begin{subfigure}{0.49\linewidth}
    \centering
    \includegraphics[width=\linewidth]{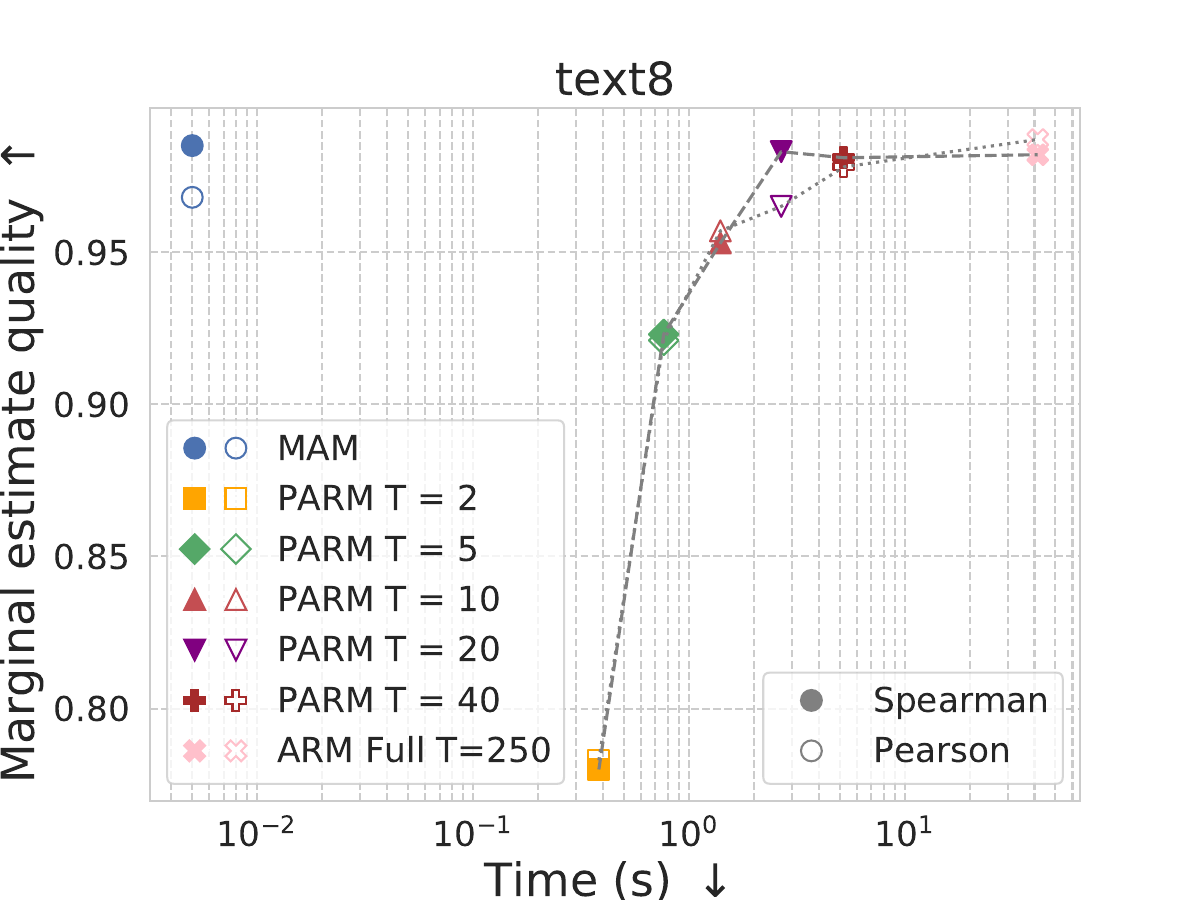}
    \end{subfigure}
  \caption{\small Comparison with Parallel AO-ARMs.}
  \label{fig:parm}
\end{figure}

Inference of AO-ARMs can be parallelized using fewer steps with dynamic programming at cost of minimal log-likelihood degradation, which makes it a strong baseline for accelerated inference as shown in \citet{hoogeboom2021autoregressive}.
In \Cref{fig:parm}, we compare the quality of \ours{} against P-AO-ARM (PARM) with varying number of sampling steps $T$. \ours{} is consistently faster and produces better-correlated marginal estimates than PARM.
PARM's effectiveness varies across datasets. Text and molecule data require more steps of PARM for accurate estimation due to their sequential dependencies. Interestingly, ImageNet32 needs much fewer PARM steps for correlated log-likelihoods (despite values being quite off), suggesting easier parallelization of sampling/inference once some pixels are filled.

\subsection{Out-of-distribution Robustness}
The marginal estimates from \ours{} are not perfectly-normalized but only approximate log-likelihood values. Hence we test the how useful and robust those approximate marginals are in real-world use cases which are often out-of-distribution with various degrees.

We tested \ours{}'s marginal estimates on generated ``synthetic'' data from masked MNIST (\Cref{subsec:mnist-app}) and Text8 examples (\Cref{subsec:text8-samples}). \ours{} $\log p$ estimates maintain a high correlation with actual log-likelihoods on data that are on-manifold but slightly out-of-distribution.
In \Cref{subsec:extrapolation}, we also tested the model's generalizability for length extrapolation on Text8. The model is trained on $D=250$ and tested on sequences with $D=300$ from the same dataset. \ours{}'s predicted $\log p$ marginals generalize gracefully to longer sequences. The quality matches Parallel AO-ARM with $10 \sim 20$ steps while using significantly less time ($2000\times$). 
Finally, in \Cref{sec:moses_additional_results},we tested the model on a more challenging task: using \ours{} model's marginal likelihood estimates trained on Molecular Sets (a general chemical space of drug-like compounds) to distinguish between two focused chemical spaces (tyrosine kinase inhibitors and organic photodiodes) that are not seen during training. We created ~$1000$ pairs consisting of one of each using datasets from \citet{subramanian2023automated}, controlling for other factors like SMILES length and chemical space to increase difficulty. \ours{} marginals correctly identified the drug molecule 74\% of the time (v.s. 79\% for AO-ARM), with 90\% alignment on marginal estimates with AO-ARM.

%% file: 7_conclusion.tex
\section{Conclusion}

In conclusion, marginalization models are a novel family of generative models for high-dimensional discrete data that offer scalable and flexible generative modeling. These models explicitly model all induced marginal distributions, allowing for fast evaluation of arbitrary marginal probabilities with a single neural net forward pass. \ours{}s also support scalable training objectives for any-order generative modeling, which previous methods struggle to achieve under the energy-based training context. Potential future work includes designing novel neural network architectures that automatically satisfy the marginalization self-consistency. 

%% file: appendix.tex
\onecolumn
\newpage
\appendix
\section{Additional Technical Details}
\label{sec:app_tech}

\subsection{Proof of Proposition~\ref{thm:mar_ml}}
\label{app:claim_proof}

\begin{proof}
From the single-step marginalization self-consistency in~\eqref{eq:one-step-marginal-conditional}, we have 
\begin{align*}
    \log p_\theta(\boldx) = \sum\nolimits_{d=1}^{D}  \log p_\phi\left(x_{\sigma(d)} |\boldx_{\sigma(<d)}\right),\;\forall \boldx, \sigma.
\end{align*}
Therefore we can rewrite the optimization in~\eqref{eq:mar_ml_constr} as:
\begin{align}
    \max_{\phi} \quad & \; \bbE_{\boldx \sim p_{\text{data}}} \bbE_{\sigma \sim \mcU(S_D)} \sum\nolimits_{d=1}^D \log p_\phi\left(x_{\sigma(d)} \mid \boldx_{\sigma(<d)}\right)
\label{eq: ml_cond_obj_marg_const} \\
  \text{s.t.} \quad & p_\theta(\boldx_{\sigma (< d)}) p_\phi(\boldx_{\sigma (d)} | \boldx_{\sigma (< d)}) = p_\theta(\boldx_{\sigma (\leq d)}), \; \forall \sigma \in S_D, \boldx \in \{1,\cdots,K\}^D, d \in [1:D]. \nonumber
\end{align}
Let $p^*$ be the optimal probability distribution that maximizes the likelihood on training data, and from the chain rule we have: 
\begin{equation*}
    p^* = \argmax_p  \bbE_{\boldx \sim p_{\text{data}}} \log p(\boldx) = \bbE_{\boldx \sim p_{\text{data}}} \bbE_{\sigma \sim \mcU(S_D)}  \sum\nolimits_{d=1}^{D}  \log p\left(x_{\sigma(d)} |\boldx_{\sigma(<d)}\right) 
\end{equation*}
Then $p^*$ is also the optimal solution to~\eqref{eq: ml_cond_obj_marg_const} the marginalization constraints are automatically satisfied by $p^*$ since it is a valid distribution. From the universal approximation theorem~\citep{hornik1989multilayer, hornik1991approximation,cybenko1989approximation}, we can use separate neural networks to model $p_\theta$ (marginals) and $p_\phi$ (conditionals), and obtain optimal solution to~\eqref{eq: ml_cond_obj_marg_const} with $\theta^*$ and $\phi^*$ that approximates $p^*$ arbitrarily well.

Specifically, if $\theta^*$ and $\phi^*$ satisfy the following three conditions below, they are the optimal solution to~\eqref{eq: ml_cond_obj_marg_const}:
\begin{align}
    p_{\phi^*}\left(x_{\sigma(d)} \mid \boldx_{\sigma(<d)}\right) & = p^*\left(x_{\sigma(d)} \mid \boldx_{\sigma(<d)}\right),\quad \forall\; \boldx,\sigma
    \label{eq:opt_cond_1}\\
    p_{\theta^*}(\boldx_s) & =  p^*(\boldx_s) Z_{\theta^*} ,\quad \forall\; \boldx, s \subseteq \{1,\cdots,D\}
    \label{eq:opt_cond_2}\\
    p_{\theta^*}(\boldx_{\sigma (< d)}) p_{\phi^*}(\boldx_{\sigma (d)} | \boldx_{\sigma (< d)}) & = p_{\theta^*}(\boldx_{\sigma (\leq d)}), \; \forall \sigma \in S_D, \boldx \in \{1,\cdots,K\}^D, d \in [1:D]
    \label{eq:opt_cond_3}
\end{align}
where $Z_{\theta^*}$ is the normalization constant of $p_{\theta^*}$ and is equal to $p_{\theta^*}\left(\left(\sv, \cdots,\sv\right)\right)$. It is easy to see from the definition of conditional probabilities that satisfying any two of the optimal conditions leads to the third one.

To obtain the optimal $\phi^*$, it suffices to solve the following optimization problem:
\begin{equation*}
    \textbf{Stage 1:} \quad 
    \max_{\phi} \; \bbE_{\boldx \sim p_{\text{data}}} \bbE_{\sigma \sim \mcU(S_D)} \sum\nolimits_{d=1}^D \log p_\phi\left(x_{\sigma(d)} \mid \boldx_{\sigma(<d)}\right)
    \nonumber
\end{equation*}
because $p^* = \argmax_p  \bbE_{\boldx \sim p_{\text{data}}} \bbE_{\sigma \sim \mcU(S_D)} \sum\nolimits_{d=1}^D  \log p^*\left(x_{\sigma(d)} |\boldx_{\sigma(<d)}\right)$ due to chain rule. Solving Stage 1 is equivalent to finding $\phi^*$ that satisfies condition~\eqref{eq:opt_cond_1}.
Then we can obtain the optimal $\theta^*$ by solving for condition~\eqref{eq:opt_cond_3} given the optimal conditionals $\phi^*$:
\begin{equation*}
    \textbf{Stage 2:} \quad \min_{\theta} \; \bbE_{\boldx \sim q(\boldx)} \bbE_{\sigma \sim \mcU(S_D)} \bbE_{d \sim \mcU(1,\cdots,D)} \left( \log [p_\theta(\boldx_{\sigma (< d)}) p_{\phi^*}(\boldx_{\sigma (d)} | \boldx_{\sigma (< d)})] - \log p_\theta(\boldx_{\sigma (\leq d)}) \right)^2
\end{equation*}
\end{proof}

\subsection{Expected Lower bound of Log-Likelihood}

Here we present the expected lower bound objective used for training AO-ARMs under maximum likelihood setting, which was first proposed by \citet{uria2014deep}. \citet{hoogeboom2021autoregressive} provided the expected lower bound perspective. 

Given an ordering $\sigma$, 
\begin{equation}
    \log p(\boldx \mid \sigma)=\sum\nolimits_{d=1}^D \log p\left(x_{\sigma(d)} \mid \boldx_{\sigma(<d)}\right).
\end{equation}

By taking the expectation over all orderings $\sigma$, we can derive a lower bound on the log-likelihood via Jensen's inequality.
\begin{align}
   \log p_\phi(\boldx) =\log \mathbb{E}_{\sigma} \, p_\phi(\boldx \mid \sigma)  &\overset{\text{Jensen's inequality}}{\geq} \bbE_{\sigma}\, \sum\nolimits_{d=1}^D \log p_\phi\left(x_{\sigma(d)} \mid \boldx_{\sigma(<d)}\right)
   \nonumber \\
    &= \mathbb{E}_{\sigma \sim \mcU(S_D)} \, 
    D \, \bbE_{d \sim \mcU(1,\ldots,D)}  \log p_\phi\left(x_{\sigma(d)} \mid \boldx_{\sigma(<d)}\right) \nonumber \\
   &= D \, \bbE_{d}\, \bbE_{\sigma} \,   \frac{1}{D-d+1} \sum\nolimits_{j \in \sigma(\geq d)}^{}  \log p_\phi\left(x_j \mid \boldx_{\sigma(<d)}\right), \label{eq:oaarm}
\end{align}
where $\sigma \sim \mcU(S_D)$, $d  \sim \mcU(1,\ldots,D)$ and $\boldx_{\sigma(<d)} = \{x_{\sigma(1)}, \ldots, x_{\sigma(d-1)}\}$. $\mcU(S)$ denotes the uniform distribution over a finite set $S$ and $\sigma(d)$ denotes the $d$-th element in the ordering.

\subsection{Algorithms}
\label{app:algorithm}
We present the algorithms for training \ours{} for maximum likelihood and energy-based training settings in Algorithm~\ref{alg:mle} and Algorithm~\ref{alg:dm}.
\input{0_algorithm}

\subsection{Connections between \ours{}s and GFlowNets}
\label{app:mam_gflownet}

In this section, we identify an interesting connection between generative marginalization models and GFlowNets. The two type of models are designed with different motivations. GFlowNets are motivated by learning a policy to generate according to an energy function and \ours{}s are motivated from any-order generation through learning to perform marginalization. However, under certain conditions, there exists an interesting connection between generative marginalization models and GFlowNets. In particular, the marginalization self-consistency condition derived from the definition of marginals in \Cref{eq:marginal} has an equivalence to the ``detailed balance'' constraint in GFlowNet under the following specific conditions.

\begin{observation}
    \label{obs:connection}
    When the directed acyclic graph (DAG) used for generation in GFlowNet is specified by the following conditions, there is an equivalence between the marginalization self-consistency condition in \Cref{eq:one-step-marginal-conditional} for \ours{} and the detailed balance constraint proposed for GFlowNet~\citep{bengio2023gflownet}. In particular, the $p_\theta(x_{\sigma(d)} | \boldx_{\sigma(<d)})$ in \ours{} is equivalent to the forward policy $P_F(\bolds_{d+1} \mid \bolds_d)$ in GFlowNet, and the marginals $p_\theta(x_{\sigma(d)})$ are equal to the flows $F(\bolds_d)$ up to a normalizing constant.
    \begin{itemize}
        
        \item DAG Condition: The DAG used for generation in GFlowNet is defined by the given tree-like structure: a sequence $\boldx$ is generated by incrementally adding one variable at each step, following a uniformly random ordering $\sigma$ i.e. $\sigma \sim \mcU(S_D)$. At step $d$, the state along the generation trajectory is defined to be $\bolds_d = \boldx_{\sigma (\leq d)}$. 
        \item Backward Policy Condition: At step $D-d$, the backward policy under the DAG is fixed by removing (un-assigning) the value of the $d+1$-th element under ordering $\sigma$ , i.e. $P_B(\bolds_{d} \mid \bolds_{d+1} ; \sigma) = \mathbbm{1}_{\{ 
        \bolds_{d} = \boldx_{\sigma (\leq d)}
        \}}$. Or equivalently, the backward policy removes (un-assigns) one of the existing variables at random, i.e. $P_B(\bolds_{d} \mid \bolds_{d+1}) = \nicefrac{1}{d+1} \mathbbm{1}_{\{  
        \bolds_{d} \subset \bolds_{d+1} 
        \}}$.
    \end{itemize}
\end{observation}

Intuitively, this is straight forward to understand, since GFlowNet generates a discrete object autoregressively. The model was proposed to enhance the flexibility of generative modeling by allowing for a learned ordering, as compared with auto-regressive models (see~\citep{zhang2022generative} Sec. 5 for a discussion). When the generation ordering is fixed, it is reduced to autoregressive models with fixed ordering, which is discussed in~\citep{zhang2022unifying}. \Cref{obs:connection} presented above for any-order ARMs can be seen as a extended result of the connection between GFlowNets and fixed-order ARMs.

We have seen the interesting connection of GFlowNets with ARMs (and \ours{}s). Next, we discuss the differences between GFlowNets and \ours{}s.
\begin{remark}
The detailed balance constraint was proposed only as a theoretical result in~\citet{bengio2023gflownet}. In actual experiments, GFlowNets are trained using either flow matching~\citep{bengio2021flow} or trajectory balance~\citep{malkin2022trajectory, zhang2022generative}. 
\end{remark}

\citet{zhang2022generative} is the most relevant GFlowNet work that targets the discrete problem setting. Training is done via minimizing the squared distance loss with trajectory balance objective. For the MLE training, it proposes to additionally learn an energy function from data so that the trajectory balance objetive can still be applied. In particular, \ours{} is different from GFlowNet in \citet{zhang2022generative} in three main aspects. 
\begin{itemize}
    \item First of all, \ours{}s target any-order generation and direct access to marginals, where as GFlowNets aim for flexibility in learning generation paths and does not offer exact likelihood or direct access to marginals under learnable generation paths. When the generation path is fixed to follow a ordering or random ordering, they are reduced to ARMs or any-order ARMs, which allow for exact likelihood. However, training with the trajectory balance objective does not offer direct access to marginals (just like how ARMs do not offer direct access to marginals but only conditionals).
    \item Second, training under MLE setting is signiticantly different: GFlowNets learn an additional learned energy function to reduce MLE training back to energy-based training, while \ours{}s directly maximizes the expected lower bound on the log-likelihood under the marginalization self-consistent constraint.
    \item Lastly, the training objective is different under energy-based training. GFlowNets are trained on squared distance under the expectation to be specified to be either on-policy, off-policy, or a mixture of both. \ours{}s are trained on KL divergence where the expectation is defined to be on-policy. It is possible though to train \ours{}s with squared distance and recently \citet{malkin2022gflownets} have shown the equivalence of the gradient of KL divergence and the on-policy expectation of the per-sample gradient of squared distance (which is the gradient actually used for training GFlowNets). 
\end{itemize}

\subsection{Additional literature on discrete generative models}
\label{app:literature}

\textbf{Discrete diffusion models}\quad
Discrete diffusion models learn to denoise from a latent base distribution into the data distribution. \citet{sohl2015deep} first proposed diffusion for binary data and was extended in~\citet{hoogeboom2021argmax} for categorical data and both works adds uniform noise in the diffusion process. A wider range of transition distributions was proposed in D3PM~\citep{austin2021structured} and insert-and-delete diffusion processes have been explored in~\citet{johnson2021beyond}.~\citet{hoogeboom2021autoregressive} explored the connection between ARMs and diffusion models with absorbing diffusion and showed that OA-ARDMs are equivalent to absorbing diffusion models in infinite time limit, but achieves better performance with a smaller number of steps.

\textbf{Discrete normalizing flow}\quad
Normalizing flows transform a latent base distribution into the data distribution by applying a sequence of invertible transformations~\citep{rippel2013high, tabak2013family, dinh2014nice,sohl2015deep, rezende2015variational, dinh2016density, kingma2016improved, papamakarios2017masked}. They have been extended to discrete data~\citep{tran2019discrete,hoogeboom2019integer} with carefully designed discrete variable transformations. Their performance is competitive on character-level text modeling, but they do not allow any-order modeling and could be limited to discrete data with small number of categories due to the use of a straight-through gradient estimators. 

\textbf{Discussion of neural generative models and Probabilistic circuits}\quad
Probabilistic circuits~\citep{choi2020probabilistic,peharz2020einsum,liu2022scaling, liu2023understanding} 
is a powerful modeling  
approach exhibiting 
fast and exact marginalization though the 
design of the model’s structure and operations. 
In contrast, neural generative models
are highly expressive, 
allowing them to
perform powerful approximate inference. %
Despite not having the exact marginalization property, the neural network approach has the advantage of 
much greater flexibility in modeling the complex distributions found in practical applications~\citep{shih2022training, hoogeboom2021autoregressive}. 
Hence, a trade-off currently exists between exact marginalization and approximate marginalization with a more expressive network. 
Our work falls 
in the neural generative models category, but directly approximates marginals. Direct modeling of marginals opens opportunities for more flexible sampling, as shown in \Cref{sec:sample_with_marginal}, and more scalable approximate marginal inference and training under EB settings.

\section{Ablation Studies}

\subsection{Testing marginal self-consistency}
\label{sec:test_msc}
The \textit{marginal self-consistency} in \ours{}s is enforced through optimizing the scalable training objective. Here we empirically examine how well they are enforced in practice. First we look at \textit{checkerboard}, a synthetic problem often used for testing clustering algorithms. More recently it has been used for testing and visualizing both continuous and discrete generative models. We define a discrete input space by discretizing the continuous coordinates of points in 2D. To be more concrete, the origin range $[-4,4]$ of each dimension is converted into a $16$-bit string following the standard way of converting float to string. The target unnormalized probability $p(\boldx)$ is set to $1$ for points within dark squares and $1e-10$ within light squares (since it is infeasible to set it to $\ln(0) = -\infty$ for a NN to learn, and in practice $1e-10$ is negligible compared to $1$). 
We trained a $5$-layer MLP with hidden node size $2048$ and residual connections on this problem on both MLE and EBM settings and $q(\boldx)$ is set to be a balanced mixture of ground truth data and samples from $p_\theta$ for MLE or uniform random for EBM:
\begin{align}
  \min_{\theta}  - \bbE_{\boldx \sim p_\text{data}} p_{\theta} \left(\boldx\right)
  + \lambda \,\bbE_{\boldx \sim q(\boldx)} \bbE_{\sigma} \bbE_{d}\! \left(  \log \textstyle\sum\nolimits_{x_{\sigma (d)}} p_\theta([\boldx_{\sigma (< d)}, x_{\sigma(d)}]) -  \log p_\theta([\boldx_{\sigma (< d)}, x_{\sigma(d)}]) \right)^2   \!\!.
  \nonumber %
\end{align}

\begin{align}
  \min_{\theta}  D_\text{KL}\!\left( p_{\theta} \left(\boldx\right)\! \parallel\! p\!\left(\boldx\right) \right) 
  + \lambda \,\bbE_{\boldx \sim q(\boldx)} \bbE_{\sigma} \bbE_{d}\! \left(  \log \textstyle\sum\nolimits_{x_{\sigma (d)}} p_\theta([\boldx_{\sigma (< d)}, x_{\sigma(d)}]) -  \log p_\theta([\boldx_{\sigma (< d)}, x_{\sigma(d)}]) \right)^2   \!\!.
  \nonumber %
\end{align}

\begin{figure}[ht!]
  \centering
  \begin{subfigure}{0.35\linewidth}
    \centering
    \includegraphics[width=\linewidth]{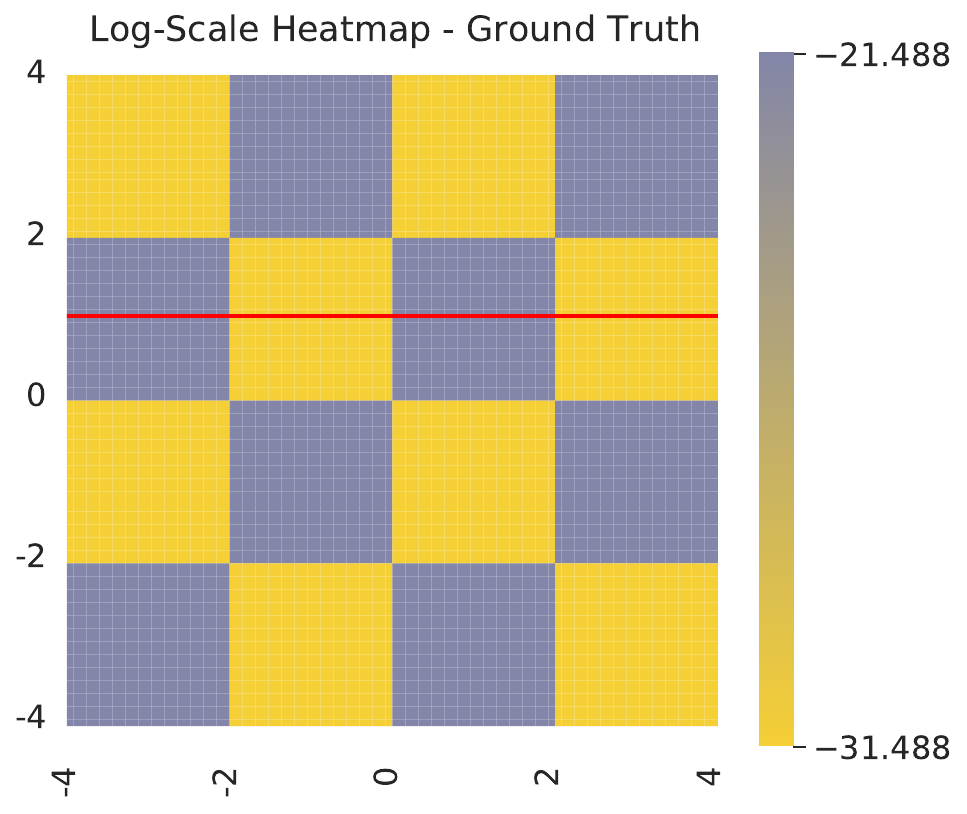}
  \end{subfigure}
  \begin{subfigure}{0.32\linewidth}
    \centering
    \includegraphics[width=\linewidth]{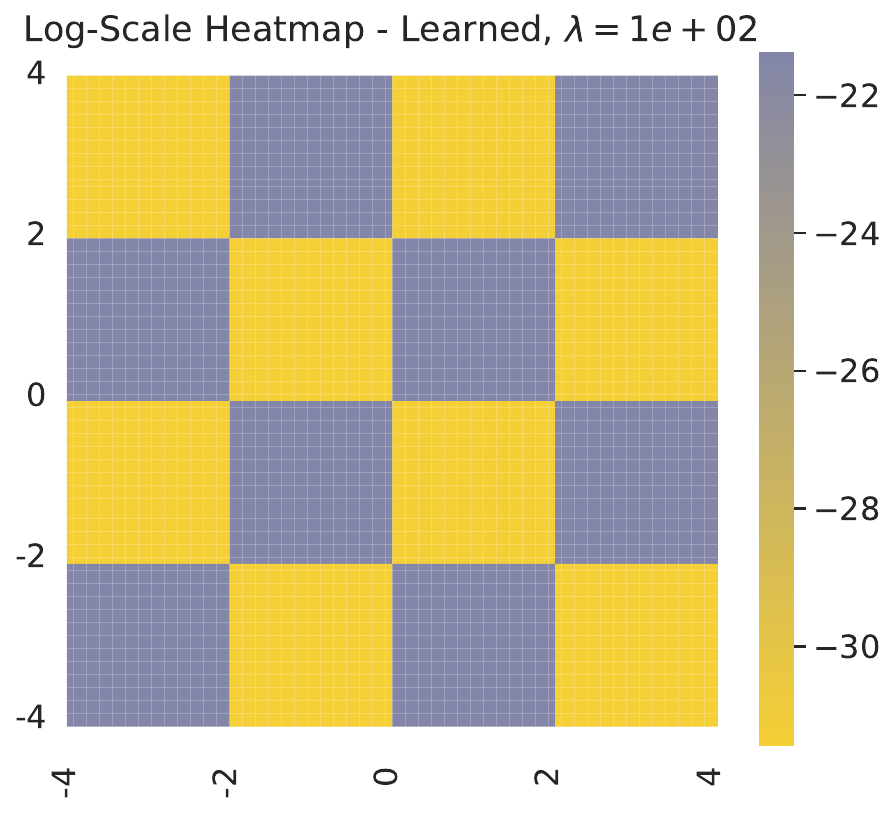}
  \end{subfigure}
  \begin{subfigure}{0.30\linewidth}
    \centering
    \includegraphics[width=\linewidth]{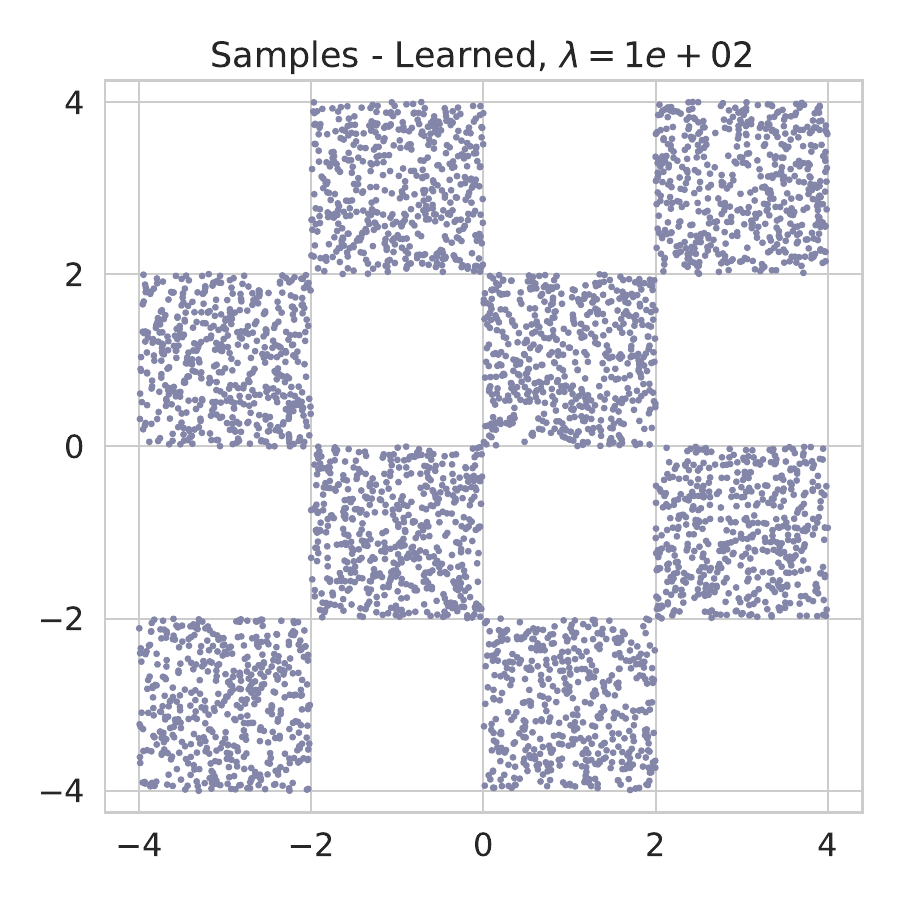}
  \end{subfigure}
  \caption{
  \small PMF heat map under EB training. The learned PMF and ground truth PMF are consistent to each other relatively well. The MSE on log $p$ (or $p$) of dark pixels is $0.0033$ (or $7.67e-20$) and the MSE on light pixels is $0.0076$ (or $3.73e-30$). We are evaluating marginals along the red line: i.e. fixing $(\boldx_{17}, \cdots, \boldx_{32}) = (0,0,0,1,1,1,\cdots,1)$, which correspond to $1$ in floating number for y-axis, and perform marginalization over $(\boldx_{1}, \cdots, \boldx_{16})$. $(0,0,\cdots)$ corresponds to $[0,2]$. $(0,1,\cdots)$ corresponds to $[2,4]$. $(1,0,\cdots)$ corresponds to $[-2,0]$. $(1,1,\cdots)$ corresponds to $[-4,-2]$. 
  }
  \label{fig:syntheitc_heat_map}
\end{figure}

\begin{figure}[ht!]
  \centering
  \begin{subfigure}{0.24\linewidth}
    \centering
    \includegraphics[width=\linewidth]{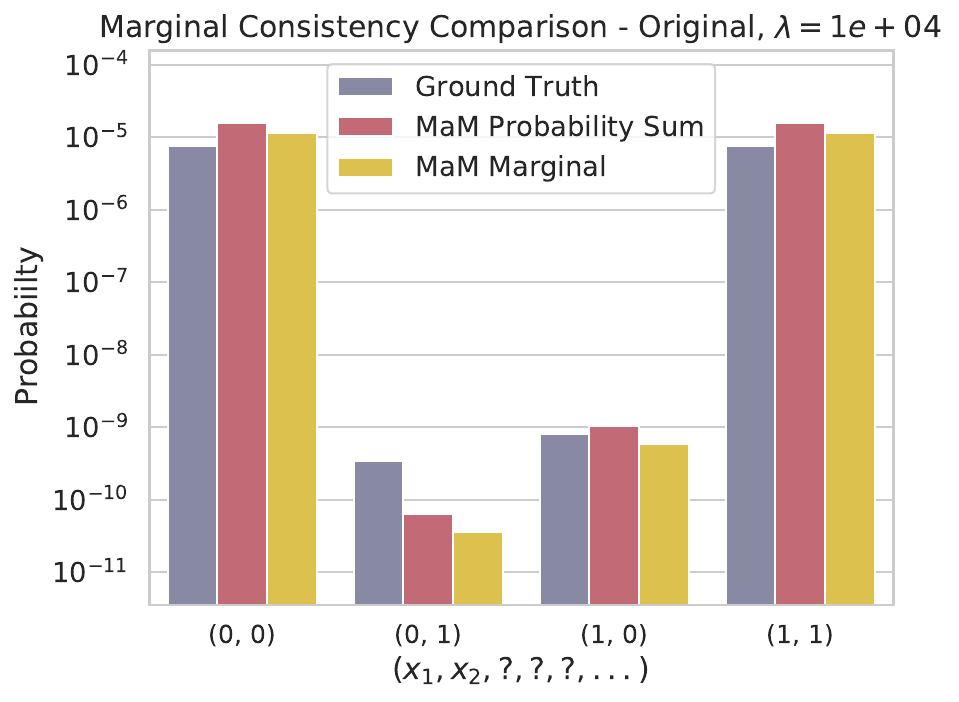}
  \end{subfigure}
  \hfill
  \begin{subfigure}{0.24\linewidth}
    \centering
    \includegraphics[width=\linewidth]{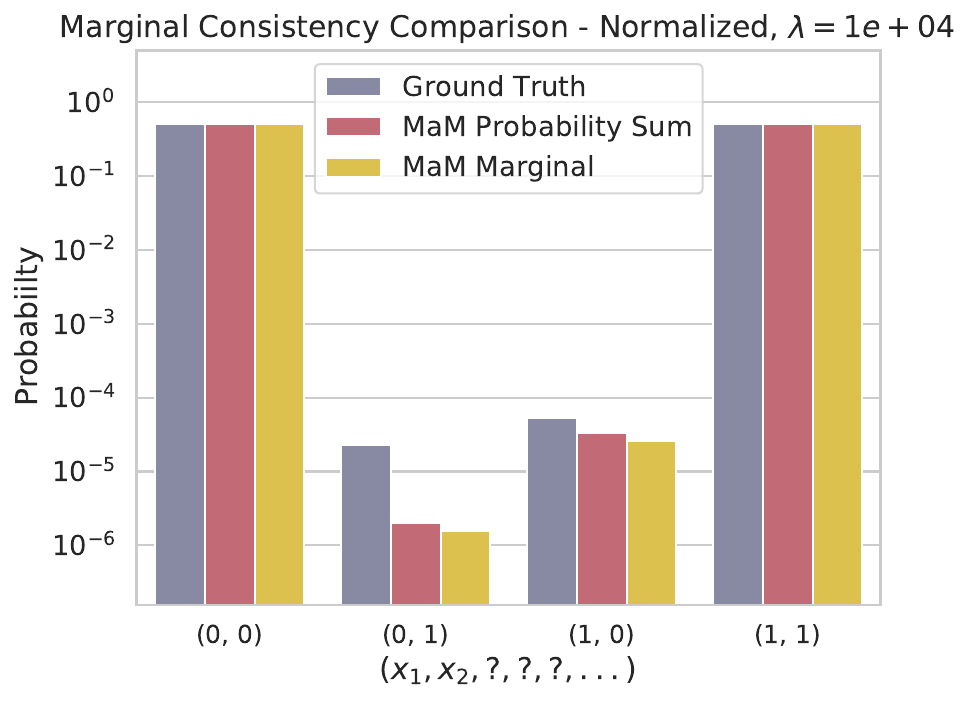}
  \end{subfigure}
  \begin{subfigure}{0.24\linewidth}
    \centering
    \includegraphics[width=\linewidth]{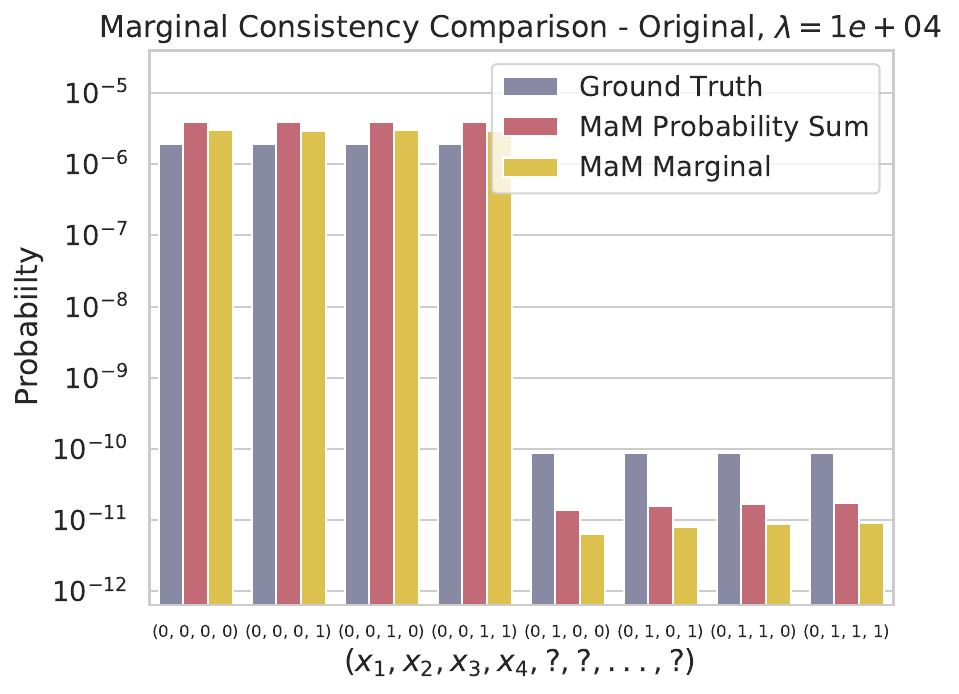}
  \end{subfigure}
  \hfill
  \begin{subfigure}{0.24\linewidth}
    \centering
    \includegraphics[width=\linewidth]{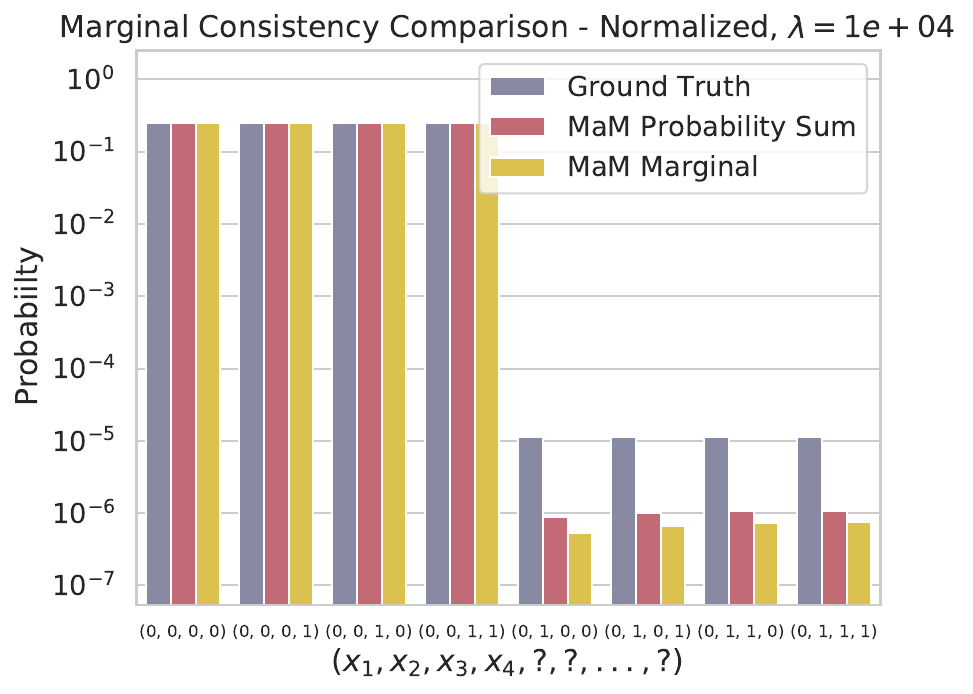}
  \end{subfigure}
  \caption{\small Marginal consistency $\lambda=1e4$ with MLE training. Ground truth: summing over ground truth PMF. \ours{} Probability Sum: summing over learned PMF from \ours{}. \ours{} Marginal: direct estimate with \ours{}. Note that $p$ for $(0,1)$ and $(1,0)$ should be in principle close to zero, but are non-zero due to float-to-int converting numerical errors.}
  \label{fig:msc_1e4}
\end{figure}

\begin{figure}[ht!]
  \centering
  \begin{subfigure}{0.24\linewidth}
    \centering
    \includegraphics[width=\linewidth]{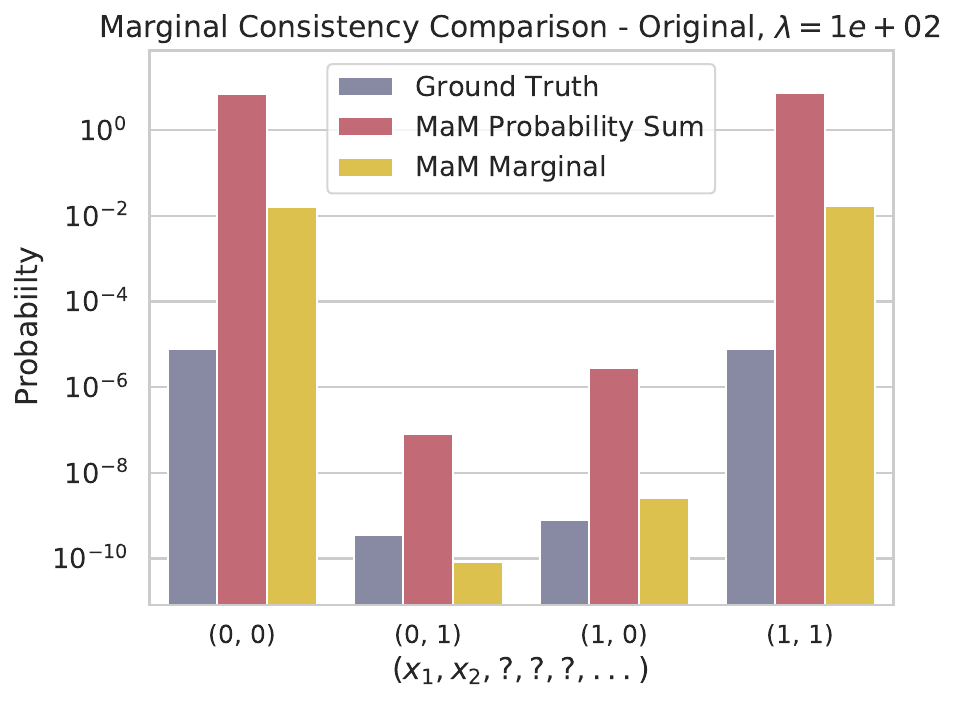}
  \end{subfigure}
  \hfill
  \begin{subfigure}{0.24\linewidth}
    \centering
    \includegraphics[width=\linewidth]{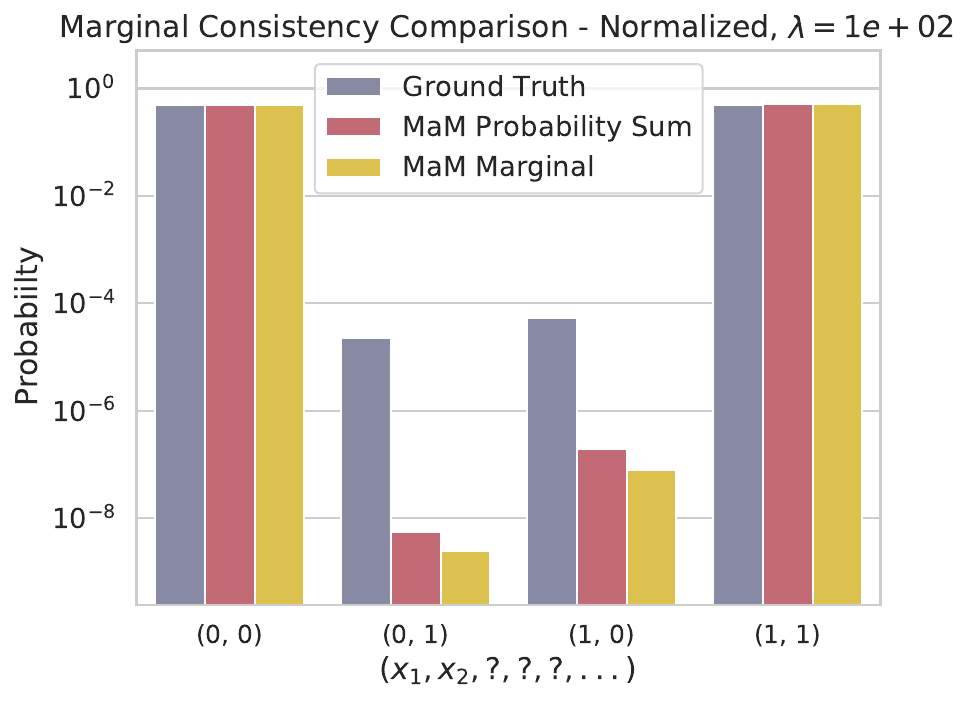}
  \end{subfigure}
  \begin{subfigure}{0.24\linewidth}
    \centering
    \includegraphics[width=\linewidth]{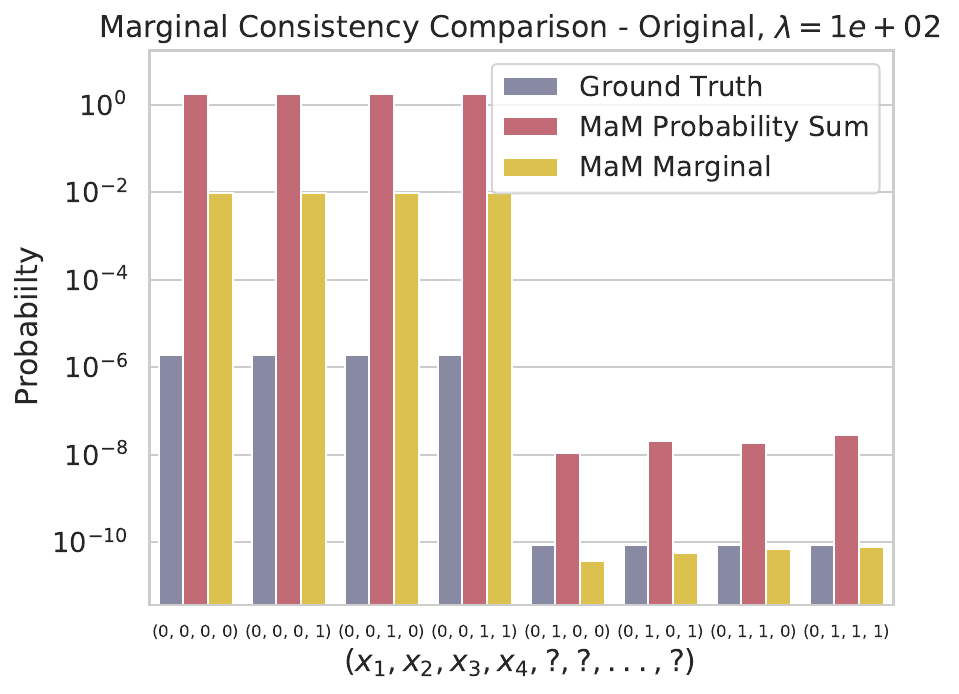}
  \end{subfigure}
  \hfill
  \begin{subfigure}{0.24\linewidth}
    \centering
    \includegraphics[width=\linewidth]{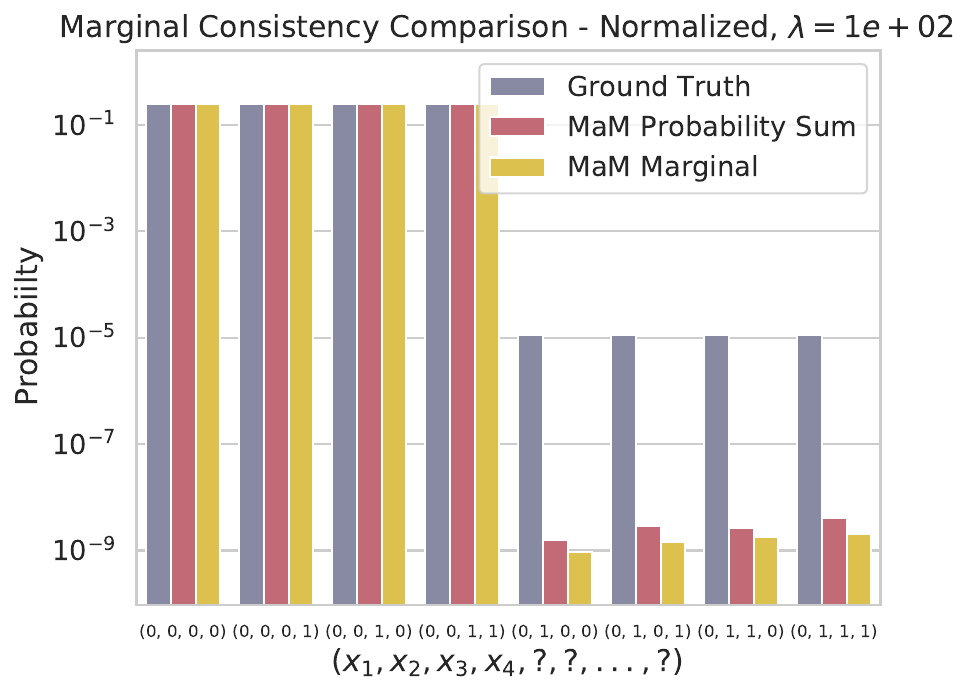}
  \end{subfigure}
  \caption{\small Marginal consistency $\lambda=1e2$ with MLE training.}
  \label{fig:msc_1e2}
\end{figure}

For this problem, only a marginal network $\theta$ is trained to predict the $\log p$ of any marginals. Upon training to convergence, the generative models perform on par or better than state of the art discrete generative models and achieve a $20.68$ test NLL. See \Cref{fig:syntheitc_heat_map} for a comparison of ground truth and learned PMF heatmap. It can be seen the PMF are approximated quite accurately. We investigate how well the marginal self-consistency are enforced, by looking at the marginal estimates of \ours{}s trained with $\lambda = 1e2$ and $\lambda = 1e4$. We evaluate marginals over the first dimension ($0-16$ bits) by fixing the second dimension ($17-32$ bits) to $1.0$ (bit string $=0001111111111111$). We plot marginals by marginalizing out bit $3-16$ (i.e. $(x_1, x_2, \cdots)$) and bit $5-16$ (i.e. $(x_1, x_2, x_3, x_4, \cdots)$). In \Cref{fig:msc_1e4}, when $\lambda=1e4$, the self-consistency are more strictly enforced, leading to matched marginals. Notice that there is some tiny residue PMF at the light squares due to the $1e-10$ approximation applied to points with $0$ probability, but they are negligible compared to the significant probability masses.
After normalizing the marginals over all possibilities, the marginals are almost exactly matched. In \Cref{fig:msc_1e2}, when $\lambda=1e2$, the self-consistency are more loosely enforced as compared to $\lambda=1e4$. But it is notable that they are only shifted by a constant as compared to the ground truth marginals. This means although marignal self-consisteny is not strictly enforced when $\lambda=1e2$, softly enforcing it leads to shifted but consistent estimates of marginals, as the NN learns to generalize and predict symmetric probabilities for symmetric regions. Using the constant-shifted marginals to sample will result in the same distribution with the ground truth, because the normalized \ours{} marginals match the ground truth almost exactly. This is observed in the samples generated under $\lambda=1e2$ in \Cref{fig:syntheitc_heat_map} and consistent normalized marginals in \Cref{fig:msc_1e2}.

\begin{figure}[ht!]
  \centering
  \begin{subfigure}{0.24\linewidth}
    \centering
    \includegraphics[width=\linewidth]{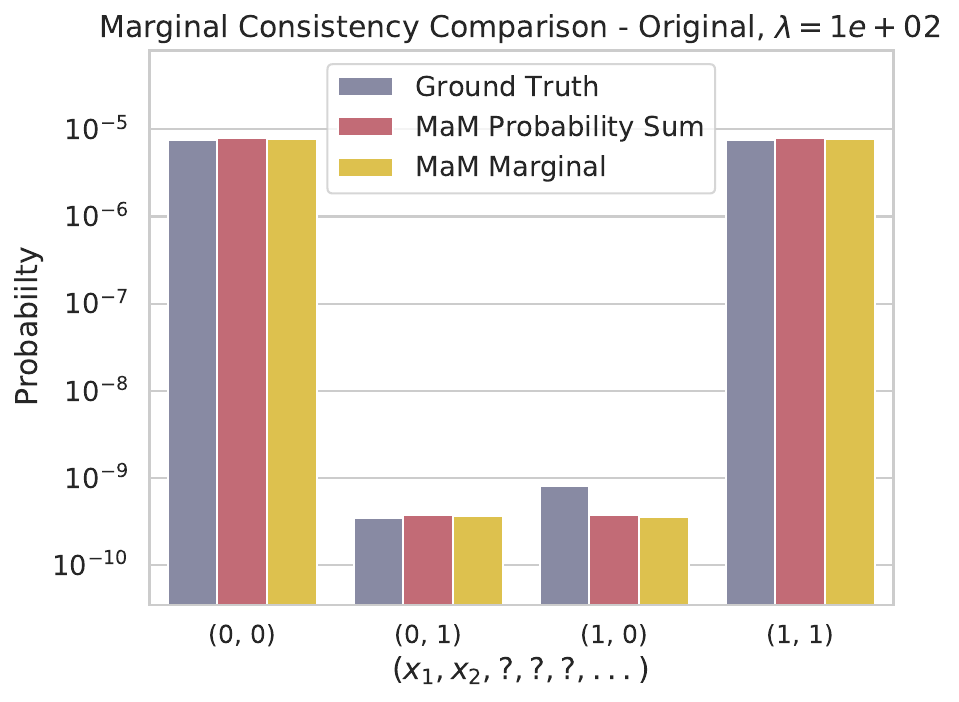}
  \end{subfigure}
  \hfill
  \begin{subfigure}{0.24\linewidth}
    \centering
    \includegraphics[width=\linewidth]{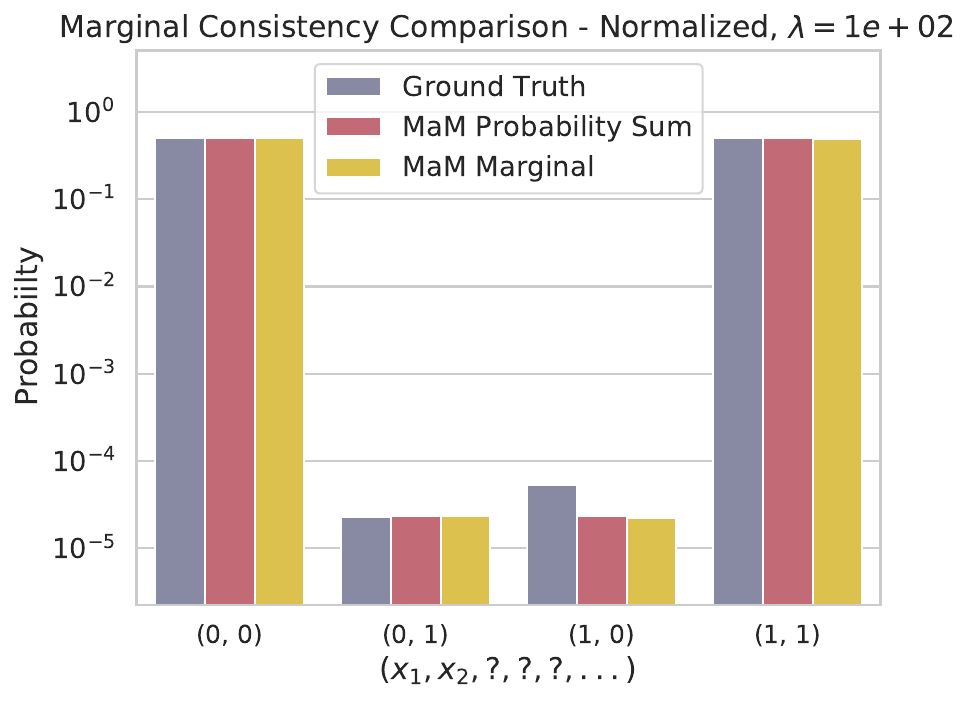}
  \end{subfigure}
  \begin{subfigure}{0.24\linewidth}
    \centering
    \includegraphics[width=\linewidth]{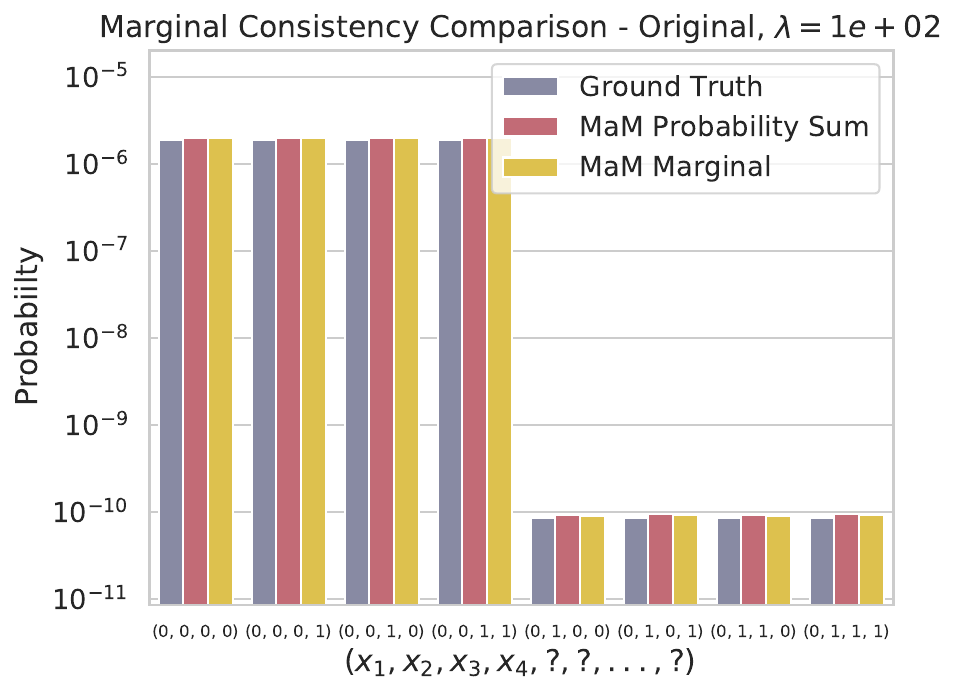}
  \end{subfigure}
  \hfill
  \begin{subfigure}{0.24\linewidth}
    \centering
    \includegraphics[width=\linewidth]{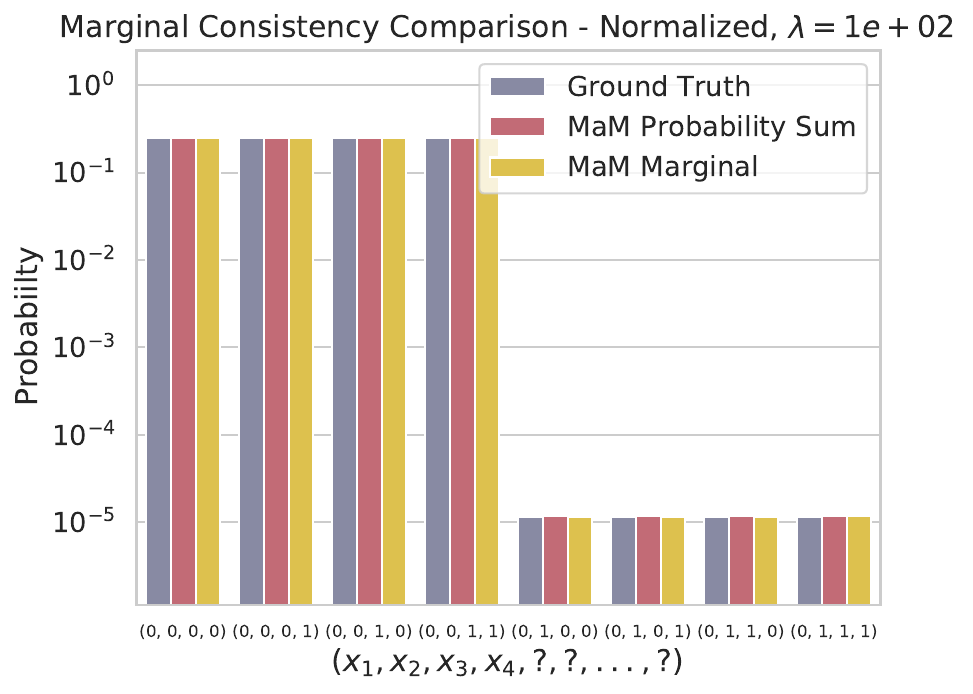}
  \end{subfigure}
  \caption{\small Marginal consistency $\lambda=1e2$ under EB training. Ground truth: summing over ground truth PMF. \ours{} Probability Sum: summing over learned PMF from \ours{}. \ours{} Marginal: direct estimate with \ours{}. The small discrepancy in $p(1, 0, ?, \cdots, ?)$ is due to the corner case of $(1, 0, 0, 0, \cdots, 0)$ be assigned to a positive value due to numerical errors in float conversion. }
  \label{fig:msc_kl_1e2}
\end{figure}

\begin{figure}[ht!]
  \centering
  \begin{subfigure}{0.205\linewidth}
    \centering
    \includegraphics[width=\linewidth]{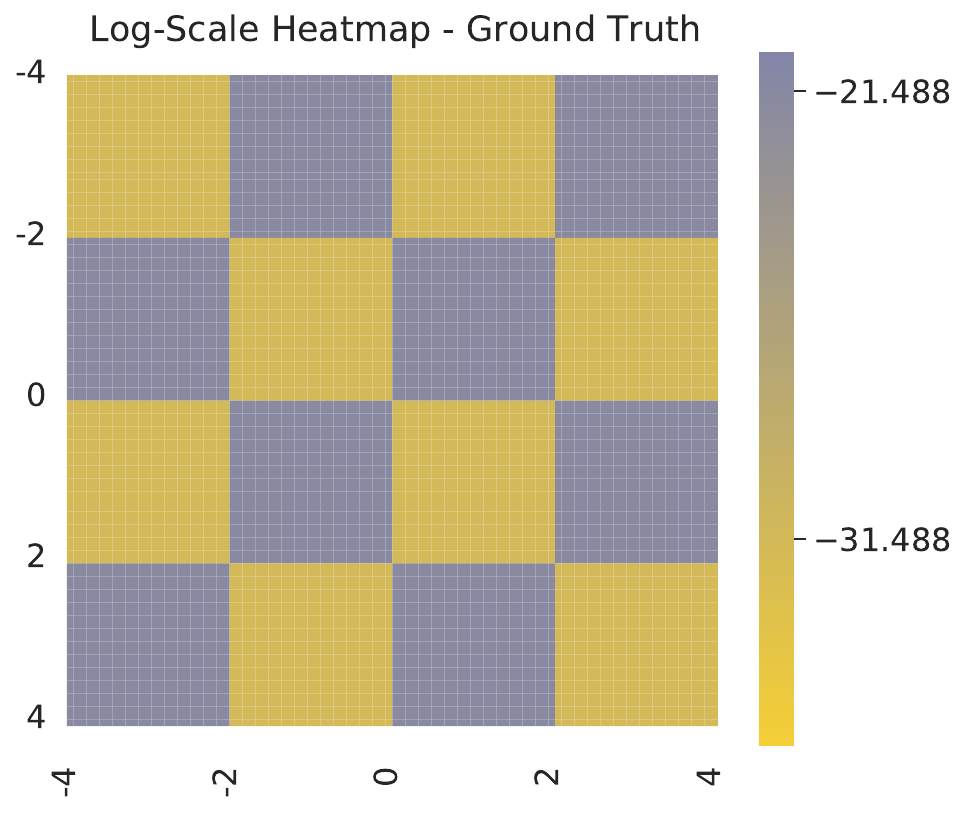}
  \end{subfigure}
  \begin{subfigure}{0.19\linewidth}
    \centering
    \includegraphics[width=\linewidth]{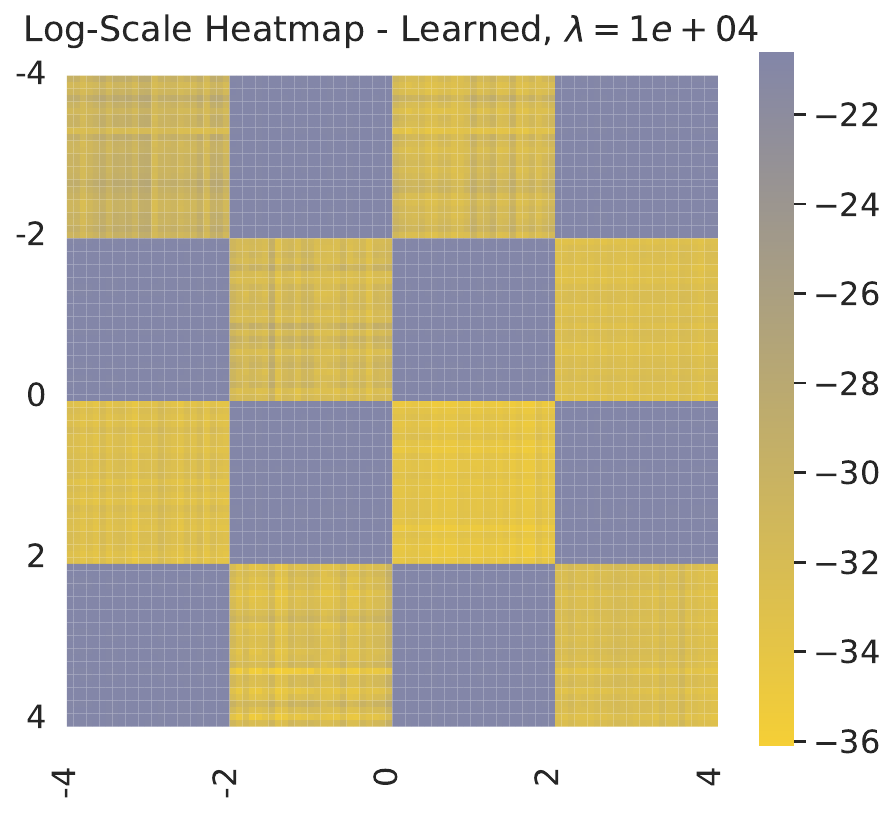}
  \end{subfigure}
  \begin{subfigure}{0.18\linewidth}
    \centering
    \includegraphics[width=\linewidth]{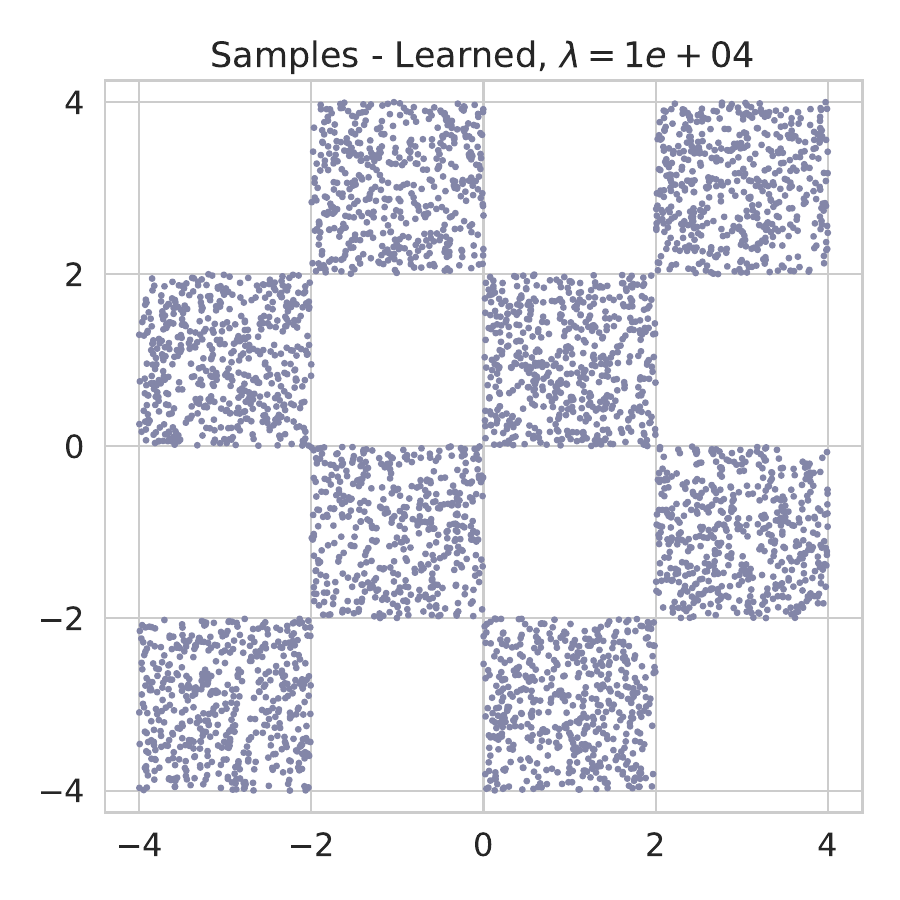}
  \end{subfigure}
  \begin{subfigure}{0.2\linewidth}
    \centering
    \includegraphics[width=\linewidth]{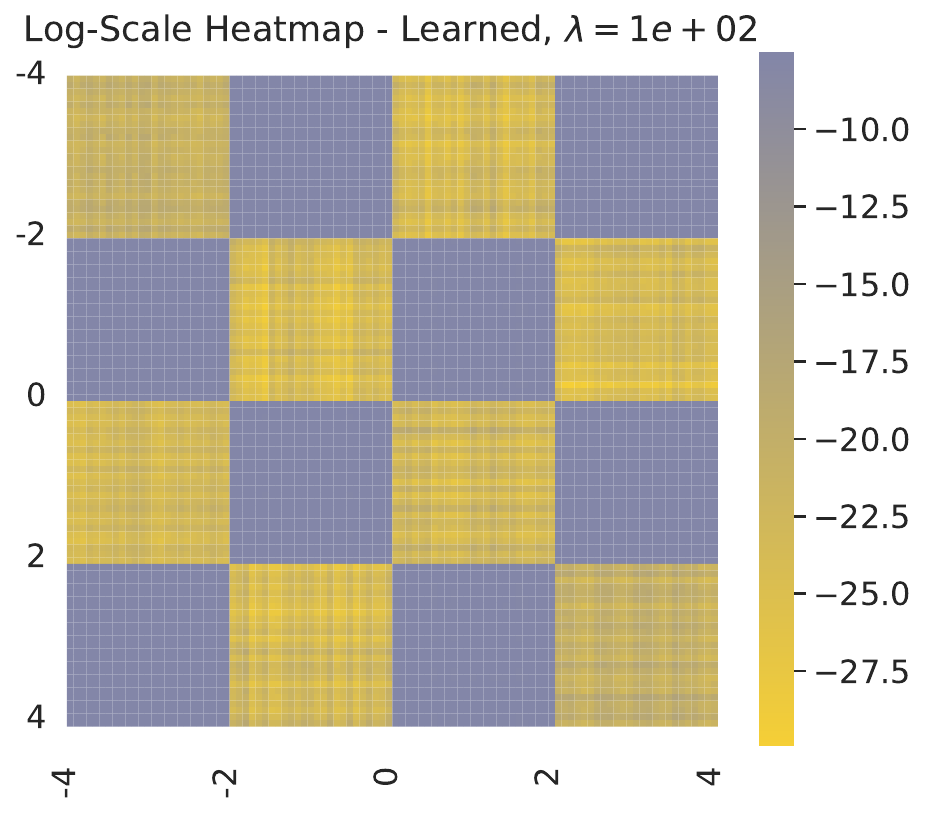}
  \end{subfigure}
  \begin{subfigure}{0.18\linewidth}
    \centering
    \includegraphics[width=\linewidth]{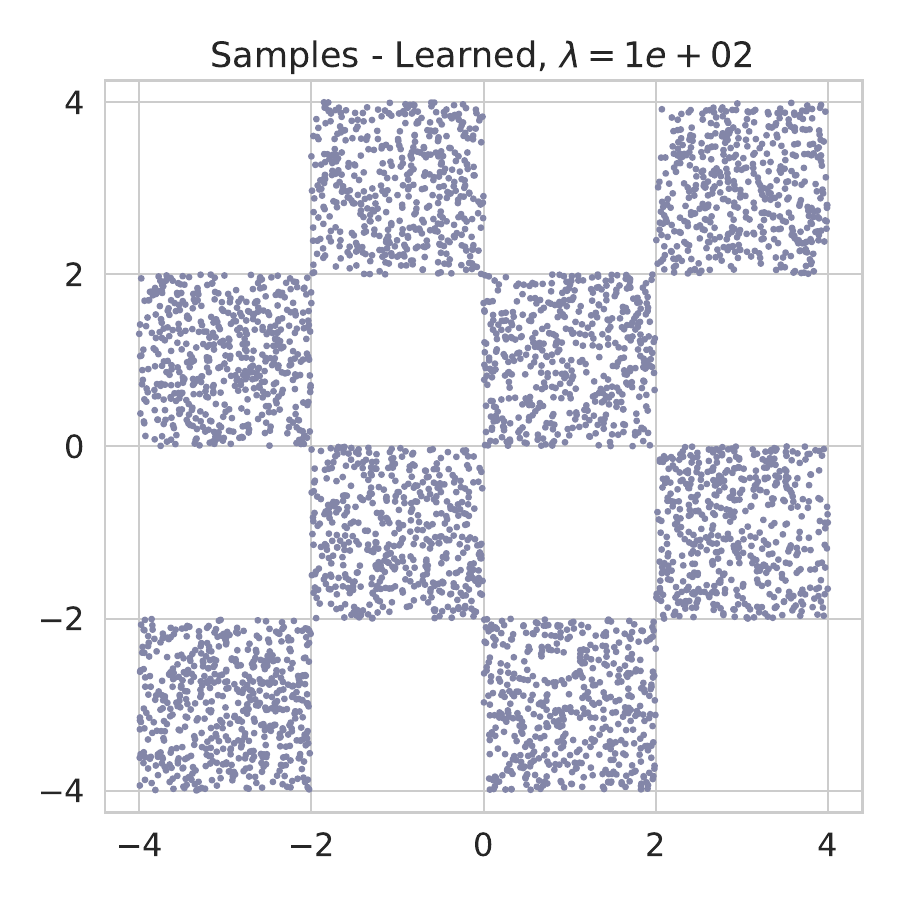}
  \end{subfigure}
  \caption{\small PMF heat map under MLE training. The learned PMF and ground truth PMF are consistent to each other relatively well. The MSE on log $p$ (or $p$) of dark pixels is $0.533$ (or $2.5e-19$) and the MSE on light pixels is $2.5$ (or $3e-28$).}
  \label{fig:syntheitc_heat_map_mle}
\end{figure}

\subsection{Sampling with marginals v.s. conditionals}
\label{sec:sample_with_marginal}
\begin{figure}[ht!]
  \centering
  \begin{subfigure}{0.245\linewidth}
    \includegraphics[width=\linewidth]{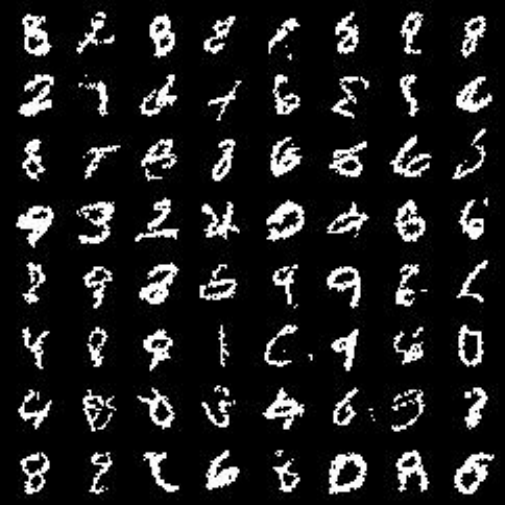}
    \caption{Marg.: $1$ pixel per step}
  \end{subfigure}
  \begin{subfigure}{0.245\linewidth}
    \includegraphics[width=\linewidth]{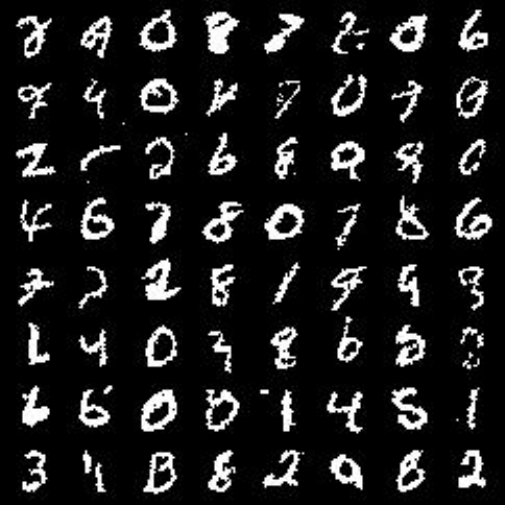}
      \caption{Marg.: $2$ pixels per step}
  \end{subfigure}
  \begin{subfigure}{0.245\linewidth}
    \includegraphics[width=\linewidth]{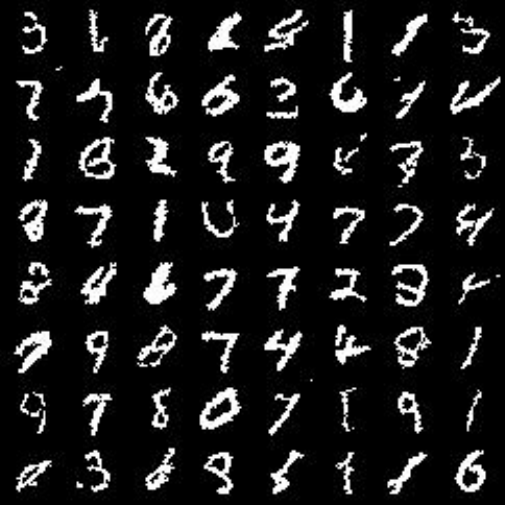}
      \caption{Marg.: $4$ pixels per step}
  \end{subfigure}
  \begin{subfigure}{0.245\linewidth}
    \includegraphics[width=\linewidth]{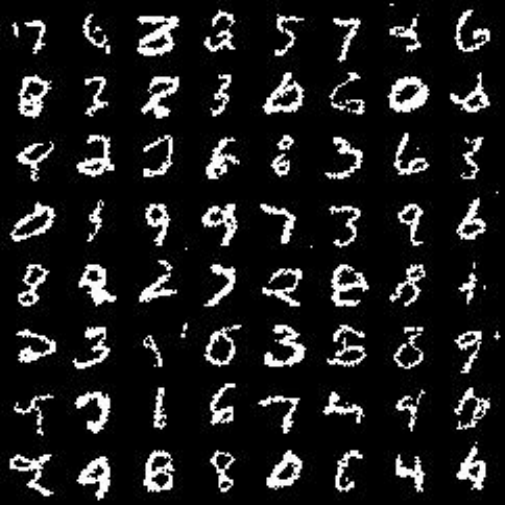}
      \caption{Marg.: $8$ pixels per step}
  \end{subfigure}\\
  \centering
  \begin{subfigure}{0.245\linewidth}
      \centering
    \includegraphics[width=\linewidth]{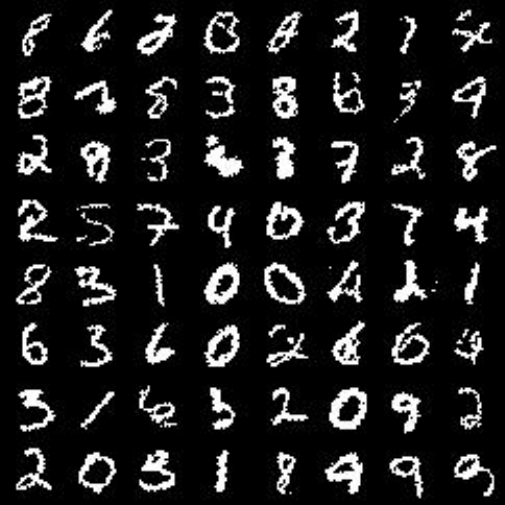}
      \caption{Cond.: $1$ pixel per step}
  \end{subfigure}
    \hspace{10mm}
    \begin{subfigure}{0.55\linewidth}
    \centering
    \includegraphics[width=\linewidth]{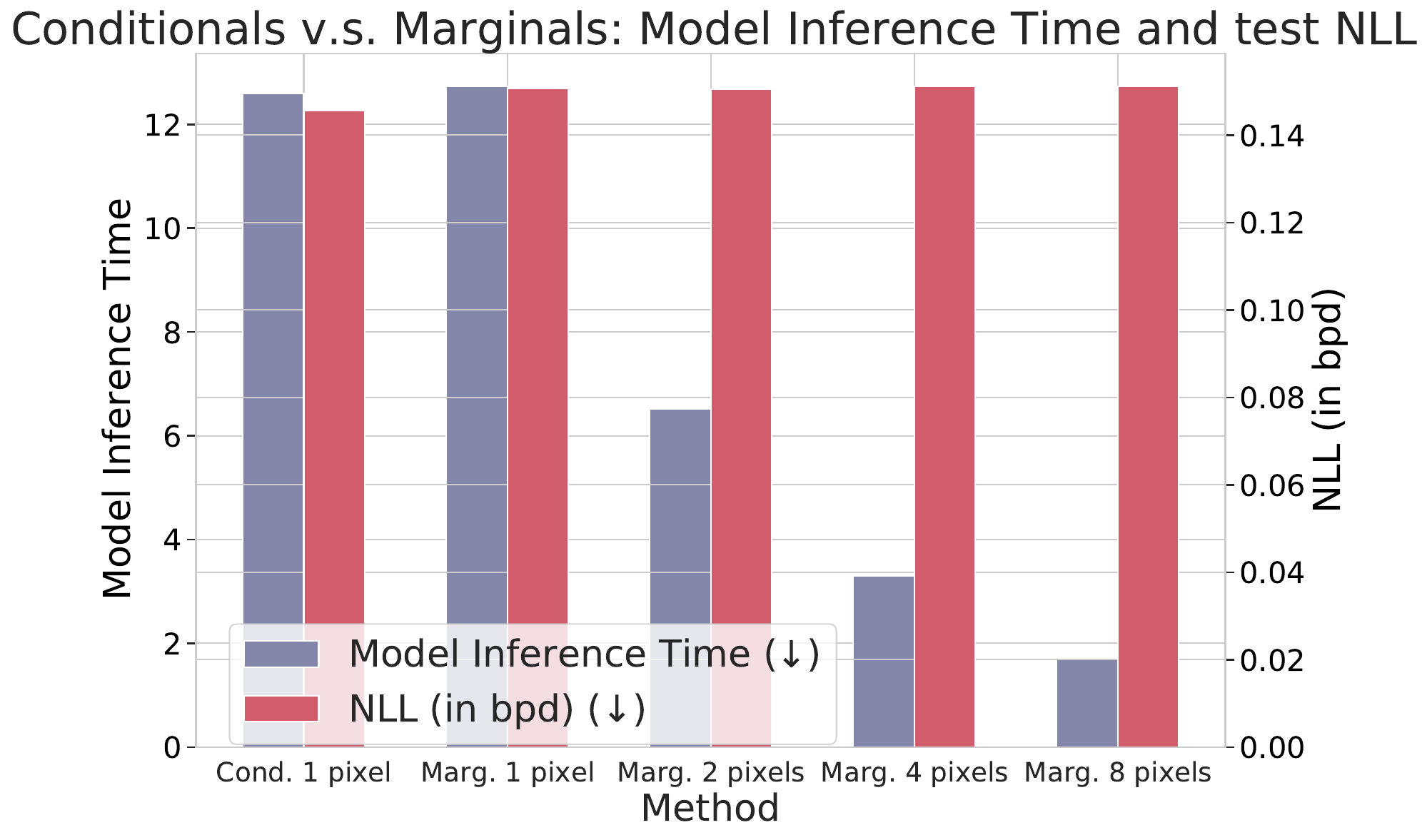}
      \caption{Inference time and NLL comparison}
  \end{subfigure}
  \caption{\small \ours{} sampling using marginal network (a-d) with different number of variables at each step {v.s.} sampling using conditional network (e) with $1$ variable at eacth step. (f) compares NLL under different sampling procedures and the model inference time.}
  \label{fig:marginal_sampling}
\end{figure}

The trained marginalization model comes with two networks. The conditional network $\phi$ estimates any-order conditionals $p_\phi(\boldx_{\sigma (d)} | \boldx_{\sigma (< d)})$, and the marginal network $\theta$ estimates arbitrary marginals $p_\theta(\boldx_{\sigma (\leq d)})$. When \ours{} is used for sampling, either network can be used. With the conditional network $\phi$, samples can be drawn autoregressively one variable at each step. Or the marginals can be used to draw variables using the normalized conditional:
\begin{equation*}
    p_\theta(\boldx_{s_i} | \boldx_{s(<i)}) = \frac{p_\theta([\boldx_{s_i},\boldx_{s(<i)}])}{\textstyle\sum\nolimits_{\boldx_{s_i}} p_\theta([\boldx_{s_i},\boldx_{s(<i)}])} \,.
\end{equation*}
where $\boldx_{s_i}$ is the next block of variables (can be multiple) to sample at step $i$ and $\boldx_{s(<i)}$ are the previously sampled variables. We show with experiments that the marginals are also effective to be used for sampling and they provide extra flexibility in the sampling procedure. We test sampling with different block sizes using the marginals with random orderings and compare them to sampling with conditionals in \Cref{fig:marginal_sampling}. 
The samples generated are of similar quality. And those different sampling procedures exhibit similar likelihood on test data. However, sampling with large block size enables to trade compute memory for less time spent (due to fewer steps) in generation inference, which we find it interesting to explore for future work. Compared with the conditional network, the marginal network allows sampling in arbitrary block variable size and ordering. This illustrates the potential utility of \ours{}s in flexible generation with tractable likelihood.

\subsection{Choice of $q$ in sampling the marginalization self-consistency for training}
\label{sec:pick_q}

In simple examples such as the synthetic checkerboard problem, it does not really matter, we have tried $p_\text{data}$ or $p_\theta$ or random, or a mixture of them. All work fairy well given that the problem is relatively easy.

In real-world data problems, it boils down to what the marginal will be used for at test time. Uniform distribution over $x$ will be a bad choice if there is a data manifold we care about. If it will be used for generation, for example in the MNIST Binary example in \Cref{sec:sample_with_marginal}, $q$ is set to a mixture of $p_\theta$ and $p_\text{data}$. If it will be used for mariginal inference on the data manifold, $p_\text{data}$ will be enough. We all know the NN is not robust on data it hasn’t trained on, and so are the marginal networks, they will not give correct estimates if we evaluate on arbitrary datapoint off the manifold or policy.

\subsection{Two-Stage v.s. Joint Training}
\label{subsec:exp_two_stage_joint}

\begin{figure}[ht!]
  \centering
  \begin{subfigure}{0.35\linewidth}
    \centering
    \includegraphics[width=\linewidth]{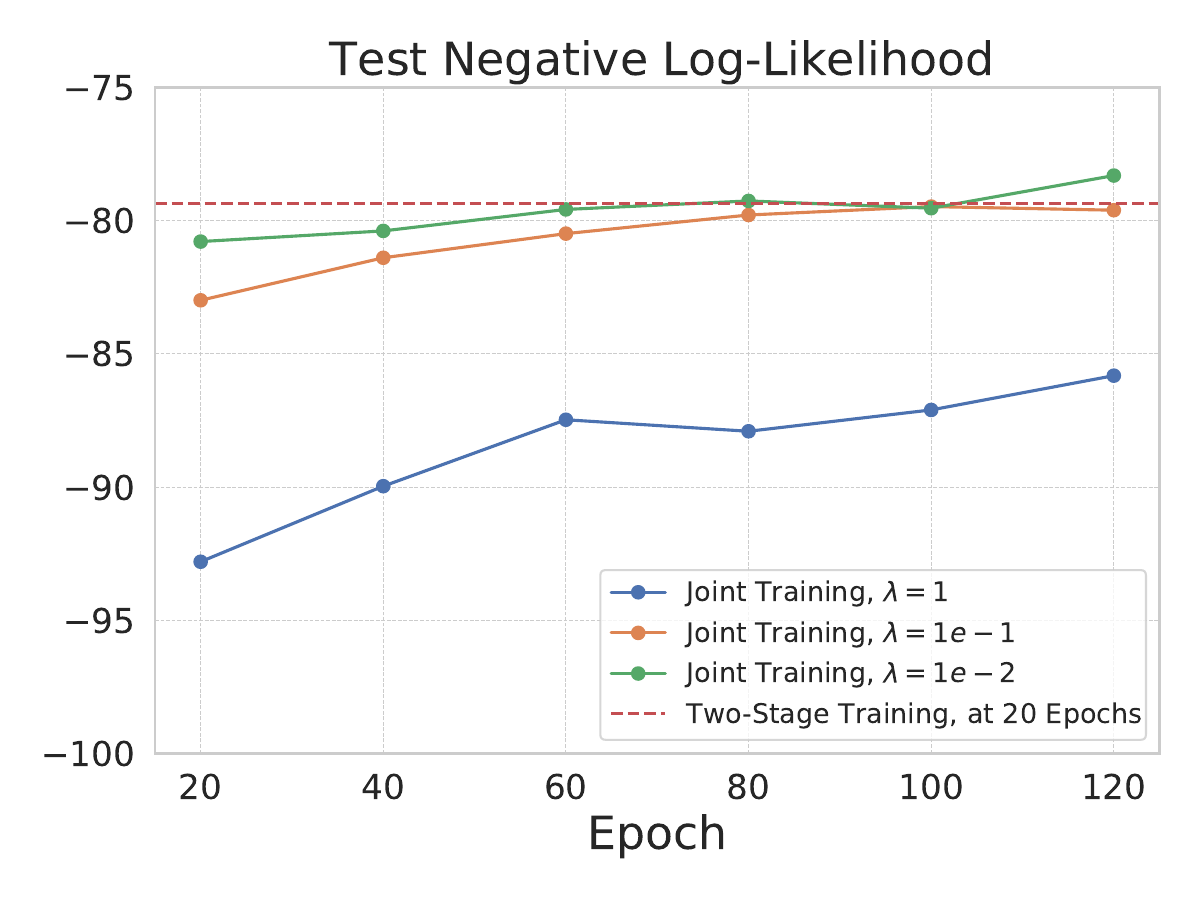}
  \end{subfigure}
    \hspace{2mm}
  \begin{subfigure}{0.35\linewidth}
    \centering
    \includegraphics[width=\linewidth]{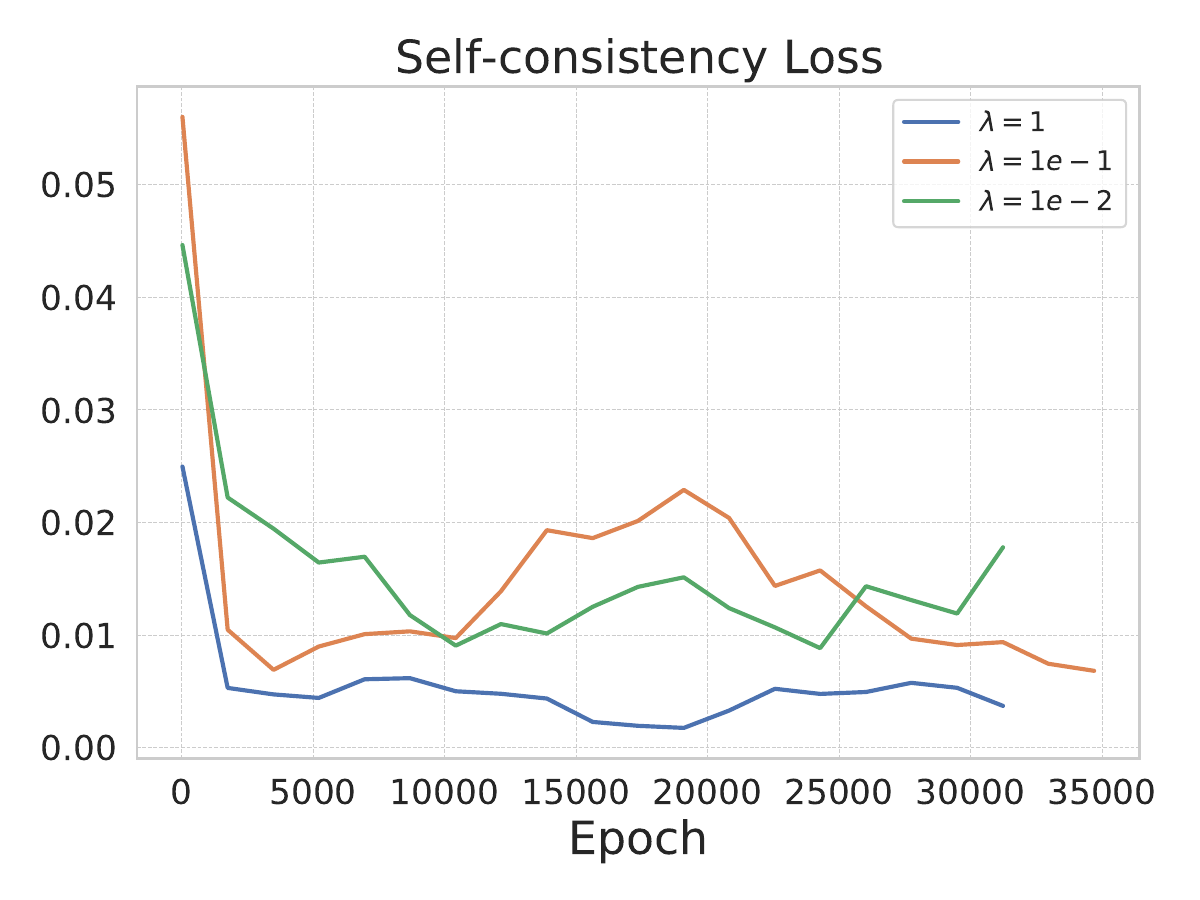}
  \end{subfigure}
  \caption{\small Two-stage training v.s. joint training on MNIST-Binary maximum likelihood training. $\lambda$ is the penalty hyperparameter of self-consistency error term.}
  \label{fig:mnist-two-stage-joint}
\end{figure}

On MNIST maximum likelihood training, we compare two-stage training and joint training in \Cref{fig:mnist-two-stage-joint}. Both training uses the decomposed conditionals for the log likelihood objective, otherwise joint training will lead to inflated log likelihoods. We observe that two-stage training converges faster than joint training (20 epochs v.s. 80 epochs) and needs less GPU memory since it only requires gradient of one model instead of two models. For joint training, it is observed that smaller $\lambda$ is preferred for fast convergence and better performance while large $\lambda$ hurts the model's inference performance.

\begin{figure}[ht!]
  \centering
  \begin{subfigure}{0.35\linewidth}
    \centering
    \includegraphics[width=\linewidth]{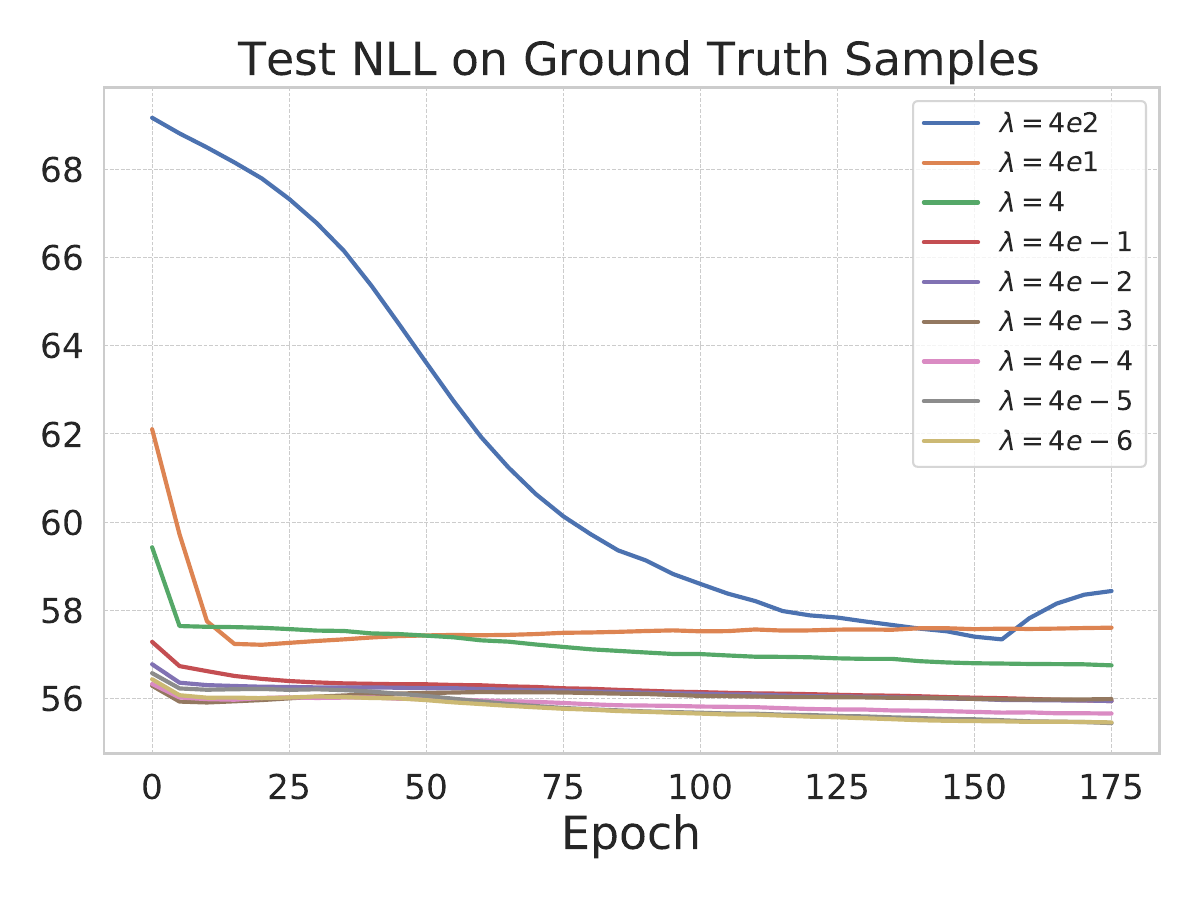}
  \end{subfigure}
    \hspace{2mm}
  \begin{subfigure}{0.35\linewidth}
    \centering
    \includegraphics[width=\linewidth]{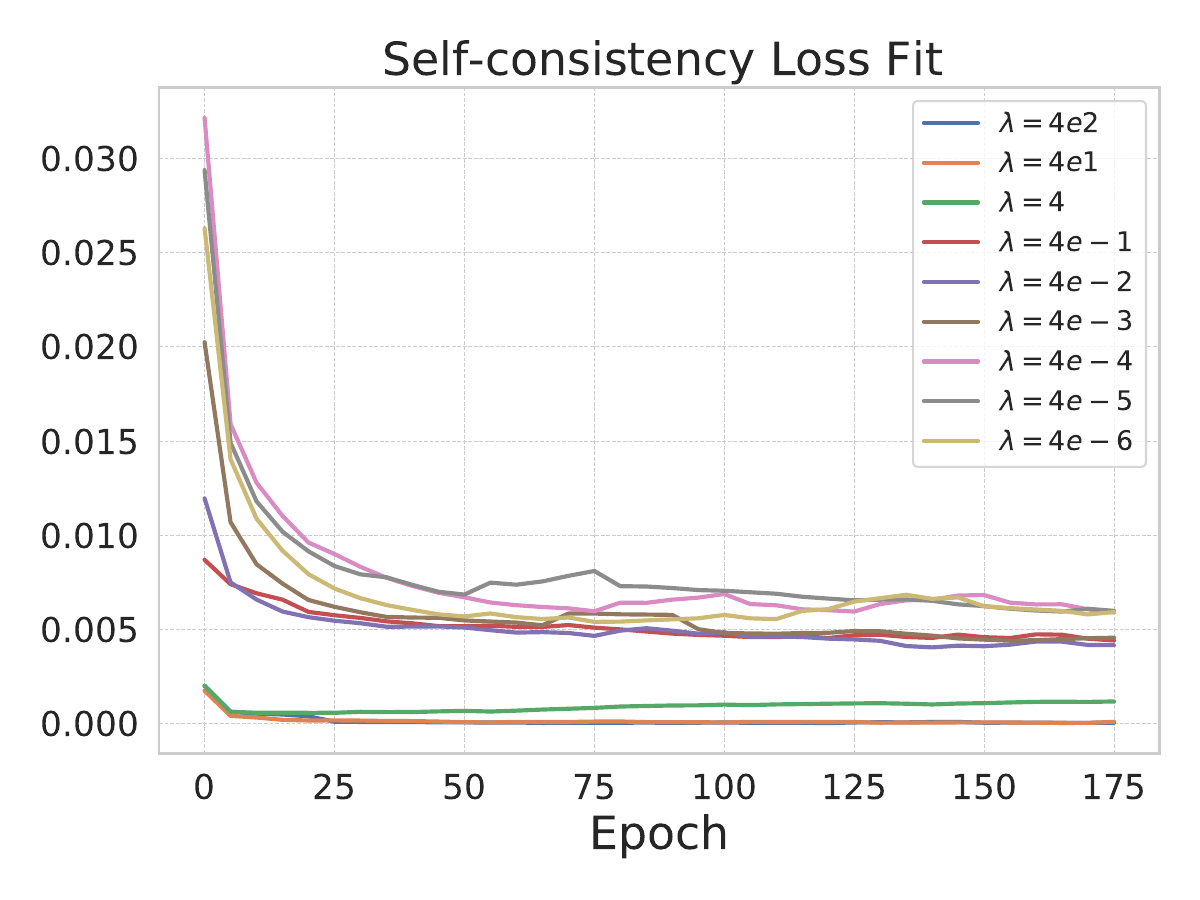}
  \end{subfigure}
  \caption{\small Two-stage training v.s. joint training on Ising Model $10 \times 10$ energy-based training. $\lambda$ is the penalty hyperparameter of self-consistency error term.}
  \label{fig:ising-lambda}
\end{figure}

For joint training under energy-based training setting, we empirically test out how $\lambda$ affects models performance. We find that a wide range of small $\lambda$ leads to best results. See \Cref{fig:ising-lambda} for training dynamics of different $\lambda$ values. Our hypothesis is that $\mathcal L_\text{KL}$ is easier to fit than $\mathcal L_\text{SC}$, since it only involves fitting one term instead of many constraints. When $\lambda$ is relatively small, $\mathcal L_\text{KL}$ is closely fitted first, then training objective is left with $\lambda \mathcal L_\text{SC}$. Since optimization with Adam is scale-invariant, the training converges to similar solutions. When $\lambda$ is too large, $\mathcal L_\text{SC}$ is first fitted very close to $0$, but this restricts the flexibility of the conditionals and marginals to fit $\mathcal L_\text{KL}$ well, hence hurting its generative performance.

\section{Additional Experiments Details}
\label{sec:app_exp}

\begin{table}[htbp!]
  \centering
  \caption{Length Extrapolation on Text8}
  \label{tab:text8-extrapolation}
  \begin{tabular}{lccc}
    \toprule
    Method & Spearman $\uparrow$ & Pearson $\uparrow$ & Time (s) $\downarrow$ \\
    \midrule
    \ours{} & 0.947 & 0.944 & 0.006 \\
    P-AO-ARM $T = 2$ & 0.859 & 0.854 & 2.223 \\
    P-AO-ARM $T = 5$ & 0.923 & 0.931 & 5.683 \\
    P-AO-ARM $T = 10$ & 0.927 & 0.931 & 11.63 \\
    P-AO-ARM $T = 20$ & 0.957 & 0.970 & 23.28 \\
    AO-ARM $T = 300$ & 0.969 & 0.966 & 349.7 \\
    \bottomrule
  \end{tabular}
\end{table}

\subsection{Dataset details}

\textbf{Binary MNIST}\quad
Binary MNIST is a dataset introduced in~\citep{salakhutdinov2008quantitative} that stochastically set each pixel to ``$1$'' or ``$0$'' in proportion to its pixel intensity. We use the training and test split of~\citep{salakhutdinov2008quantitative} provided in~\href{https://github.com/yburda/iwae/tree/master}{https://github.com/yburda/iwae/tree/master}~\citep{burda2015importance}.

\textbf{CIFAR-10}\quad
The CIFAR-10 dataset \citep{krizhevsky2009learning} comprises 60,000 32x32 color images across 10 classes, split into 50,000 training and 10,000 test images. It's used for image recognition and classification tasks in machine learning.

\textbf{ImageNet32}\quad
ImageNet32 \citep{deng2009imagenet,chrabaszcz2017downsampled} is a downsampled variant of the ImageNet dataset, resized to 32x32 pixels. It maintains the diversity of the original with over 14 million images across thousands of categories, but in a lower resolution for computational efficiency.

\textbf{Molecular Sets}\quad 
The molecules in MOSES are represented either in SMILES~\citep{weininger1989smiles} or SELFIES~\citep{krenn2020self} strings. We construct a vocabulary (including a stop token) from all molecules and use discrete valued strings to represent molecules. It is worth noting that \ours{} can also be applied for modeling molecules at a coarse-grained level with predefined blocks, which we leave for future work.

The test set used for evaluating likelihood estimate quality is constructed in a similar manner to Binary MNIST, by drawing sets of random samples from the test dataset.

\textbf{text8}\quad 
In this dataset, we use a vocabulary of size $27$ to represent the letter alphabet with an extra value to represent spaces. 

The test set of datasets used for evaluating likelihood estimate quality is constructed in a similar manner to Binary MNIST, each set is generated by randomly masking out portions of a test text sequence (by $50$, $100$, $150$, $200$ tokens) and generating samples.

\textbf{Ising model}\quad 
The Ising model is defined on a 2D cyclic lattice. The $\boldJ$ matrix is defined to be $\sigma \boldA_N$, where $\sigma$ is a scalar and $\boldA_N$ is the adjacency matrix of a $N \times N$ grid. Positive $\sigma$ encourages neighboring sites to have the same spins and negative $\sigma$ encourages them to have opposite spins. The bias term $\boldsymbol{\theta}$ places a bias towards positive or negative spins. In our experiments, we set $\sigma$ to $0.1$ and $\boldsymbol{\theta}$ to $\mathbf{1}$ scaled by $0.2$.
Since we only have access to the unnormalized probability, we generate $2000$ samples following~\citep{grathwohl2021oops} using Gibbs sampling with $1,000,000$ steps for $10 \times 10$ and $30 \times 30$ lattice sizes. Those data serve as ground-truth samples from the Ising model for evaluating the test log-likelihood.

\textbf{Molecular generation with target property}\quad
During training, we need to optimize on the loss objective on samples generated from the neural network model. However, if the model generates SMILES strings, not all strings correspond to a valid molecule, which makes training at the start challenging when most generated SMILES strings are invalid molecules. Therefore, we use SELFIES string representation as it is a $100\%$ robust in that every SELFIES string corresponds to a valid molecule and every molecule can be represented by SELFIES.

\subsection{Training details}

\textbf{Binary MNIST, CIFAR10, ImageNet32}
\begin{itemize}
    \item Pixel values are converted to scalar values as input. 
    ``$0$'', ``$1$'' for Binary MNIST, ``$0-255$'' for CIFAR-10 and ImageNet.
    ``$\sv$'' takes the value $0$. For each pixel, there is an additional mask indicating if it is a ``$\sv$''. 
    \item U-Net with 4 ResNet Blocks for MNIST, 32 ResNet Blocks for CIFAR-10 sand ImageNet, interleaved with attention layers for both AO-ARM and \ours{}. \ours{} uses two separate neural networks for learning marginals $\phi$ and conditionals $\theta$. Input resolution is $1 \times 28 \times 28$ or $3 \times 32 \times 32$ with $256$ channels used.
    \item The mask is concatenated to the input. $3/4$ of the channels are used to encode input. The remaining $1/4$ channels encode the mask cardinality (see~\cite{hoogeboom2021autoregressive} for details).
    \item \ours{} first learns the conditionals~$\phi$ and then learns the marginals~$\theta$ by finetuning on the downsampling blocks and an additional MLP with $2$ hidden layers of dimension $4096$. We observe it is necessary to distill the marginals by not only finetuning on the additional MLP but also on the downsampling blocks to get a good fitting of the marginal probability, which shows marginal network and conditional network rely on different features to make the final prediction.
    \item Batch size is $128$ for MNIST and $32$ for CIFAR-10 and ImageNet. Adam is used with learning rate $0.0001$. Gradient clipping is set to $100$. Both AO-ARM and \ours{} conditionals are trained for $100$ epochs on MNIST, $800$ epochs on CIFAR-10, $16$ epochs on ImageNet. \ours{} marginals are finetuned from the trained conditionals for $25$ epochs on MNIST, $25$ epochs on CIFAR-10 and $3$ epochs on ImageNet. 
    
    The effectiveness of the proposed two-stage training is validated during experiments. Distilling marginals from conditionals are much faster and easier than learning conditionals and marginals jointly from scratch. And distilling marginals require much fewer epochs than fitting the conditionals.
\end{itemize}

\textbf{MOSES and text8}
\begin{itemize}
    \item Transformer with $12$ layers, $768$ dimensions, $12$ heads, $3072$ MLP hidden layer dimensions for both AO-ARM and \ours{}. Two separate networks are used for \ours{}.
    \item SMILES or SELFIES string representation and ``$\sv$'' are first converted into one-hot encodings as input to the Transformer.
    \item \ours{} first learns the conditionals~$\phi$ and then learns the marginals~$\theta$ by finetuning on the MLP of the Transformer.
    \item Batch size is $512$ for MOSES and $256$ for text8.
    \item AdamW is used with learning rate $0.0005$, betas $0.9/0.99$, weight decay $0.001$. Gradient clipping is set to $0.25$. Both AO-ARM and \ours{} conditionals are trained for $1600$ epochs for text8 and $200$ epochs for MOSES. The \ours{} marginals are finetuned from the trained conditionals for $200$ epochs.
\end{itemize}

\textbf{Ising model and molecule generation with target property}
\begin{itemize}
    \item Ising model input are of $\{0,1,\sv\}$ values and are one-encoded as input to the neural network. The same is done for molecule SELFIES strings.
    \item MLP with residual layers, $3$ hidden layers, feature dimension is $2048$ for Ising model. $6$ hidden layers, feature dimension $4096$ for molecule target generation. 
    \item Adam is used with learning rate of $0.0001$. Batch size is $512$ and $4096$ for molecule target generation. ARM, GFlowNet and \ours{} are trained with $19,800$ steps for the Ising model. ARM and \ours{} are trained with $3,000$ steps for molecule target generation.
    \item Separate networks are used for conditionals and marginals of \ours{}. They are trained jointly with penalty parameter $\lambda$ set to $4$.
\end{itemize}

\textbf{Compute}
\begin{itemize}
    \item All models are trained on a single NVIDIA A100. The evaluation time is tested on an NVIDIA GTX 1080Ti.
\end{itemize}

\subsection{Testing details}
\textbf{Batch size for measuring Spearman and Pearson correlation}
\begin{itemize}
    \item The Spearman and Pearson correlation are measured on batch size of $16$ and averaged over $20$ random batches.
    \item For each batch, pre-screening is performed such that there exist no two samples having similar log-likelihoods that are statistically indistinguishable.  The thresholds are set according to the input dimension of the problem. $10.0$ is used for text8, CIFAR-10, ImageNet32. $5.0$ is used for MNIST-Binary. $2.0$ is used for MOSES.
\end{itemize}

\subsection{Additional results on Images}
\label{sec:app_image}

\subsubsection{CIFAR-10}

\begin{figure}[ht!]
  \centering
  \begin{subfigure}{0.90\linewidth}
    \centering
    \includegraphics[width=\linewidth]{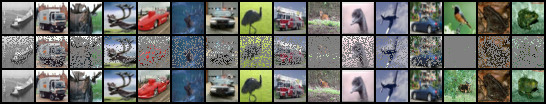}
  \end{subfigure}
  \caption{\small. {CIFAR-10: conditional generation.}}
    \label{fig:cifar10cond}
\end{figure}

\begin{figure}[ht!]
  \centering
  \begin{subfigure}{0.40\linewidth}
    \centering
    \includegraphics[width=\linewidth]{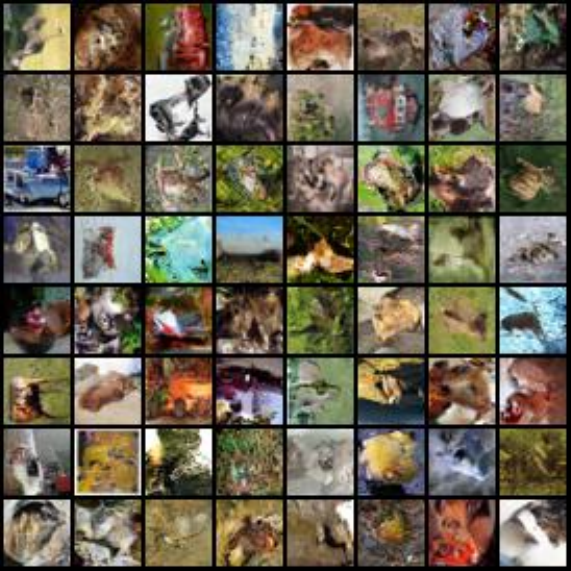}
  \end{subfigure}
  \caption{\small. {CIFAR-10: generated samples. Note that sometimes images are flipped because MaM is trained on augmented images.}}
    \label{fig:cifar10}
\end{figure}

We train MaMs conditionals for $800$ epochs and then further train $25$ epochs to fit the marginals. 
\ours{} achieves a test NLL of $2.88$ bpd (if we continue training to $3000$ epochs, test NLL will get close to $2.69$ bpd shown in the AO-ARM literature \citep{hoogeboom2021autoregressive}). 
Test NLL is compared in \Cref{tab:cifar10}. MaM achieve highly correlations in terms of $\log p$ estimate when compared with AO-ARM $\log p$'s. 
The marginal self-consistency error is averaged $\sim 0.3$ in $\log p$ values. 
Generated samples are shown in \Cref{fig:cifar10cond} and \Cref{fig:cifar10}.

\subsubsection{ImageNet32}

\begin{figure}[ht!]
  \centering
  \begin{subfigure}{0.90\linewidth}
    \centering
    \includegraphics[width=\linewidth]{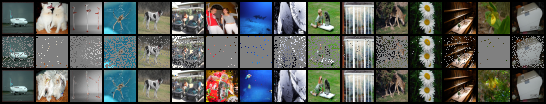}
  \end{subfigure}
  \caption{\small. {ImageNet32: conditional generation.}}
    \label{fig:imagenet_cond}
\end{figure}

\begin{figure}[ht!]
  \centering
  \begin{subfigure}{0.40\linewidth}
    \centering
    \includegraphics[width=\linewidth]{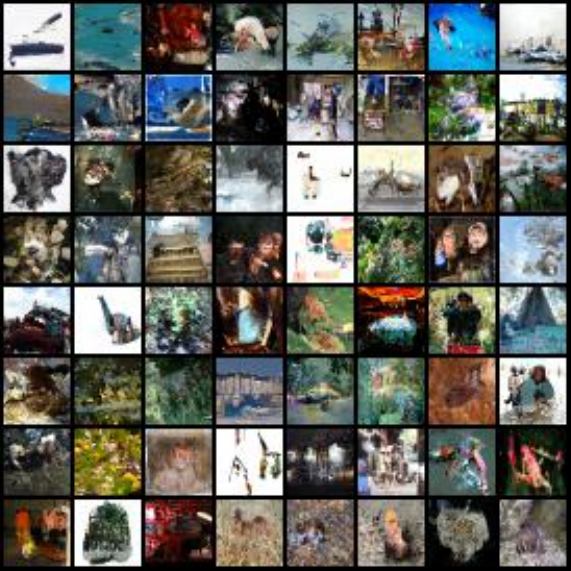}
  \end{subfigure}
  \caption{\small. {ImageNet32: generated samples.}}
    \label{fig:imagenet}
\end{figure}

We train MaMs conditionals for $16$ epochs and train $3$ more epochs for fitting the marginals.
\ours{} achieves a test NLL of $2.88$ bpd (if we continue training to $3000$ epochs, test NLL will get close to $2.69$ bpd shown in the AO-ARM literature \citep{hoogeboom2021autoregressive}). 
Test NLL is compared in \Cref{tab:imagenet}. MaM achieve highly correlations in terms of $\log p$ estimate when compared with AO-ARM $\log p$'s. 
The marginal self-consistency error is averaged $\sim 0.3$ in $\log p$ values. 
Generated samples are shown in \Cref{fig:imagenet_cond} and \Cref{fig:imagenet}. 

\subsubsection{Binary MNIST}
\label{subsec:mnist-app}
\textbf{Likelihood estimate on partial Binary MNIST images}

\begin{figure}[ht]
  \centering
  \includegraphics[width=1\textwidth]{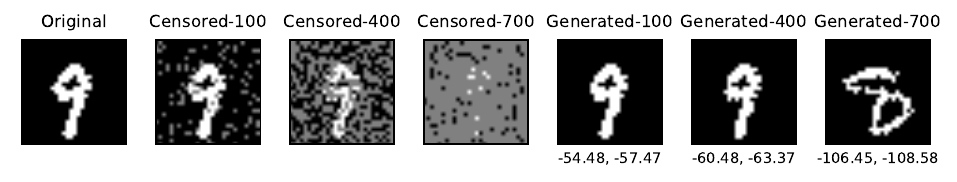}
  \caption{\small An example of the data generated (with $100/400/700$ pixels masked) for comparing the quality of likelihood estimate. Numbers below the images are LL estimates from \ours{}'s marginal network (left) and AO-ARM-E's ensemble estimate (right).}
  \label{fig:mnist-syn-examples}
\end{figure}

\begin{figure}[htbp!]
  \centering
  \includegraphics[width=0.7\textwidth]{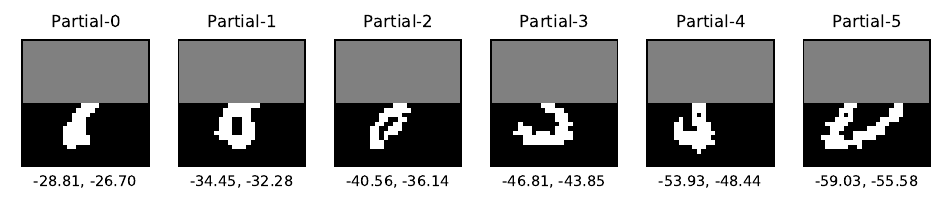}
  \caption{\small An example set of partial images for evaluating marginal likelihood estimate quality. The numbers in the captions show the log-likelihood calculated using learned marginals (left) v.s. learned conditionals (right)}
  \label{fig:partial-mnist}
\end{figure}

Figure~\ref{fig:partial-mnist} illustrates an example set of partial images that we evaluate and compare likelihood estimate from \ours{} against ARM. Table~\ref{tab:mnist-partial} contains the comparison of the marginal likelihood estimate quality and inference time.

\begin{table}[htbp!]
  \centering
  \caption{Marginal estimates on Binary-MNIST partial images}
  \label{tab:mnist-partial}
  \begin{tabular}{lcc}
    \toprule
    &{Pearson $\uparrow$}  & {Marg. inf. time (s) $\downarrow$}\\
    \midrule
    AO-ARM & {0.997} & 49.75 $\pm$ 0.03\\
    \ours{}  & {0.995} & {0.02 $\pm$ 0.00} \\
    \bottomrule
  \end{tabular}
\end{table}

\textbf{Likelihood estimate on synthetic Binary MNIST images}

Figure~\ref{fig:mnist-syn-examples} illustrates an example of ``synthetic'' MNIST images generated from masked MNIST images that we evaluate and compare likelihood estimate from \ours{} against ARM. Table~\ref{tab:mnist-syn} shows the marginal likelihood estimate shows strong correlation with actual $\log p$ from ARM, demonstrating strong generalizing to data on the manifold but not seen during training. 

\begin{table}[htbp!]
  \centering
  \caption{Marginal estimates on Binary-MNIST ``synthetic'' images}
  \label{tab:mnist-syn}
  \begin{tabular}{lcc}
    \toprule
    &{Pearson $\uparrow$} \\
    \midrule
    AO-ARM & {0.993}\\
    \ours{}  & {0.993}\\
    \bottomrule
  \end{tabular}
\end{table}

\textbf{Generated samples}

\begin{figure}[htbp!]
  \centering
  \includegraphics[width=1\textwidth]{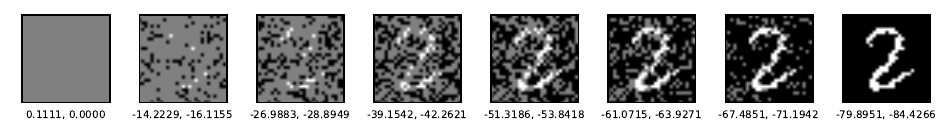}
  \caption{\small An example of the trajectory every $112$ step when generating an MNIST digit following a random order. The future pixels are generated by conditioning on the existent filled-in pixels. The numbers in the captions show the log-likelihood calculated using learned marginals (left) v.s. learned conditionals (right)}
  \label{fig:gen-mnist}
\end{figure}

\begin{figure}[htbp!]
  \centering
  \includegraphics[width=1\textwidth]{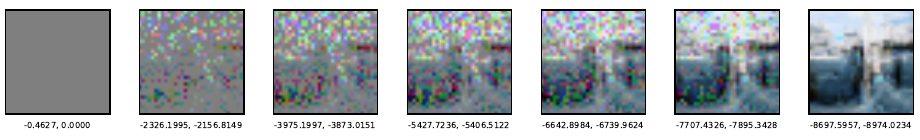}
  \caption{\small An example of the trajectory when generating an ImageNet image following a random order. The future pixels are generated by conditioning on the existent filled-in pixels. The numbers in the captions show the log-likelihood calculated using learned marginals (left) v.s. learned conditionals (right).}
  \label{fig:gen-imagenet}
\end{figure}

\begin{figure}[htbp!]
  \centering
  \includegraphics[width=0.8\textwidth]{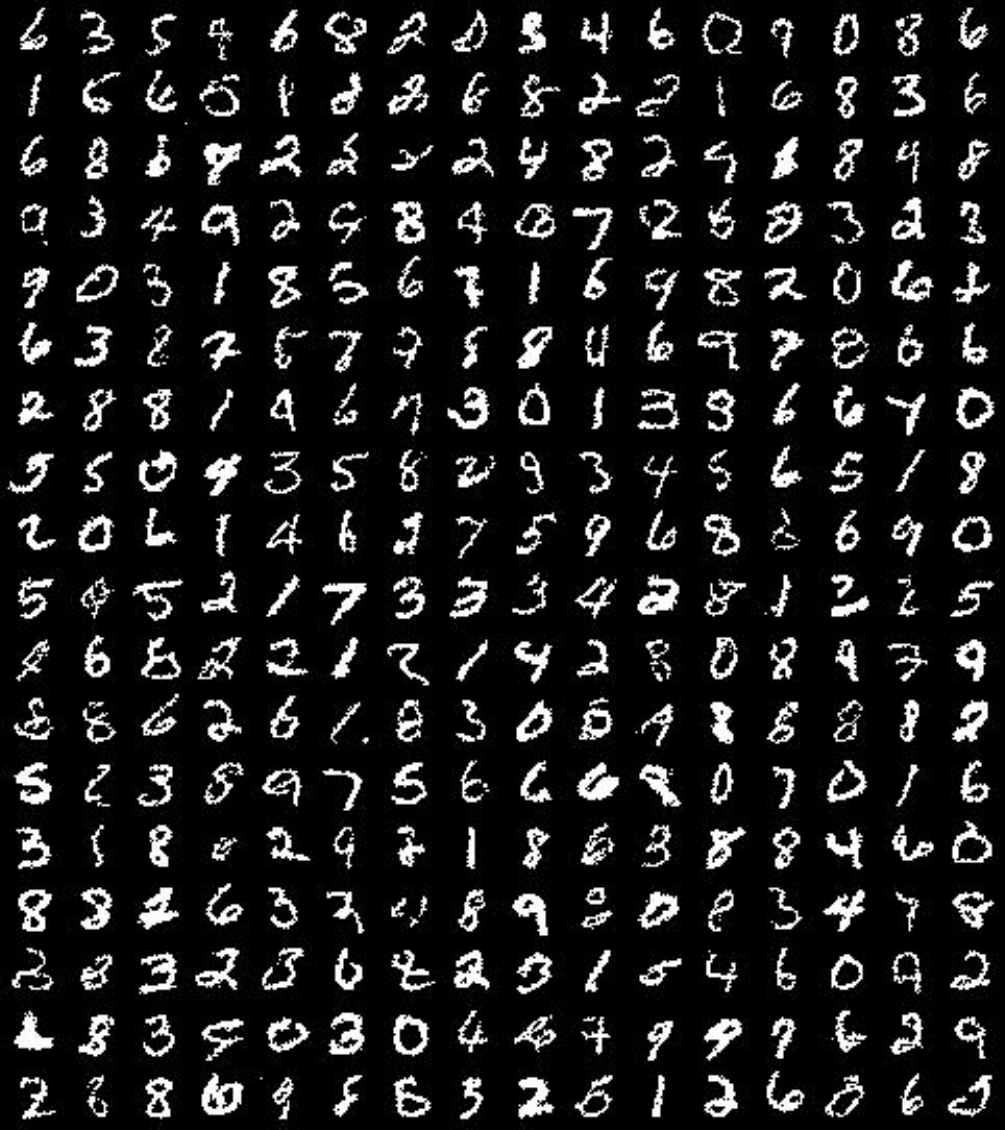}
  \caption{\small Generated samples: Binary MNIST}
  \label{fig:samples-mnist}
\end{figure}

Figure~\ref{fig:gen-mnist} shows how a digit is generated pixel-by-pixel following a random order. 
We show generated samples from \ours{} using the learned conditionals $\phi$ in Figure~\ref{fig:samples-mnist}.

\subsection{Additional results on MOSES}
\label{sec:moses_additional_results}

\begin{table}[t]
  \centering
  \caption{Character modeling on Molecular Sets}
  \label{tab:moses}
  \begin{tabular}{lccc}
    \toprule
    & {NLL (bpc) $\downarrow$} &{Pearson $\uparrow$}  & {Time (s) $\downarrow$}\\
    \midrule
    AO-ARM$^\am$ & {0.655} & 0.994 &  19.32$\pm$ 0.01\\
    \rowcolor{shadecolor} \ours{}$^\am$ & {0.655} & {0.995} & {0.006$\pm$0.00} \\
    \bottomrule
  \end{tabular}
\end{table}

\begin{table}[htbp!]
  \centering
  \caption{Performance Comparison on MOSES}
  \label{tab:moses-sota}
  \begin{tabularx}{\linewidth}{l*{8}{X}}
    \toprule
    \textbf{Model} & \textbf{Valid$\uparrow$} & \textbf{Unique 10k$\uparrow$} & \textbf{Frag Test$\downarrow$} & \textbf{Scaf TestSF$\uparrow$} & \textbf{Int Div1$\uparrow$} & \textbf{Int Div2$\uparrow$} & \textbf{Filters$\uparrow$} & \textbf{Novelty$\uparrow$} \\
    \midrule
    Training data        & 1.0   & 1.0     & 1.0   & 0.9907 & 0.8567  & 0.8508  & 1.0  & 1.0   \\
    HMM          & 0.076 & 0.5671  & 0.5754 & 0.049   & 0.8466  & 0.8104  & 0.9024  & \textbf{0.9994}  \\
    NGram        & 0.2376 & 0.9217 & 0.9846 & 0.0977  & \textbf{0.8738}  & \textbf{0.8644}  & 0.9582  & \textbf{0.9694}  \\
    CharRNN      & \textbf{0.9748} & 0.9994 & \textbf{0.9998} & \textbf{0.1101}  & \textbf{0.8562}  & \textbf{0.8503}  & \textbf{0.9943}  & 0.8419  \\
    JTN-VAE      & \textbf{1.0}   & \textbf{0.9996}  & 0.9965 & \textbf{0.1009}  & 0.8551  & 0.8493  & \textbf{0.976}   & 0.9143  \\
    \ours{}-SMILES  & 0.7192 & \textbf{0.9999} & \textbf{0.9978} & \textbf{0.1264}  & 0.8557  & 0.8499  & \textbf{0.9763}  & \textbf{0.9485}  \\
    \ours{}-SELFIES & \textbf{1.0}   & \textbf{0.9999}  & \textbf{0.997}  & 0.0943  & \textbf{0.8684}  & \textbf{0.8625}  & 0.894   & 0.9155  \\
    \bottomrule
  \end{tabularx}
\end{table}

\begin{figure}[htbp!]
  \centering
  \begin{subfigure}{0.45\linewidth}
    \centering
    \includegraphics[width=\linewidth]{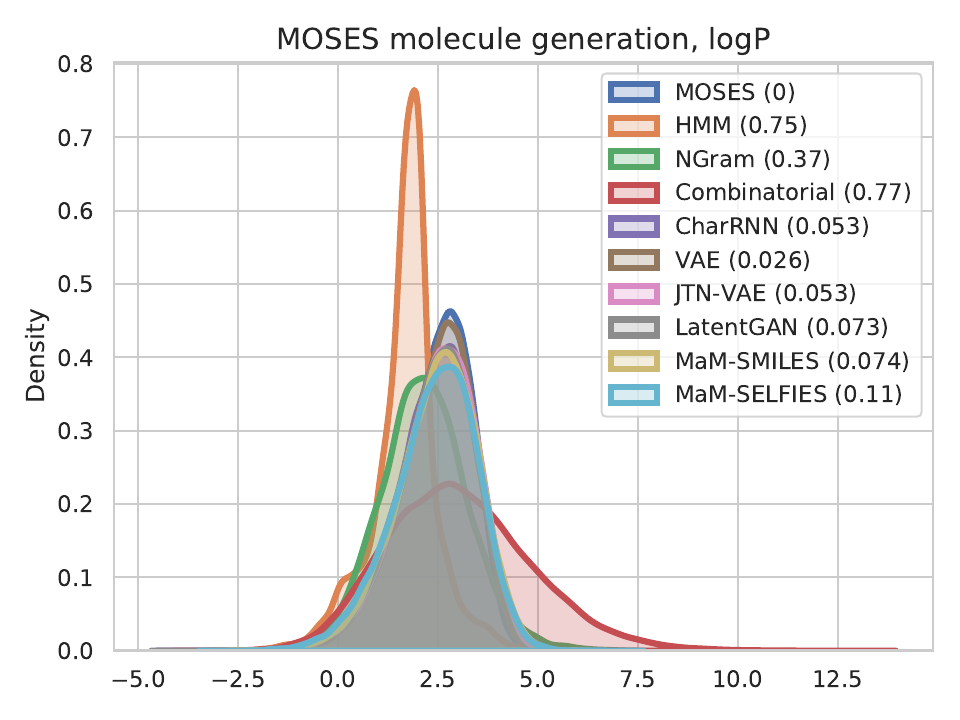}
  \end{subfigure}
  \hspace{2mm}
  \begin{subfigure}{0.45\linewidth}
    \centering
    \includegraphics[width=\linewidth]{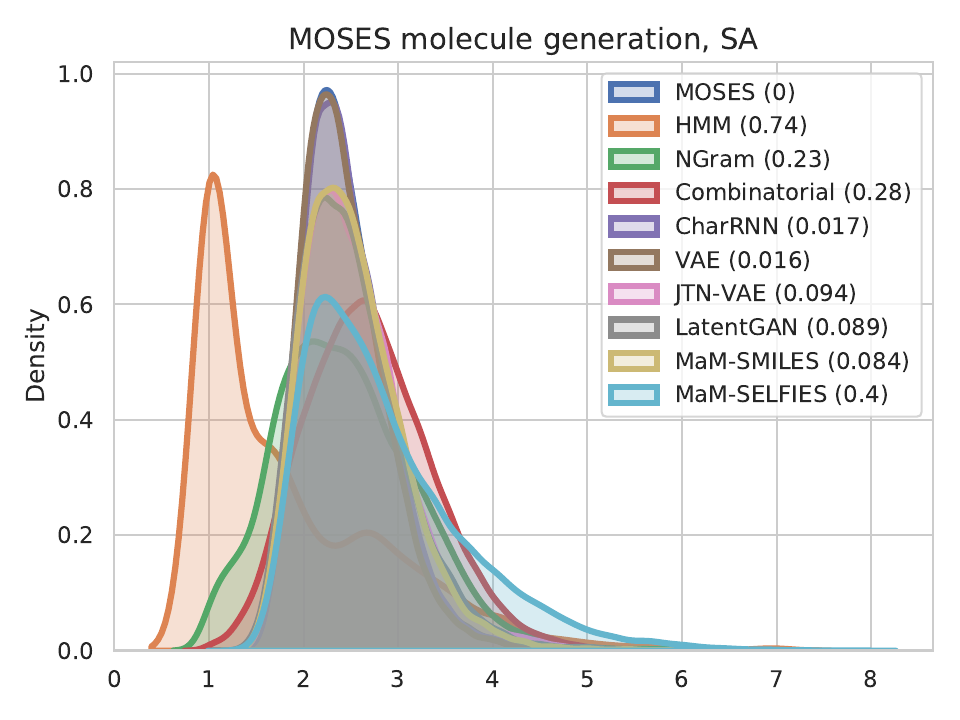}
  \end{subfigure}
  \vspace{0.5cm}
  \begin{subfigure}{0.45\linewidth}
    \centering
    \includegraphics[width=\linewidth]{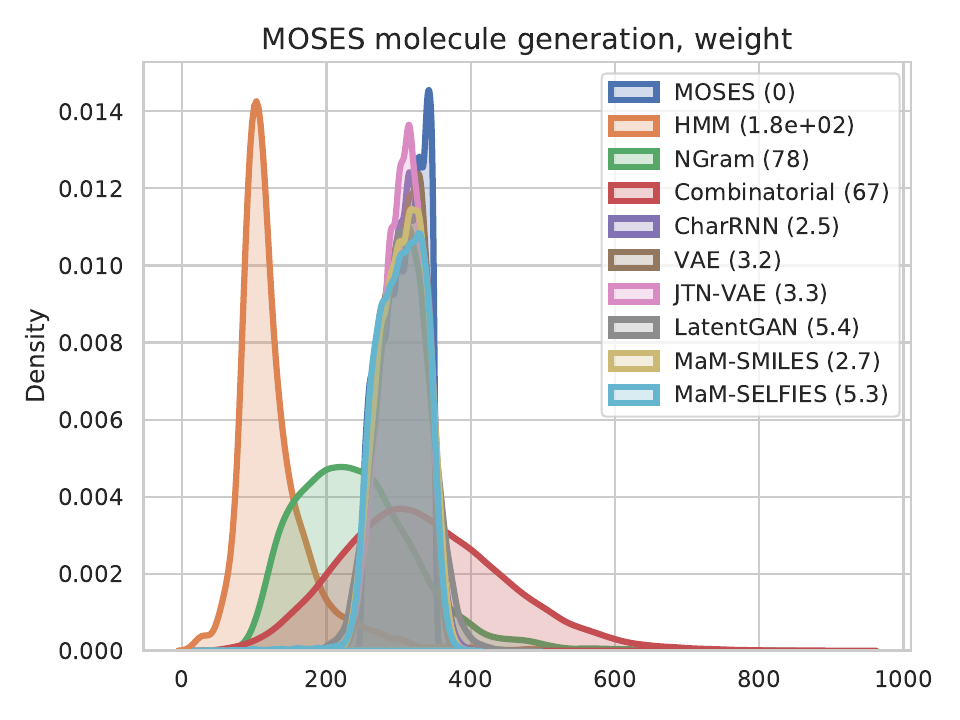}
  \end{subfigure}
  \hspace{2mm}
  \begin{subfigure}{0.45\linewidth}
    \centering
    \includegraphics[width=\linewidth]{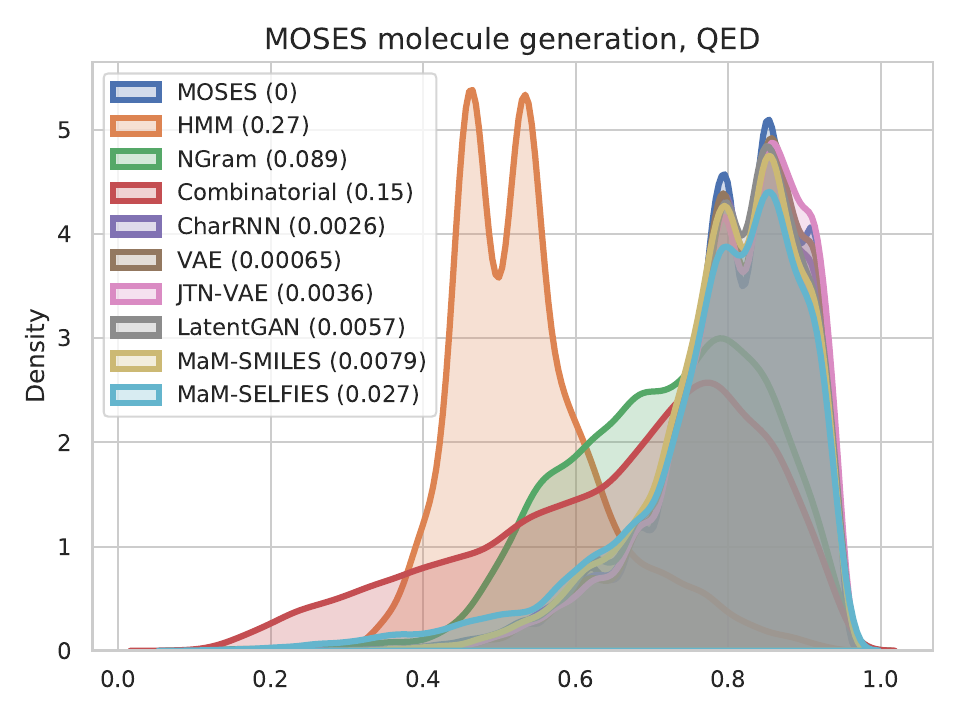}
  \end{subfigure}
  \caption{\small KDE plots of lipophilicity (logP), Synthetic Accessibility (SA), Quantitative Estimation of Drug-likeness (QED), and molecular weight for generated molecules. $30,000$ molecules are generated for each method.}
  \label{fig:moses-sota}
  \vspace{-0.5cm}
\end{figure}

\subsubsection{Comparing \ours{} with SOTA on MOSES molecule generation}

We compare the quality of molecules generated by \ours{} with standard baselines and state-of-the-art methods in Table~\ref{tab:moses-sota} and Figure~\ref{fig:moses-sota}. Details of the baseline methods are provided in~\citep{polykovskiy2020molecular}.
\ours{}-SMILES/SELFIES represents \ours{} trained on SMILES/SELFIES string representations of molecules.
\ours{} performs either better or comparable to SOTA molecule generative modeling methods. The major advantage of \ours{} and AO-ARM is that their order-agnostic modeling enables generation in any desired order of the SMILES/SELFIES string (or molecule sub-blocks).

\subsubsection{Generated molecular samples}

\begin{figure}[hp!]
  \centering
  \includeinkscape[width=1\textwidth]{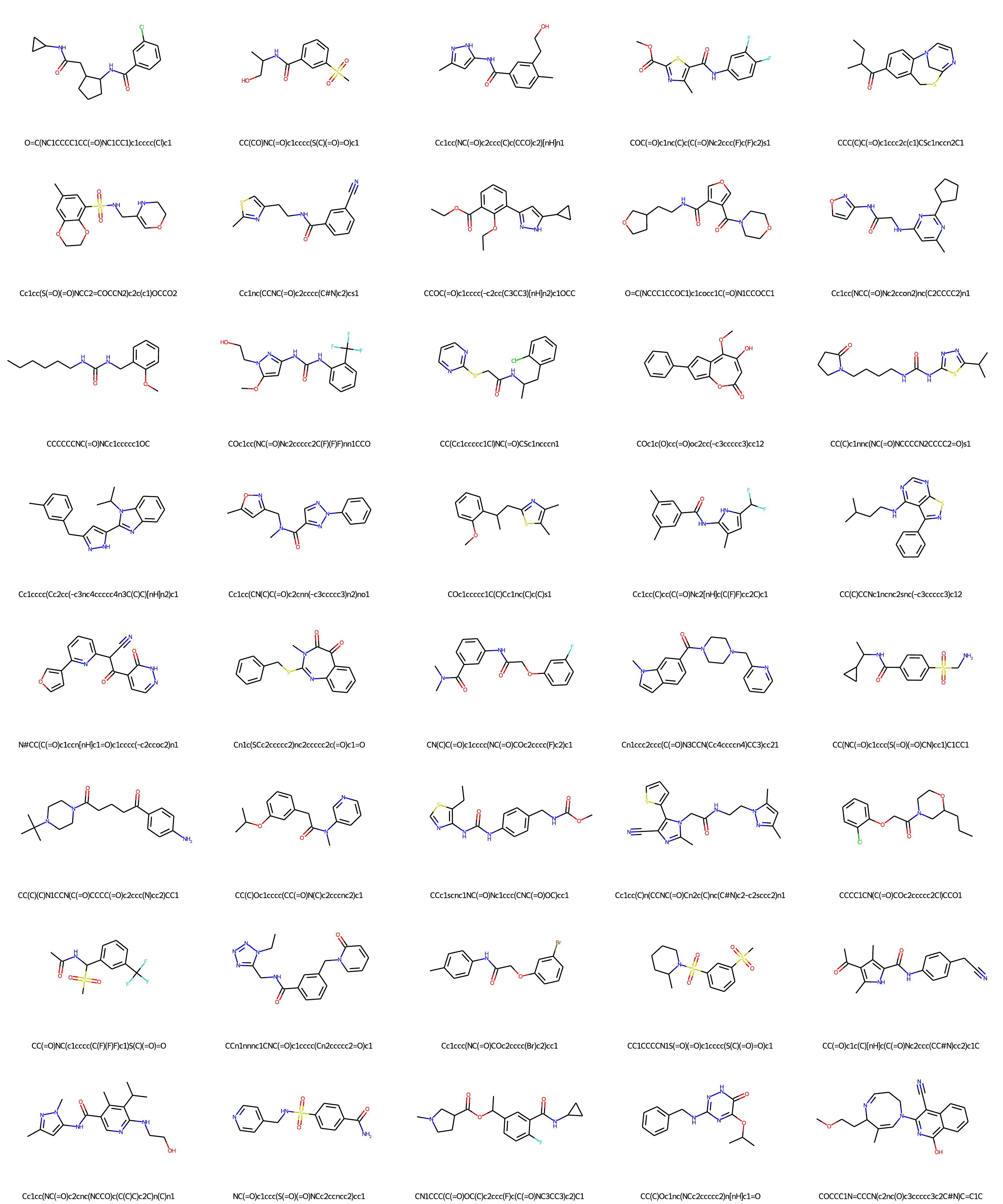}
  \caption{\small Generated samples from \ours{}-SMILES: MOSES}
  \label{fig:samples-moses-smiles}
\end{figure}

\begin{figure}[hp!]
  \centering
  \includeinkscape[width=1\textwidth]{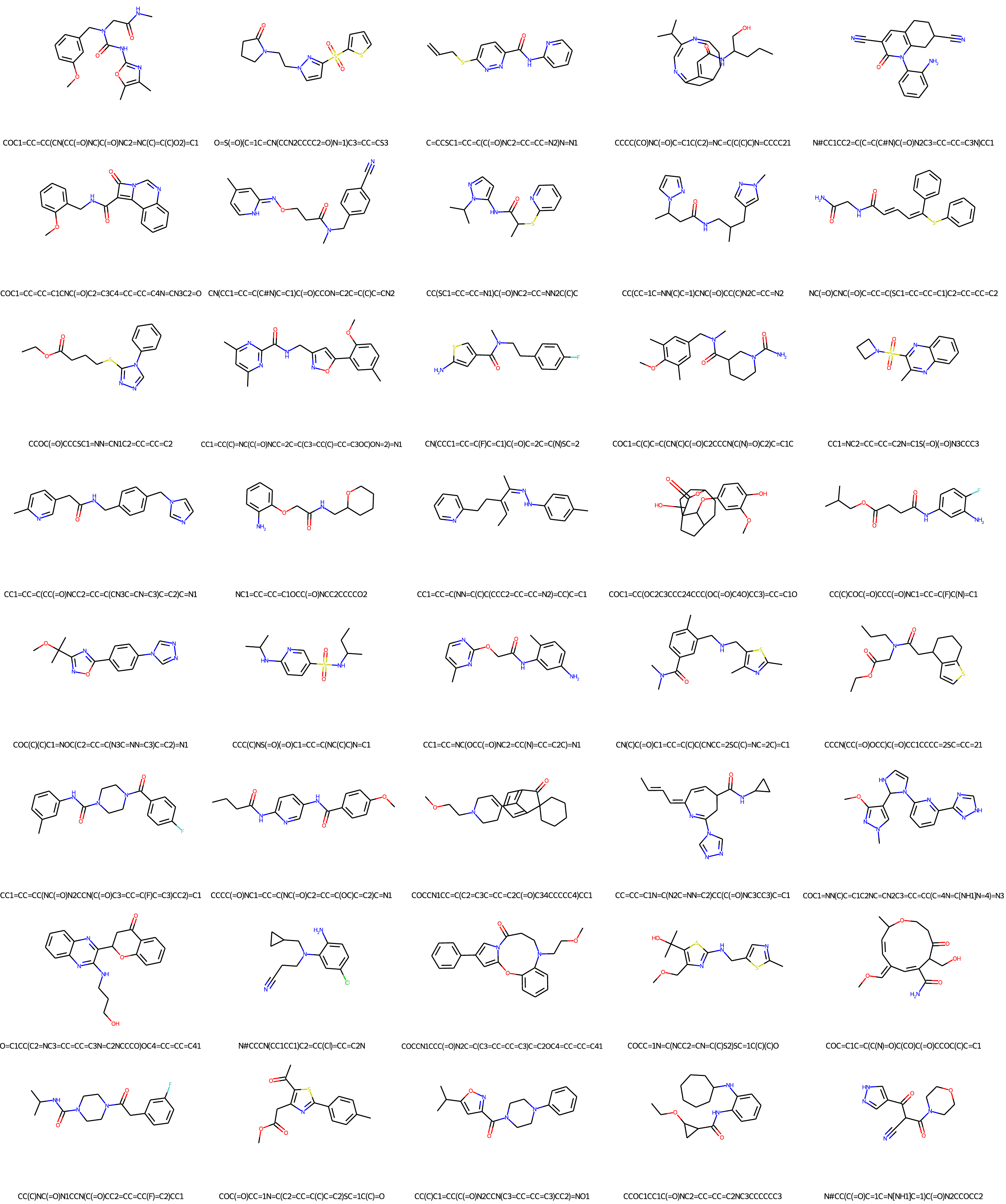}
  \caption{\small Generated samples from \ours{}-SELFIES: MOSES}
  \label{fig:samples-moses-selfies}
\end{figure}

Figure~\ref{fig:samples-moses-smiles} and~\ref{fig:samples-moses-selfies} plot the generated molecules from \ours{}-SMILES and \ours{}-SELFIES.

\subsubsection{Out-of-distribution test on differentiating Drug vs. Photodiode}

We challenged the model with a more difficult OOD task: distinguishing between tyrosine kinase inhibitors (a specific type of drug) and organic photodiodes from focused chemical spaces while the MAM model is trained on a general chemical space of drug-like compounds. The tyrosine kinase inhibitors should be considered by the model to have higher likelihood given that it has more similar properties (such as moderate weight and lipophilicity) to the molecules in ZINC.

We created ~$1000$ pairs consisting of one of each using datasets from \citet{subramanian2023automated}, controlling for other factors like SMILES length and chemical space to increase difficulty. Despite this, MAM's marginals correctly identified the drug molecule 74\% of the time (vs. 79\% for AO-ARM), with 90\% alignment between marginal estimates and AO-ARM $\log p$’s.

\clearpage
\subsection{Additional results on text8}
\label{subsec:text8-samples}

\subsubsection{Length extrapolation on Text8}
\label{subsec:extrapolation}
In \Cref{tab:text8-extrapolation}, we evaluated the model's ability to handle length extrapolation on Text8. We trained data with $D=250$ and tested on sequences with $D=300$.

\textbf{Robustness}\quad
\ours{}'s predicted $\log p$ marginals maintain a high correlation with those calculated using AO-ARM conditionals, even on longer sequences. Its quality matches Parallel AO-ARM with 20 steps.

\textbf{Graceful Extrapolation} \quad
We observe the absolute errors in $\log p_\theta(x) - \log p_\textrm{ARM}(x)$ increase due to challenge from OOD prediction, but the variance of these errors remains surprisingly similar to that observed when $D=250$. This indicates that MAM gracefully extrapolates the relative scales among $\log p$ values, explaining the high observed correlation in the Table.

\subsubsection{Samples used for evaluating likelihood estimate quality}

We show an example of a set of generated samples from masking different portions of the same text, which is then used for evaluating and comparing the likelihood estimate quality. Their log-likelihood calculated using the conditionals with the AO-ARM are in decreasing order. We use \ours{} marginal network to evaluate the log-likelihood and compare its quality with that of the AO-ARM conditionals.

\begin{myquote}
{\fontfamily{courier}\selectfont Original text:}

{\fontfamily{cmtt}\selectfont  the subject of a book by lawrence weschler in one nine nine five entitled mr wilson s cabinet of wonder and the museum s founder david wilson received a macarthur foundation genius award in two zero zero three the museum claims to attract around six}
\end{myquote}

\begin{myquote}
{\fontfamily{courier}\selectfont Text generated from masking out $50$ tokens:}

{\fontfamily{cmtt}\selectfont  \;the\_su\_je\_t of a b\_ok\_by\_la\_r\_nce \_es\_h\_\_\_ \_n o\_\_\_nine n\_ne five entitled mr\_wilson s\_cabinet of wonder and the museum s founder \_\_vid w\_l\_o\_ r\_\_eive\_ a macarthur fou\_\_a\_\_on \_e\_\_\_s\_awa\_d in two \_ero z\_r\_ \_hree \_he museum c\_aims \_o attr\_ct ar\_u\_d s\_\_}
\\
{\fontfamily{cmtt}\selectfont 
\;the subject of a book by lawrence heschell in one nine nine five entitled mr wilson s cabinet of wonder and the museum s founder david wilson received a macarthur foundation dennis award in two zero zero three the museum claims to attract around sev}
\end{myquote}

\begin{myquote}
{\fontfamily{courier}\selectfont Text generated from masking out $100$ tokens:}

{\fontfamily{cmtt}\selectfont  \_the\_su\_je\_t \_f \_\_b\_\_k\_\_y\_l\_\_r\_nc\_ \_es\_h\_\_\_\_\_n o\_\_\_nine n\_ne five\_
\\
enti\_l\_d
mr\_wil\_o\_ \_\_c\_b\_\_et of wond\_r an\_ \_h\_ mu\_eu\_ s f\_u\_der\_\_\_vid
\\
\_w\_l\_\_\_\_\_\_\_eive\_ a\_maca\_thur f\_u\_\_a\_\_\_n \_e\_\_\_\_\_a\_a\_d \_\_ two \_er\_ z\_r\_ \_h\_ee\_\_\_\_ museum c\_a\_ms\_\_o \_\_tr\_ct ar\_u\_\_ \_\_\_}
\\
{\fontfamily{cmtt}\selectfont  \;the subject of a book by lawrence bessheim in one nine nine five entitled mr wilson s cabinet of wonder and the museum s founder david wilson received a macarthur foundation leaven award in two zero zero three the museum claims to detract around the}
\end{myquote}

\begin{myquote}
{\fontfamily{courier}\selectfont Text generated from masking out $150$ tokens:}

{\fontfamily{cmtt}\selectfont  \_the\_\_u\_\_\_\_t\_\_f \_\_\_\_\_\_\_\_\_\_l\_\_r\_nc\_ \_es\_h\_\_\_\_\_n o\_\_\_n\_\_e n\_ne\_\_ive\_
\\
e\_ti\_l\_\_m\_\_wil\_\_\_\_\_\_c\_\_\_\_et of won\_\_\_ an\_\_\_\_\_\_\_\_\_\_\_\_ s\_\_\_u\_der\_\_\_vid\_
\\
w\_\_\_\_\_\_\_\_\_eiv\_\_ a\_\_a\_a\_th\_\_\_\_\_\_\_\_a\_\_\_n \_e\_\_\_\_\_a\_\_\_\_\_\_\_ two\_\_e\_\_ z\_r\_ \_\_\_e\_\_\_\_\_ \_use\_m c\_a\_ms\_\_\_ \_\_tr\_ct\_a\_\_\_\_\_ \_\_\_}
\\
{\fontfamily{cmtt}\selectfont
\;the tudepot of europe de laurence desthefs in one nine nine five entitled mr wild the cabinet of wonder anne cedallica s founder david wright received arnasa the culmination team sparked in two zero zero three the museum claims to retract athlet c a}
\end{myquote}

\begin{myquote}
{\fontfamily{courier}\selectfont Text generated from masking out $200$ tokens:}

{\fontfamily{cmtt}\selectfont
\_t\_\_\_\_\_\_\_\_\_\_\_\_f \_\_\_\_\_\_\_\_\_\_l\_\_r\_\_\_\_ \_\_\_\_\_\_\_
\_\_\_\_\_o\_\_\_\_\_\_e\_\_n\_\_\_iv\_\_
\\
e\_\_i\_l\_\_
\_\_\_\_wil\_\_\_\_\_\_c\_\_\_\_\_t\_\_\_ w\_\_\_\_\_\_a\_\_\_\_\_\_\_\_\_\_\_\_\_\_\_\_\_\_\_\_der\_\_\_\_\_d\_
\\
w\_\_\_\_\_
\_\_\_\_e\_\_\_\_\_\_\_\_a\_a\_\_\_\_\_\_\_\_\_\_\_a\_\_\_n\_e\_\_\_\_\_a\_\_\_\_\_\_\_t\_\_\_\_\_\_\_\_\_\_\_\_\_
\\
\_\_\_e\_\_\_\_\_ \_u\_e\_\_\_c\_a\_\_s\_\_\_ \_\_\_r\_c\_\_a\_\_\_\_\_ \_\_\_}
\\
{\fontfamily{cmtt}\selectfont
\;the builder of the pro walter a a e sec press one nine nine five esciele the wild men convert of wark flax notes the world undergroand whirl spiken america ascent and martin decree a letter to the antler s default museum chafes in america ascent vis}
\end{myquote}

\subsection{Additional experiments on Ising model}

\textbf{Generated samples}

\begin{figure}[t]
  \centering
  \begin{subfigure}{0.35\linewidth}
    \centering
    \includegraphics[width=\linewidth]{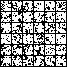}
  \end{subfigure}
  \hspace{2mm}
  \begin{subfigure}{0.35\linewidth}
    \centering
    \includegraphics[width=\linewidth]{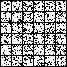}
  \end{subfigure}
  \caption{\small Samples: $10 \times 10$ Ising model. Ground truth (left) v.s. \ours{} (right).}
  \label{fig:samples-ising-10d}
\end{figure}

\begin{figure}[t]
  \centering
  \begin{subfigure}{0.35\linewidth}
    \centering
    \includegraphics[width=\linewidth]{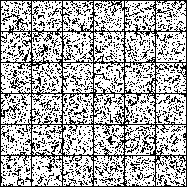}
  \end{subfigure}
    \hspace{2mm}
  \begin{subfigure}{0.35\linewidth}
    \centering
    \includegraphics[width=\linewidth]{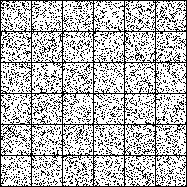}
  \end{subfigure}
  \caption{\small Samples: $30 \times 30$ Ising model. Ground truth (left) v.s. \ours{} (right).}
  \label{fig:samples-ising-30d}
\end{figure}

We compare ground truth samples and \ours{} samples in Figure~\ref{fig:samples-ising-10d} and~\ref{fig:samples-ising-30d}.

\subsection{Additional experiments on molecule target generation}

\subsubsection{Target property energy-based training on lipophilicity (logP)}

\begin{figure}[ht!]
  \centering
  \begin{subfigure}{0.35\linewidth}
    \centering
    \includegraphics[width=\linewidth]{figures/mols/sample_score_logp_55d_tau1.0.pdf}
  \end{subfigure}
    \hspace{2mm}
  \begin{subfigure}{0.35\linewidth}
    \centering
    \includegraphics[width=\linewidth]{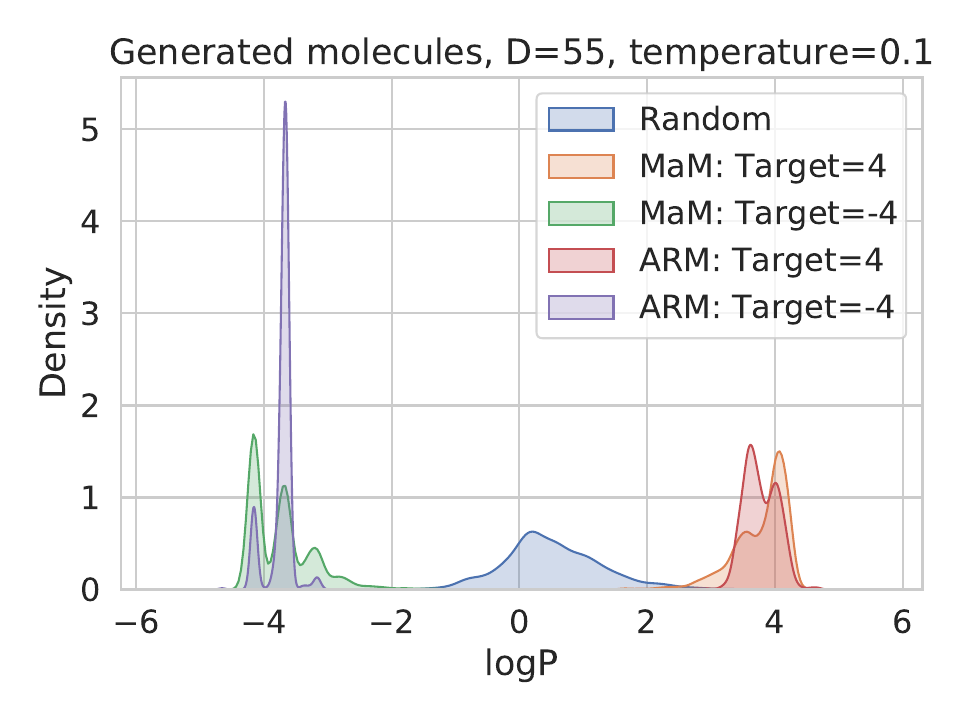}
  \end{subfigure}
  \caption{\small Target property matching with different temperatures. $2000$ samples are generated for each method.}
  \label{fig:mol-55d-logp}
\end{figure}

\begin{figure}[ht!]
  \centering
  \begin{subfigure}{0.35\linewidth}
    \centering
    \includegraphics[width=\linewidth]{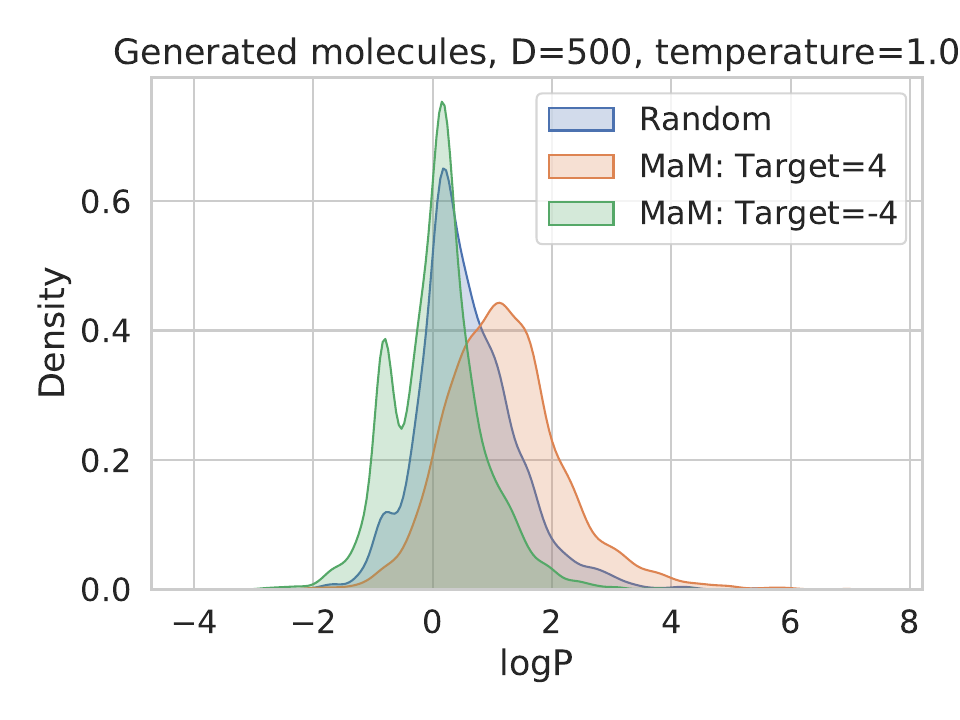}
  \end{subfigure}
    \hspace{2mm}
  \begin{subfigure}{0.35\linewidth}
    \centering
    \includegraphics[width=\linewidth]{figures/mols/sample_score_logp_500d_tau0.1.pdf}
  \end{subfigure}
  \caption{\small Target property matching with different temperatures. $2000$ samples are generated for each method.}
  \label{fig:mol-500d-logp}
\end{figure}

Figure~\ref{fig:mol-55d-logp} and~\ref{fig:mol-500d-logp} show the logP of generated samples of length $D=55$ towards target values $4.0$ and $-4.0$ under distribution temperature $\tau =1.0$ and $\tau=0.1$. For $\tau=1.0$, the peak of the probability density (mass) appears around $2.0$ (or $-2.0$) because there are more valid molecules in total with that logP than molecules with $4.0$ (or $-4.0$), although a single molecule with $4.0$ (or $-4.0$) has a higher probability than $2.0$ (or $-2.0$). When the temperature is set to much lower ($\tau=0.1$), the peaks concentrate around $4.0$ (or $-4.0$) because the probability of logP value being away from $4.0$ (or $-4.0$) quickly diminishes to zero. We additionally show results on molecules of length $D=500$. In this case, logP values are shifted towards the target but their peaks are closer to $0$ than when $D=55$, possibly due to the enlarged molecule space containing more molecules with logP around 0. Also, this is validated by the result when $\tau=0.1$ for $D=500$, the larger design space allows for more molecules with logP values that are close to, but not precisely, the target value.

\subsubsection{Conditionally generated samples}

\begin{figure}[htp!]
  \centering
  \includeinkscape[width=1\textwidth]{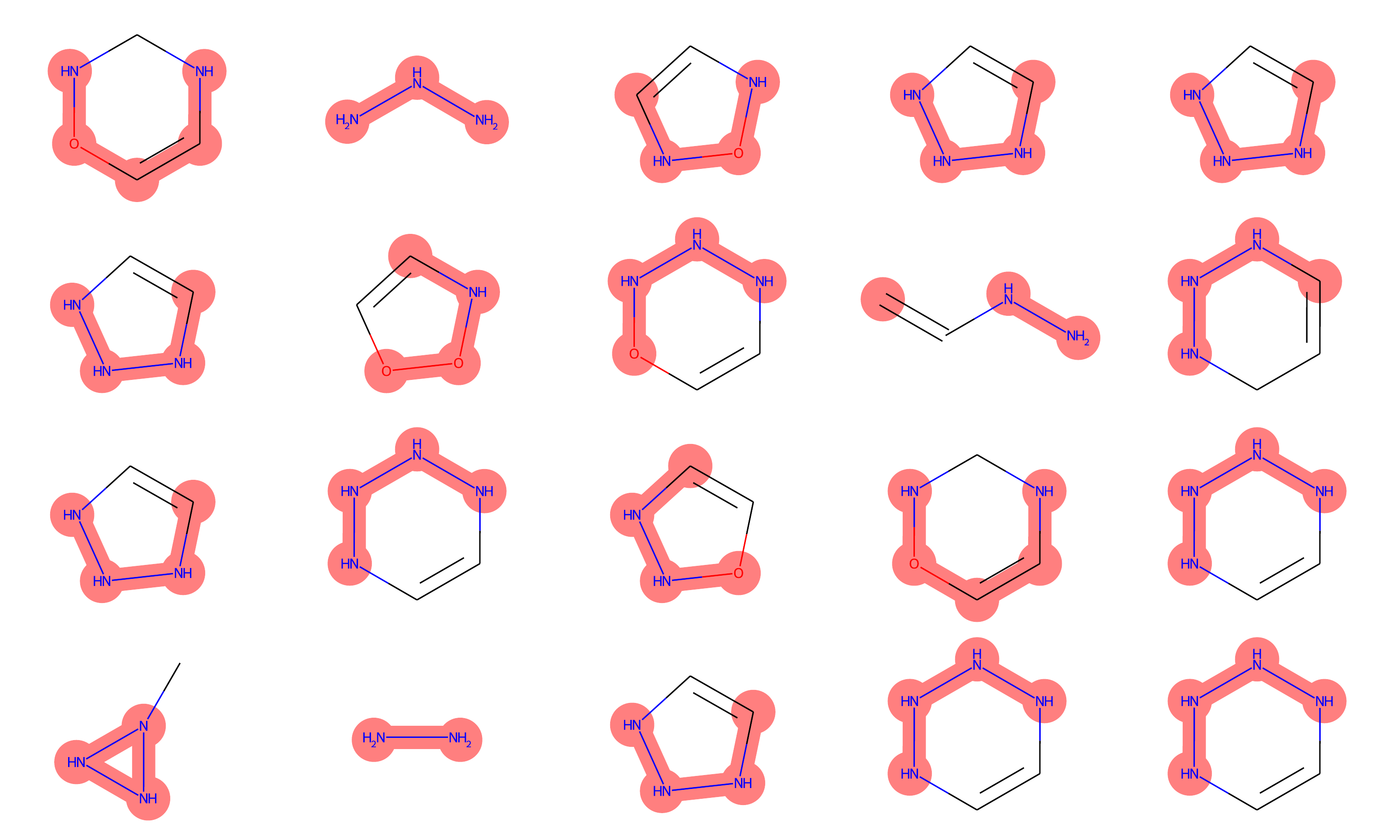}
  \caption{\small Generated samples from masking out the left 4 SELFIES characters of a Benzene. Shaded region are the impainted structures.}
  \label{fig:dm_mol_cond_gen_left}
\end{figure}

\begin{figure}[htp!]
  \centering
  \includeinkscape[width=1\textwidth]{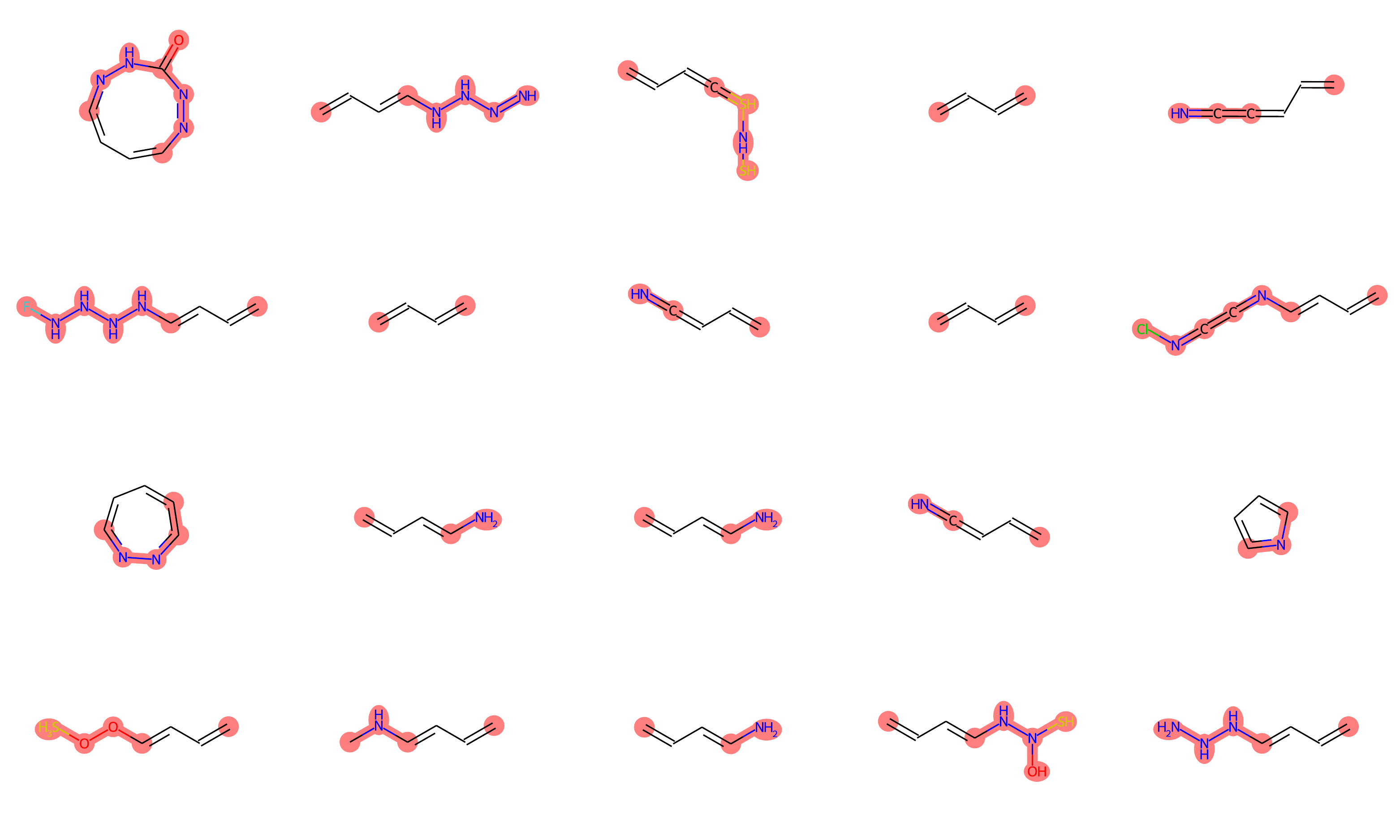}
  \caption{\small Generated samples from masking out the right 4-20 SELFIES characters of a Benzene. Shaded region are the impainted structures.}
  \label{fig:dm_mol_cond_gen_right}
\end{figure}

More samples from conditionally generating towards low lipophilicity ($\text{target}=-4.0$, $\tau=1.0$) from user-defined substructures of Benzene. We are able to generate from any partial substructures with any-order generative modeling of \ours{}. Figure~\ref{fig:dm_mol_cond_gen_left} shows conditional generation from masking out the left $4$ SELFIES characters. Figure~\ref{fig:dm_mol_cond_gen_right} shows conditional generation from masking the right $4\sim20$ characters.

%% file: 0_algorithm.tex
\begin{figure}[th!] 
\begin{minipage}[t]{1\linewidth}
  \begin{algorithm}[H]
  \caption{MLE training of \ours{}s}
  \begin{algorithmic}
    \State \textbf{Input}: Data $\mcD_\text{train}$, $q(\boldx)$, network $\theta$ and $\phi$
    \State \textbf{Stage 1:} Train $\phi$ with \Cref{eq:oaarm} used in AO-ARM
    \For{minibatch $\boldx \sim \mcD_\text{train}$}
    \State Sample $\sigma \sim \mcU(S_D)$, $d \sim \mcU(1,\cdots,D)$
    \State $\mcL \gets 
    \frac{D}{D-d+1}  \sum\nolimits_{j \in \sigma(\geq d)}^{}  \log p_\phi\left(x_j | \boldx_{\sigma(<d)}\right)$
    \State Update $\phi$ with gradient of $\mcL$
    \EndFor
    \State \textbf{Stage 2:} Train $\theta$ to distill the marginals from optimized conditionals $\phi$
    \For{minibatch $\boldx \sim q(\boldx)$}
    \State Sample $\sigma \sim \mcU(S_D)$, $d \sim \mcU(1,\cdots,D)$
    \State $\mcL \gets$ squared error of the inconsistencies in \Cref{eq:one-step-marginal-conditional}
    \State Update $\theta$ with gradient of $\mcL$
    \EndFor
    \end{algorithmic}
  \label{alg:mle}
  \end{algorithm}
\end{minipage}
\hspace{0.02\linewidth}
\begin{minipage}[t]{1\linewidth}
  \begin{algorithm}[H]
  \caption{Energy-based training of \ours{}s}
  \begin{algorithmic}
    \State \textbf{Input}: $q(\boldx)$, network $\theta$ and $\phi$, Gibbs sampling block size $M$
    \State \textbf{Joint training of $\phi$ and $\theta$:}
    \For{$j$ in $\{1, \cdots, N\}$}
    \State Sample $\sigma \sim \mcU(S_D)$
    \State Update $\boldx \sim p_\phi(\boldx_{\sigma(\leq M)} | \boldx_{\sigma(>M)})$ \\
    \Comment{Persistent block Gibbs sampling}
    \State Sample $\Tilde{\boldx} \sim q(\boldx)$
    \State Sample $\Tilde{d} \sim 
    \mcU(1,\cdots,D)$, $\Tilde{\sigma} \sim \mcU(S_D)$
    \State $\mcL_\text{penalty} \gets$ squared error of \Cref{eq:one-step-marginal-conditional}, for $\Tilde{d}$ and $\Tilde{\sigma}$ with $\Tilde{\boldx}$
    \State $\nabla_{\theta,\phi} D_\text{KL} \gets$ REINFORCE est. with $\boldx$
    \State $\nabla_{\theta,\phi}  \gets \nabla_{\theta,\phi} D_\text{KL} + \lambda \nabla_{\theta,\phi} \mcL_\text{penalty}$
    \State Update $\theta$ and $\phi$ with gradient
    \EndFor
    \end{algorithmic}
    \label{alg:dm}
  \end{algorithm}
\end{minipage}
\end{figure}